\documentclass[10pt,journal,compsoc]{IEEEtran}

\usepackage{algorithm}
\usepackage{algorithmicx}
\usepackage{algpseudocode}
\usepackage{amsmath}
\usepackage{diagbox}
\usepackage{graphics}
\usepackage{epsfig}
\usepackage{threeparttable}
\usepackage{xspace}
\usepackage{graphicx}
\usepackage{amssymb}
\usepackage{threeparttable}
\usepackage{color}
\usepackage[normalem]{ulem}
\usepackage{multirow}
\usepackage{float}
\usepackage{amsfonts}
\usepackage{bm}
\usepackage{array}
\usepackage[table]{xcolor}
\usepackage{colortbl}
\usepackage{cite}
\usepackage[pagebackref=false,breaklinks=true,colorlinks,bookmarks=false]{hyperref}

\usepackage{sidecap}

\def\ie{\emph{i.e.~}}
\def\eg{\emph{e.g.~}}
\def\etal{{\em et al.~}}
\def\aka{{\em a.k.a~}}

\def\figpath{./Figs}

\extrafloats{100}

\definecolor{mygray}{gray}{.8}

\makeatletter
\newcommand{\thickhline}{%
    \noalign {\ifnum 0=`}\fi \hrule height 1pt
    \futurelet \reserved@a \@xhline
}
\makeatother

\makeatletter
\DeclareRobustCommand\onedot{\futurelet\@let@token\@onedot}
\def\@onedot{\ifx\@let@token.\else.\null\fi\xspace}

\def\eg{\emph{e.g}\onedot} 
\def\ie{\emph{i.e}\onedot}

\def\etal{\emph{et al}\onedot}
\makeatother

\begin{document}
%
%
%

\title{Saliency Prediction in the Deep Learning Era: \\
Successes, Limitations, and Future Challenges }

\author{Ali Borji,~\IEEEmembership{Member,~IEEE}
\IEEEcompsocitemizethanks{
\IEEEcompsocthanksitem MarkableAI Inc.
\IEEEcompsocthanksitem Email: aliborji@gmail.com
%
%
}
}

\markboth{IEEE TRANSACTIONS ON PATTERN ANALYSIS AND MACHINE INTELLIGENCE}%
{Shell \MakeLowercase{\textit{et al.}}: Bare Demo of IEEEtran.cls for Journals}

\IEEEtitleabstractindextext{
\begin{abstract}
Visual saliency models have enjoyed a big leap in performance in recent years, thanks to advances in deep learning and large scale annotated data. Despite enormous effort and huge breakthroughs, however, models still fall short in reaching human-level accuracy. In this work, I explore the landscape of the field emphasizing on new deep saliency models, benchmarks, and datasets. A large number of image and video saliency models are reviewed and compared over two image benchmarks and two large scale video datasets. Further, I identify 
factors that contribute to the gap between models and humans and discuss the remaining issues that need to be addressed to build the next generation of more powerful saliency models. 
Some specific questions that are addressed include: in what ways current models fail, how to remedy them, what can be learned from cognitive studies of attention, how explicit saliency judgments relate to fixations, how to conduct fair model comparison, and what are the emerging applications of saliency models.

\end{abstract}

}
\maketitle

\IEEEdisplaynontitleabstractindextext
\IEEEpeerreviewmaketitle

\IEEEraisesectionheading{\section{Introduction}\label{sec:introduction}}

\IEEEPARstart{V}isual attention enables humans to rapidly analyze complex scenes and devote their limited perceptual and cognitive resources to the most pertinent subsets of sensory data. It acts as a shiftable information processing bottleneck, allowing only objects within a circumscribed region to reach higher levels of processing and visual awareness~\cite{koch1987shifts}.  

\begin{figure}
\centering
        \includegraphics[width=1\linewidth]{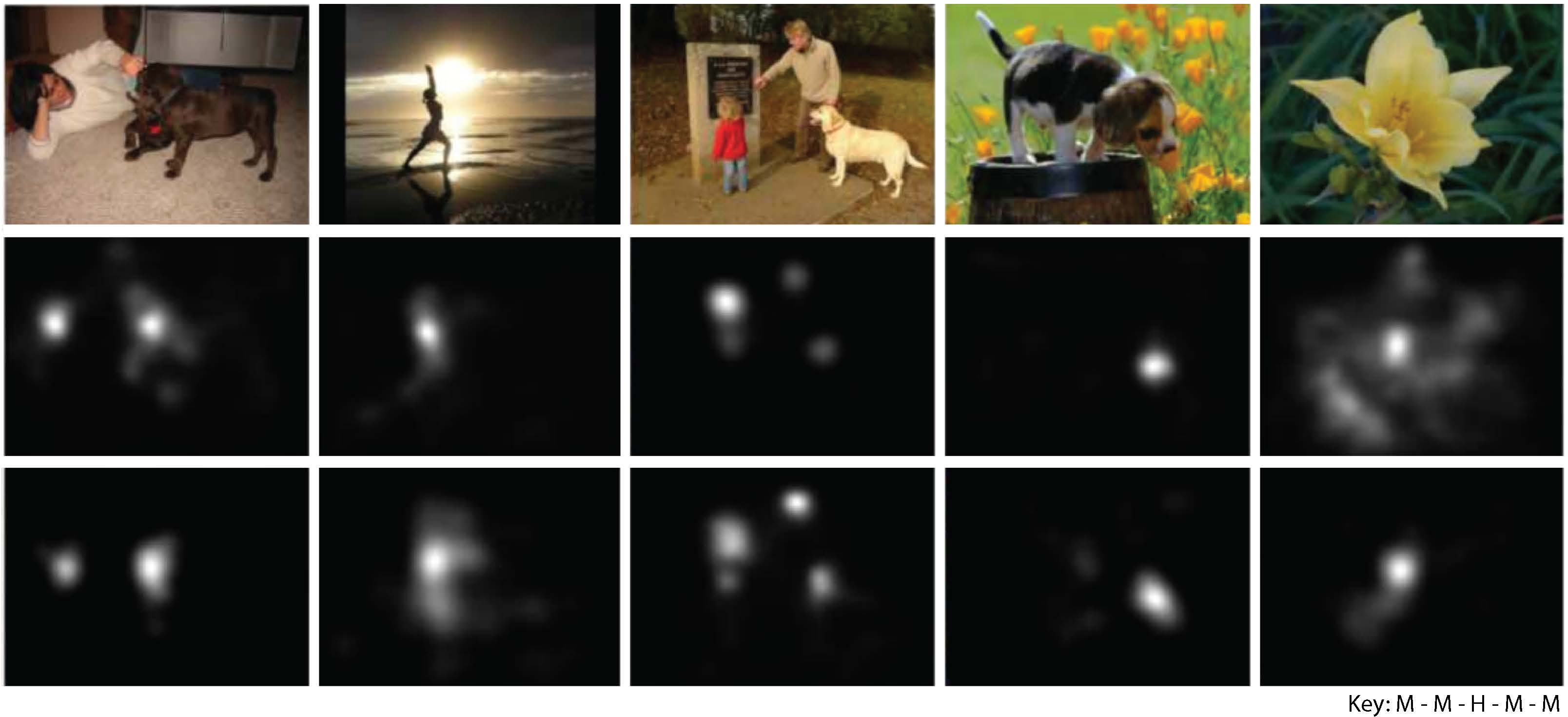}
            \vspace*{-10pt}
	\caption{Some images with fixation maps and prediction maps from the SALICON model~\cite{huang2015salicon}. Maps in the second and third rows belong to either \textit{Model (M) or Humans (H)} (\ie are complementary). Try to guess which one is which. Zoom on the bottom-right text to see the answer.} 
	\label{turing}
            \vspace*{-10pt}	
\end{figure}

Broadly speaking, the literature on attentional models can be split into two categories: task-agnostic approaches (\ie finding the salient pieces of information, \aka bottom-up (BU) saliency~\cite{treisman1980feature,Itti_etal98pami,koch1987shifts,borji2013state}) and 
task-specific methods (\ie finding information relevant to the ongoing behavior, task, or goal 
~\cite{hayhoe2005eye,Borji_etal14smc}).
Bottom-up salience is the most extensively studied aspect of visual guidance. The model by Itti \etal~\cite{Itti_etal98pami} in 1998 triggered a lot of interests in visual saliency in the computer vision community. 
Since then, there has been a lot of follow up works using better handcrafted features and learning methods. These  works have been appearing in top venues in the field on a regular basis.
After a short period of depression and performance saturation ($\sim$2010-2014) and driven by three factors: a) behavioral studies discovering cues that attract gaze (\eg gaze direction), b) large scale crowd-sourced data  (\eg mouse movements) indicating what is salient, and c) resurgence of deep neural networks (DNNs), a cascade of deep visual saliency models has emerged. These models have shown tremendous improvements over the classic models on well-established saliency benchmark datasets and perform close to the human inter-observer (human IO) model (a model built from fixations of other people) such that in some cases it is almost impossible to tell apart model predictions from actual human fixation maps. This is illustrated in Fig.~\ref{turing}.

The new NN-based models, however, still suffer from fundamental problems causing them to underperform humans~\cite{BylinskiiECCV2016}. It is thus crucial to step back and study where saliency models currently stand in their capabilities, and where improvements can to be made. Further, a systematic effort is needed to address the newly  surfaced challenges, reevaluate the current successes, and explore whether models are indeed continuously getting better or saturating in performance. To this end, I ask the following questions: 
what saliency models predict, 
where and why they fail, 
how can models be evaluated in a finer-grained way, 
what are the remaining cues that attract attention but are not accounted by models, 
to what degree explicit saliency measures agree with eye movements, 
how well saliency models generalize, 
what are the new types of data that can be used to uncover model strengths and weaknesses, 
which measures are most representatives of model performance, 
what are the new applications made possible in light of the recent advances in deep learning, 
and how can saliency benefit from or contribute to the other domains.  
Before delving into these questions, I first review the landscape of the saliency field and provide a quantitative comparison of a large number of static and dynamic saliency models over image and video datasets
using multiple scores. 
Results are compiled from our own recent conference publications and online benchmarks as well as colleagues' publications in this area.

This work updates my last review of the field in 2013~\cite{borji2013state,borji2013quantitative} and focuses on deep visual saliency models. There has been a lot of efforts and progress since then in terms of building models\footnote{It is worth noting that bottom-up saliency models have gradually shifted from locating low-level conspicuous image regions to predicting eye movements. Please refer to~\cite{Borji_etal13vr} for a discussion on this.}, collecting data, understanding the limitation of models, and designing evaluation measures. The reader is referred to~\cite{bruce2015computational, BylinskiiECCV2016,bylinskii2018different,tavakoli2017saliency,Borji_etal13iccv, koehler2014saliency, bylinskii2015towards} for critical reviews.

\subsection{Significance and Applications} The significance of visual saliency models is two folds. First, they present testable predictions that can be utilized for understanding human attention mechanisms at behavioral and neural levels. Indeed, a large number of cognitive studies have utilized saliency models for model-based hypothesis testing. Second, predicting where people look in images and videos is useful in a wide variety of applications across several domains (\eg computer vision, robotics, neuroscience, medicine, assistive systems, healthcare, human-computer interaction, and defense). Some example applications include gaze-aware compression and summarization, activity recognition, object segmentation, recognition and detection, image captioning, question answering, advertisement, novice training, patient diagnosis, and surveillance. 

\subsection{Scope} The principal focus of this work is on modeling visual saliency and attention during free viewing of natural scenes\footnote{As a default scene analysis task, the free viewing paradigm offers a wealth of insights regarding the cues that attract attention.}, and particularly on the recent progress. 
Some research areas are related but will not be covered here including \textit{top-down attention and egocentric gaze prediction}\footnote{Top-down attention regards gaze mechanisms during complex daily tasks such as coffee making, driving, and game playing.},
\textit{salient object detection and segmentation}\footnote{The goal in this class of models is to detect and segment salient objects in a scene rather than predicting where people look. Please see~\cite{borji2012salient} for a review of these models.}, \textit{co-saliency detection}\footnote{It regards locating the salient regions from multiple images.}, as well as \textit{saliency for deep learning visualization}\footnote{This line of research attempts to explain activations of a neural network by means of techniques based on backpropagation (\eg~\cite{yosinski2015understanding}). The goal is to determine where a neural network looks in classifying an object, rather than replicating human fixations.}. 

\section{Saliency in the Deep Learning Era}

\subsection{Benchmarks}

\begin{figure}[t]
	\centering
    \includegraphics[width=1\linewidth]{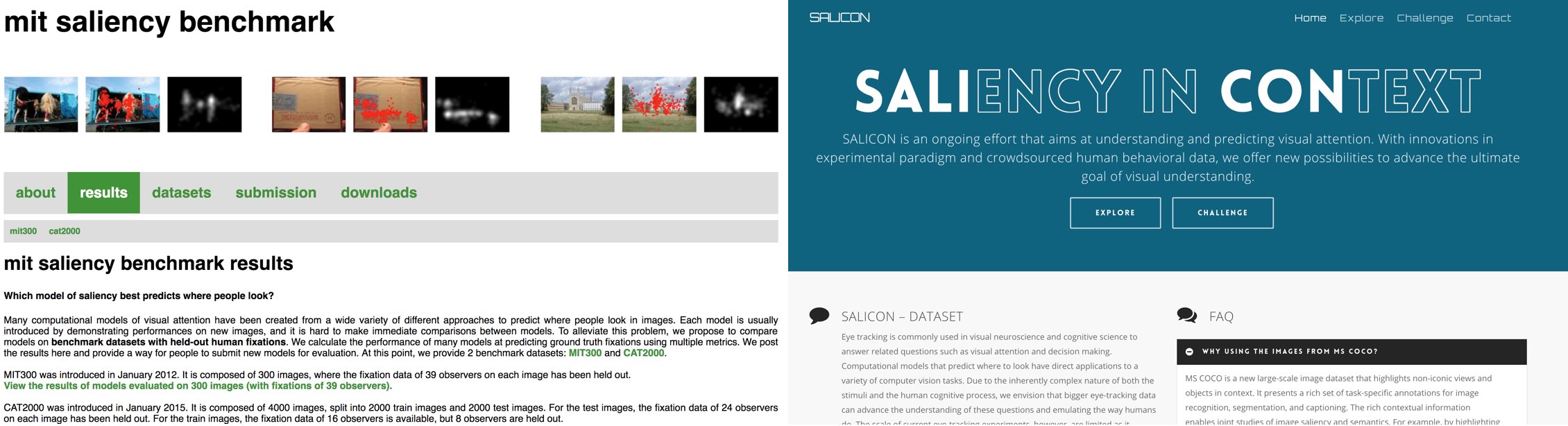} 
   	    \vspace*{-10pt} 
	\caption{Two major saliency benchmarks: \href{http://saliency.mit.edu}{MIT} (left) and \href{http://salicon.net/}{SALICON} (right). The former compares models on fixations over two datasets (MIT300 and CAT2000), whereas the latter considers mouse trajectories.}
	\label{fig:benchs}
	    \vspace*{-10pt}
\end{figure}

Benchmarks have been instrumental for advances in computer vision. In the saliency domain, they have sparked a lot of interest and have spurred a lot of interesting ideas over the past several years. Two of the most influential image-based ones\footnote{MIT: \href{saliency.mit.edu}{saliency.mit.edu} \& SALICON: \href{http://salicon.net}{http://salicon.net}} include MIT and SALICON (Fig.~\ref{fig:benchs}).

\textbf{MIT:} The MIT benchmark is currently the gold standard for evaluating
and comparing image-based saliency models. It supports eight evaluation measures for comparison and reports results over two eye movement datasets 1) MIT300 and 2) CAT2000. As of October 2018, we have evaluated 85 models over the MIT300 dataset, out of which 26 are NN-based models ($\sim$30\% of all submissions). The CAT2000 dataset has 30 models evaluated to date (9 are NN-based). In addition, 5 baselines are also computed on both datasets.

\textbf{SALICON:} Also known as the LSUN saliency challenge\footnote{Together with LSUN challenge, and including iSUN dataset~\cite{xu2015turkergaze}.}, it is relatively new and is primarily based on the SALICON dataset. It offers results over 7 scores and uses the same evaluation tools as the MIT benchmark.

These two benchmarks are complementary to each other. The former evaluates models with respect to actual fixations but suffers from small scale data. The latter mitigates the scale problem but considers noisy click data as a proxy of attention. Further, as opposed to SALICON, we maintain an up-to-date leader-board in the MIT benchmark.

\subsection{Datasets}
\begin{figure}[t]
	\centering
    \includegraphics[width=1\linewidth]{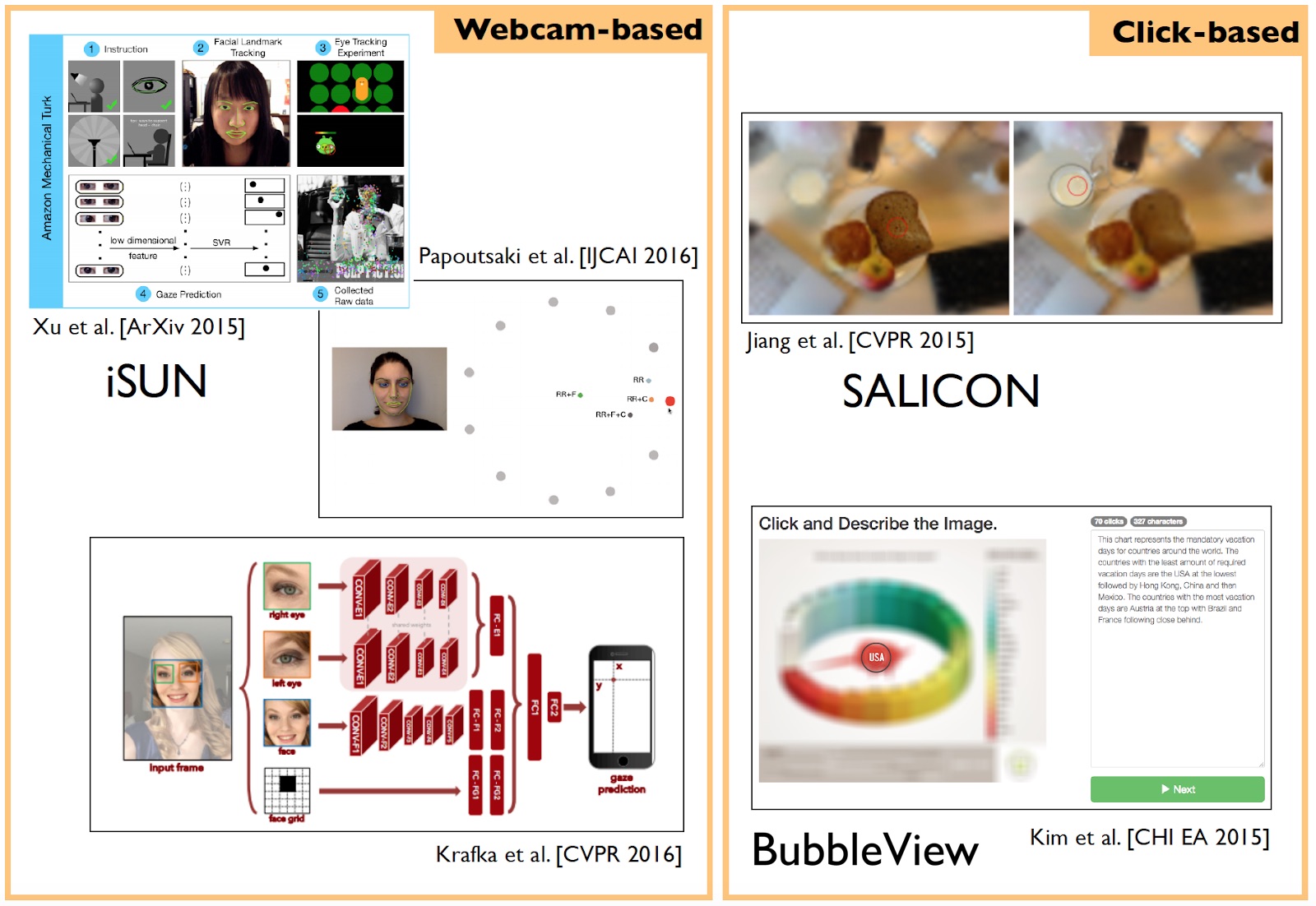} 
    		    \vspace*{-10pt}
	\caption{New crowd-sourcing methodologies to collect attention data including TurkerGaze~\cite{xu2015turkergaze}, SALICON~\cite{jiang2015salicon}, and BubbleView~\cite{kim2017bubbleview}.}
	\label{fig:header2}
	    \vspace*{-10pt}
\end{figure}

Traditionally, validation of saliency models has been done by comparing their outputs against eye movements of humans watching complex image or video stimuli (\eg~\cite{parkhurst2002modeling,bruce2005saliency}). New databases have emerged by following two trends, 1) increasing the number of images, and 2) introducing new measurements to saliency by providing contextual annotations (\eg image categories, regional properties, etc.). To annotate large scale data, researchers have resorted to crowd-sourcing schemes such as gaze tracking using webcams~\cite{xu2015turkergaze} or mouse movements~\cite{jiang2015salicon,kim2017bubbleview} as alternatives to lab-based eye trackers (Fig.~\ref{fig:header2}). Deep supervised saliency models rely heavily on these sufficiently large and well-labeled datasets. Here, I review some of the most recent and influential image and video datasets. The discussion of pros and cons of these datasets is postponed to Section~\ref{challenges}. For a review of fixation datasets pre-deep learning era please consult~\cite{winkler2013overview}.

\noindent {\bf Image datasets.} Three of the most popular image datasets used for training and testing models are as follows. 


$\bullet$ \textbf{MIT300:} This dataset is a collection of 300 natural images from the Flickr Creative Commons and personal
collections~\cite{mit-saliency-benchmark}. It contains eye movement data of 39 observers which results in a fairly robust ground-truth to test models against. It is a challenging dataset for saliency models, as images are highly
varied and natural. Fixation maps of all images are held out and used by the MIT Saliency Benchmark for evaluating models.

$\bullet$ \textbf{CAT2000:} Released in 2015, this is a relatively larger dataset consisting of 2000 training images and 2000 test images spanning 20 categories such as Cartoons, Art, Satellite, Low resolution images, Indoor, Outdoor, Line drawings, etc.~\cite{mit-saliency-benchmark}. Images in this dataset come from search engines and computer vision
datasets. The training set contains 100 images per category and has fixation annotations from 18 observers. The test test, used for evaluation, contains the fixations of 24 observers. Both MIT300 and CAT2000 datasets are collected using the EyeLink1000 eye-tracker.

$\bullet$ \textbf{SALICON:} Introduced by Jiang~\etal~\cite{jiang2015salicon}, it is currently the largest crowd-sourced saliency dataset. Images in this dataset come from the Microsoft COCO dataset and contain MS COCO's pixelwise
semantic annotations. The SALICON dataset contains 10000 training images, 5000 validation images and 5000 test images. 
Mouse movements are collected using Amazon Mechanical Turk (AMT) via a psychophysical paradigm known as \textit{mouse-contingent saliency annotation} (Fig.~\ref{fig:header2}). Despite minor discrepancies between eye movements and mouse movements, nevertheless this dataset introduces an acceptable and scalable method for the collection of further data for saliency modeling. Currently, many deep saliency models are first trained on the SALICON dataset and are then finetuned on the MIT1000~\cite{judd2009learning} or CAT2000 datasets for predicting fixations.

\begin{figure*}[t]
	\centering
    \includegraphics[width=.9\linewidth ]{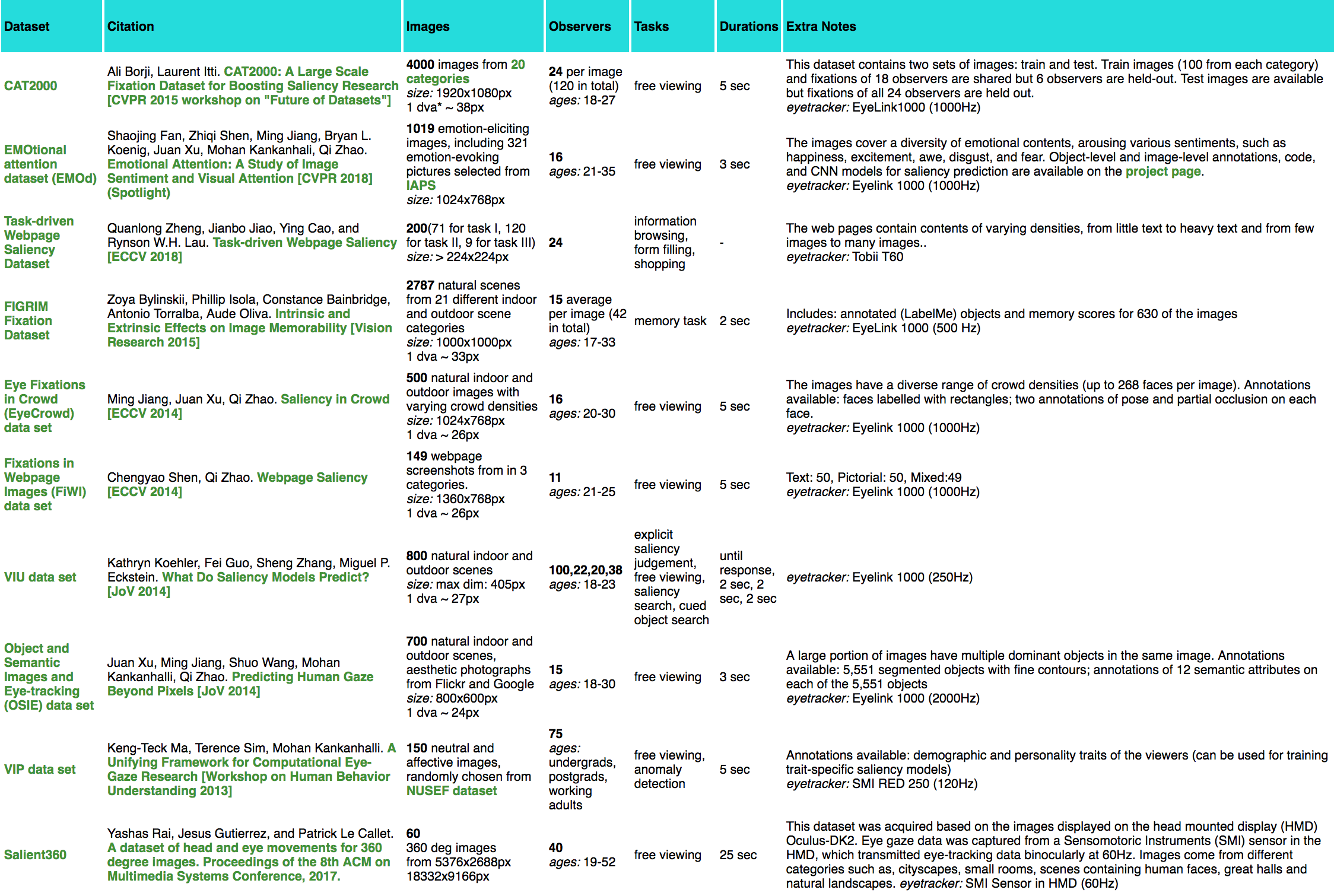} \\
    \vspace*{-10pt}
	\caption{Recent fixation datasets over still images. Please refer to the MIT Saliency Benchmark and the text for details.}
	\label{fig:dbs}
    \vspace*{-5pt}	
\end{figure*}

Fig.~\ref{fig:dbs} shows a list of recent image datasets. Some datasets have been collected to study visual attention over specific stimuli (\eg viewing infographics~\cite{bylinskii2017understanding}, crowds~\cite{jiang2014saliency}, webpages~\cite{shen2014webpage,zheng2018task}, emotional content~\cite{ramanathan2010eye,fan2018emotional}), or certain tasks (\eg driving~\cite{borji2012probabilistic,palazzi2016should}).

\noindent {\bf Video datasets.} 
DIEM~\cite{mital2011clustering}, HOLLYWOOD-2~\cite{mathe2015actions}, and UCF-Sports~\cite{mathe2015actions} are the three widely used datasets for video saliency research. 
DIEM contains 84 videos from various categories (\eg advertisements, documentaries, sport
events, and movie trailers; ranging in duration from 20 to more than 200 seconds) and free-viewing fixation data from 50 subjects. 
HOLLYWOOD-2 is the largest dynamic eye tracking dataset containing 823 training and 884 validation sequences, with fixation data of 16 subjects. The videos in this dataset are short video sequences from a set of 69 Hollywood movies, containing 12 different human action classes, ranging from answering phone, eating, driving, running, etcetera. UCF-Sports dataset contains 150 videos on 9 sports action classes such as diving,
swinging, and walking, with an average duration of 6.39 seconds. It is watched by 16 subjects.

Recently, two newcomers have joined the league: 

$\bullet$ \textbf{DHF1K:} In~\cite{wang2018revisiting}, we introduced a dataset comprising a total of 1,000 video sequences viewed by 17 subjects. Videos contain 582,605 frames covering a wide range of scenes, motions, and activities. It has the total duration of 19,420 seconds. The dynamic stimuli were displayed on a 19 inch display (resolution 1440 x 900px). 

$\bullet$ \textbf{LEDOV:} Introduced by Jiang~\etal~\cite{jiang2018deepvs}, this dataset includes diverse content spanning daily actions, sports, social activities, and art performance. It contains 179,336 frames, 5,058,178 fixations, and 6,431 seconds. All 538 videos have at least 720px resolution and 24 Hz frame rate.



\begin{figure*}
	\centering
    \includegraphics[width=.9\linewidth ]{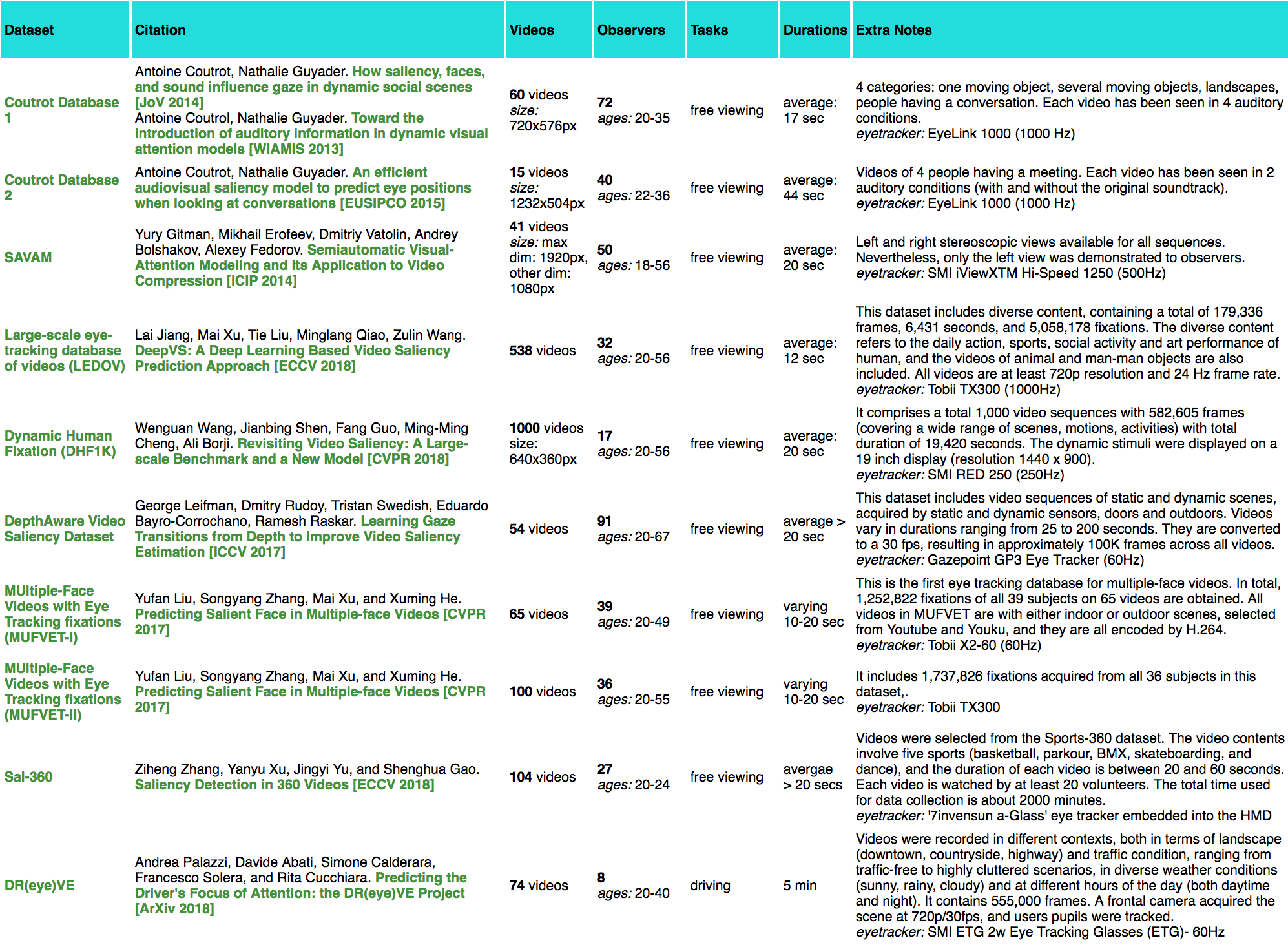} \\
    \vspace*{-10pt}
	\caption{Recent fixation datasets over videos. Please refer to the MIT Saliency Benchmark and the text for details.}
		\label{fig:viddbs}
    \vspace*{-5pt}	
\end{figure*}

Fig.~\ref{fig:viddbs} shows a list of recent video saliency datasets. Some datasets have been collected for special purposes such as studying multi-modal attention (\eg Coutrot 1 \& 2~\cite{coutrot2014saliency}), exploring attention on videos with multiple faces (MUFVET I \& II~\cite{liu2017predicting}), or modeling 3D saliency (\eg Sal-$360^\circ$~\cite{zhang2018saliency,assens2017saltinet}).

\subsection{Evaluation Measures}

Several measures have been proposed for evaluating saliency models. In general, they fall into two categories: a) \textit{distribution-based} (\ie comparing smoothed prediction and fixation maps) and b) \textit{location-based} (\ie computing some statistics at fixated locations). Pearson's Correlation Coefficient (CC), 
Kullback-Leibler divergence (KL), 
Earth Mover's Distance (EMD)~\cite{rubner2000earth}, 
and 
Similarity or histogram intersection (SIM)~\cite{rubner2000earth} fall under the first category. Normalized Scanpath
Saliency (NSS)~\cite{Peters_etal05vr}, 
Area under ROC Curve (AUC) and its variants including AUC-Judd~\cite{judd2009learning}, AUC-Borji~\cite{bylinskii2014saliency}, and Shuffled AUC (sAUC)~\cite{zhang2008sun} fall under the second category. Recently, Information Gain (IG) has also been added to this set~\cite{kummerer2015information}. It 
measures the average information gain of the saliency map over the center prior baseline at fixated locations~\cite{bylinskii2018different}. Rankings produced by these scores often do not match with each other. Nonetheless, it is commonly agreed that these measures are complementary to each other and assess different aspects of saliency maps. Further, it is believed that a good model should perform well over a variety of measures. For a detailed description on how these measures are computed, please consult~\cite{mit-saliency-benchmark,bylinskii2018different}. I will return to model evaluation and scoring later in Section~\ref{challenges}.

The MIT benchmark ranks models based on the AUC-Judd, whereas in SALICON ranking is based on the sAUC. 
The MIT benchmark will soon switch to NSS, suggested by the recent research~\cite{bylinskii2018different}, and the fact that it favors maps that visually better match with human fixation map (see~\cite{xia2018learning}). 


\subsection{Classic (Non-deep) Bottom-up Saliency Models}
Computational modeling of bottom-up attention dates back to the seminal works by Treisman and Gelade in 1980~\cite{treisman1980feature}, the computational architecture by Koch and Ullman in 1985~\cite{Koch_Ullman85}, and the bottom-up model of Itti \etal in 1998~\cite{Itti_etal98pami}.
Rooted in these works, several saliency models have been proposed. They fall into different categories (\eg Bayesian, learning-based, spectral, cognitive), according to our study in 2013~\cite{borji2013state}. 
The early models mainly relied on extracting simple feature maps such as intensity, color, and orientation in a scale space and combined them after performing center-surround and normalization operations. Subsequent models incorporated mid- and higher level features (\eg face and text~\cite{2009Cerf}, gaze direction~\cite{Parks_etal15vr}) to better predict gaze. 

A thorough examination of all classic saliency models is certainly not feasible in this limited space. Instead, I list a number of highly influential static and dynamic saliency models as follows: 
Attention for Information Maximization (AIM)~\cite{bruce2005saliency}, Graph-based Visual Saliency (GBVS)~\cite{harel2006graph}, Saliency Using Natural statistics (SUN)~\cite{zhang2008sun}, Spectral Residual saliency (SR)~\cite{hou2007saliency}, Adaptive Whitening Saliency (AWS)~\cite{garcia2012saliency}, Boolean Map based Saliency (BMS)~\cite{zhang2013saliency}, and the Judd model~\cite{judd2009learning}. Some classic video saliency models include:
AWS-D~\cite{leboran2017dynamic}, OBDL~\cite{hossein2015many}, Xu \etal~\cite{xu2017learning}, PQFT~\cite{guo2008spatio}, and Rudoy \etal~\cite{rudoy2013learning}. 

\subsection{Deep Saliency Models}

The success of convolutional neural networks (CNN)~\cite{LeCun_Cortes} on large scale object recognition corpora~\cite{deng2009imagenet}, has brought along a new wave of saliency models that perform markedly better than 
traditional saliency models based on hand-crafted features. 
Researchers leverage existing CNNs that are trained for scene recognition and re-purpose them to predict saliency. Often some architectural novelties are also introduced. These models are trained in an end-to-end manner, effectively formulating saliency as a regression problem (See Fig.~\ref{fig:timeline}.A). To remedy the lack of sufficiently large scale fixation datasets, deep saliency models are pre-trained on large image datasets and are then fine-tuned on small scale eye movement or click datasets. This procedure allows models to re-use the semantic visual knowledge already learned in CNNs and successfully transfer it to the task of saliency. In what follows, I review some of the landmark deep spatial and spatiotemporal saliency models.

\subsubsection{Static Saliency Models}

\begin{figure*}[t]
	\centering
    \includegraphics[width=1\linewidth]{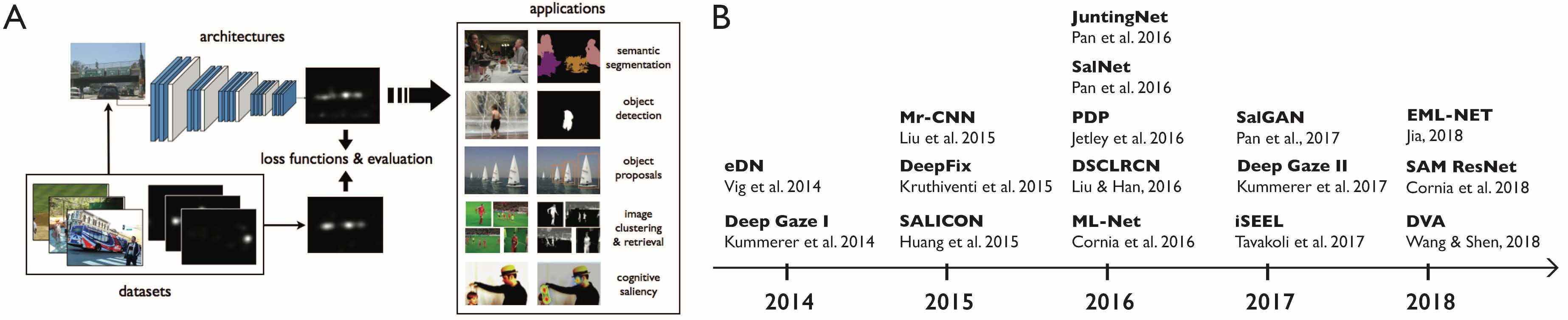} \\
	\caption{A) Common structure of deep saliency models. Pre-trained models for object recognition are adopted and fine-tuned for saliency recognition, and are applied to different tasks, B) A timeline of deep-learning based visual saliency since 2014.} 
	\label{fig:timeline}
	       \vspace*{-10pt}
\end{figure*}

Fig.~\ref{fig:timeline}.B presents a timeline of deep saliency models since their inception in 2014. It forms the basis according to which models are reviewed here.

\textbf{eDN:} Put forth by Vig \etal~\cite{vig2014large}, the eDN (ensembles of deep networks) was the first attempt to leverage CNNs for predicting image saliency. First, it generates a large number of richly-parameterized 1 to 3 layer networks using biology-inspired hierarchical features. Then, it uses hyperparameter optimization to search for independent models that are predictive of saliency and combines them into a single model by training a linear SVM.

\textbf{DeepGaze I \& II:}  K{\"u}mmerer \etal~\cite{kummerer2014deep} 
proposed to employ a relatively deeper CNN (pre-trained AlexNet~\cite{krizhevsky2012imagenet}) for saliency prediction (5 layers compared to eDN's 1-3 layers). The outputs of the convolutional layers were used to create and train a linear model to compute image salience. More recently, they introduced the DeepGaze II model~\cite{kummerer2017understanding} 
built upon DeepGaze I. It further explores the unique contributions between
low-level and high-level features towards fixation prediction and obtains the best performance on the MIT300 dataset in terms of the AUC-Judd score.

\textbf{Mr-CNN:} Liu \etal~\cite{liu2015predicting} proposed the multi-resolution CNN (Mr-CNN) model in which three different CNNs, each trained on a different scale, are followed by two fully connected layers. They fine-tuned their model over image patches centered at fixated and non-fixated locations.

\textbf{SALICON:} To help bridge the semantic gap (inability of models in predicting
gaze driven by strong semantic content) in saliency modeling, Huang \etal~\cite{huang2015salicon} proposed a deep CNN (based on AlexNet, VGG-16, and GoogLeNet) that combines information from two pre-trained CNNs, each on a different image scale (fine and coarse). The two CNNs are then concatenated to produce the final map.

\textbf{DeepFix:} This model, by Kruthiventi \etal~\cite{kruthiventi2017deepfix}, is the first application of fully convolutional neural networks (FCNN) for saliency prediction. 
It employs 5 convolution blocks with weights initialized from the VGG-16 which are followed by 
two of their novel Location Based Convolutional (LBC) layers to capture semantics at multiple scales. It also incorporates Gaussian priors to further improve learned weights.

\textbf{ML-Net:} 
Instead of using features at the final CNN layers, this model~\cite{cornia2016deep} combines feature maps extracted from different levels of the VGG network to compute saliency.
  To model center prior, it learns a set of Gaussian parameters end-to-end, as opposed to using a fixed Gaussian or feeding Gaussians to convolutional layers. Further, it benefits from a new loss function to satisfy three objectives: 1) to better measure similarity with the ground truth, 2) to make prediction maps invariant to their maximum, and 3) to give higher weights to locations with higher fixation probability.

\textbf{JuntingNet and SalNet:} Pan \etal~\cite{pan2016shallow} proposed a shallow CNN (JuntingNet) and a deep CNN (SalNet) for saliency prediction. The former is inspired by
the AlexNet and uses three convolutional and two fully connected layers, which are all randomly initialized (\ie trained from scratch). The latter, contains eight convolutional layers with the first three being initialized from the VGG network.

\textbf{PDP:} The Probability Distribution
Prediction (PDP), proposed by Jetley \etal~\cite{jetley2016end}, formulates the saliency as a generalized
Bernoulli distribution and trains a model to learn this distribution. It contains a deep neural network trained completely end-to-end using a novel loss function that pairs the classic softmax loss with functions that compute the distances between different probability distributions. According to Jetley \etal's results, the new loss functions are more efficient than traditional loss functions such as Euclidean and Huber loss for the task of saliency prediction.

\textbf{DSCLRCN:} The DSCLRCN model (Deep Spatial Contextual Long-term Recurrent Convolutional Neural network) was introduced by Liu and Han~\cite{liu2016deep} in 2016. It first learns local saliency of small image regions using a CNN. Then, it scans the image both horizontally and vertically using a deep spatial long short-term model (LSTM) to 
capture global context. These two operations allow DSCLRCN to simultaneously and effectively incorporate local and global contexts to infer image saliency.

\textbf{SalGAN:} Pan \etal~\cite{pan2017salgan} utilized Generative Adversarial Networks (GANs)~\cite{goodfellow2014generative} to build the SalGAN model. It is
composed of two modules, a generator and a discriminator. The generator is learned via back-propagation using binary cross entropy loss on existing saliency maps, which is then passed to the discriminator that is trained to identify whether the provided saliency map was synthesized by the generator, or built from the actual fixations.

\textbf{iSEEL:} In~\cite{tavakoli2017exploiting}, we proposed a saliency model 
based on inter-image similarities and ensemble of Extreme Learning Machines
(ELM)~\cite{huang2006extreme}. First, a set of images similar to a given image are retrieved. A saliency predictor is then learned from each image in this set using an ELM, resulting in an ensemble. Finally, the saliency maps by the ensemble's members are averaged to construct the final map.

\textbf{EML-Net:} The main idea in Expandable Multi-Layer NETwork (EML-NET) is that the encoder and decoder in a FCNN can be separately trained for scalability~\cite{jia2018eml}. Furthermore, the encoder can contain more than one CNN model to extract features, and the models can have different architectures or be pre-trained on different datasets.
After feeding an image to the decoder, filter maps from two different CNNs are computed and combined to form the final saliency maps.

\textbf{DVA:} Wang \etal~\cite{wang2017deep} proposed the Deep Visual Attention (DVA) model in which an encoder-decoder architecture is trained over multiple scales to predict pixel-wise saliency. The encoder part is a stack of convolutional layers. Three decoders are formed by taking inputs from three different stages of the encoder network which are finally fused to generate a saliency map.

\textbf{SAM Nets:} Saliency Attentive Models (SAM-ResNet \& SAM-VGG) by Cornia~\etal~\cite{cornia2016predicting} combine a fully convolutional network with a recurrent convolutional network, endowed with a spatial attentive mechanism.

\textbf{FUCOS:} Bruce \etal~\cite{bruce2016deeper} present a model deemed Fully Convolutional Saliency (FUCOS), that is applied to either gaze, or salient object prediction. It integrates pre-trained layers from large-scale
CNN models and is then fine-tuned on the PASCAL-Context dataset~\cite{mottaghi2014role}.

\textbf{Attentional Push:} Gorji and Clark~\cite{gorji2017attentional} proposed to augment standard deep saliency models with shared attention. 
Their model contains two pathways: an Attentional Push pathway which learns the gaze location of the scene actors and a saliency pathway. These are followed by a shallow augmented saliency CNN which combines them and generates the augmented saliency. 
They initialized and train the Attentional Push CNN by minimizing the classification error of following the actors' gaze location on a 2D grid using a large-scale gaze-following dataset. The
Attentional Push CNN is then fine-tuned along with the augmented saliency CNN to minimize the Euclidean distance between the augmented saliency and ground truth fixations using an eye-tracking dataset, annotated with the head and
the gaze location of the scene actors. This model performs well but
suffers from the limitation that it requires the human annotations for the actors' head location during evaluation. The reason is that in many of the images, the actors are looking sideways or looking away from the camera, which makes face detection a very challenging task, even for the best-performing face detectors. 

Some other deep static saliency models include~\cite{sun2017integrated,dodge2018visual,Zhao_etal15cvpr,mahdi2017deepfeat,xu2017personalized}. My focus in this section was on models that predict fixation density maps. Deep learning has also been exploited to build scanpath prediction models (\eg~\cite{assens2018pathgan,wloka2018active}).

\subsubsection{Dynamic Saliency Models}

\begin{figure}
	\centering
    \includegraphics[width=1\linewidth]{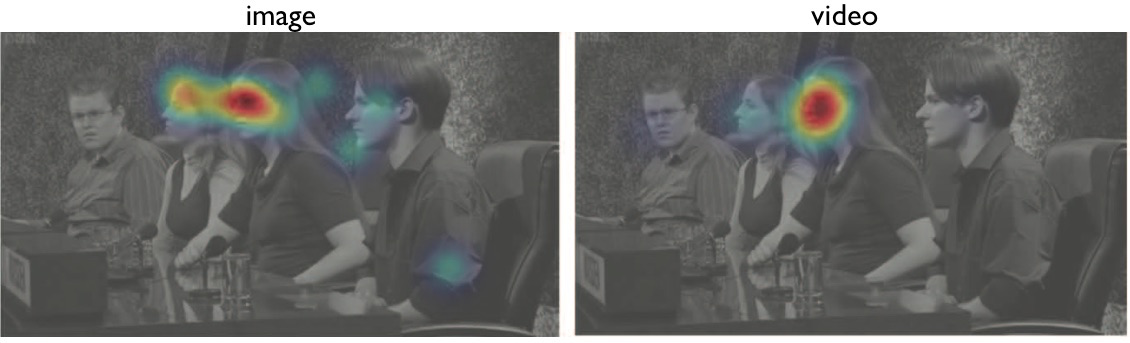} 
	       \vspace*{-10pt}    
	\caption{A) Image vs. video saliency. Rudoy \etal~\cite{rudoy2013learning} showed the same image to observers twice: once in isolation for 3 seconds (left) and once embedded in a video (right). As it can be seen, video fixation map is more concentrated on a single object whereas image saliency is dispersed. See also~\cite{nguyen2013static}. Figure from~\cite{rudoy2013learning}.}
	\label{fig:gorjiModels}
       \vspace*{-15pt}
\end{figure}

Observers look at videos and images differently due to two reasons (Fig.~\ref{fig:gorjiModels}.A). First, observers have much less time to view each video frame (about 1/30 of a second) compared to 3 to 5 seconds over still images. Second, motion is a key component that is missing in still images and strongly attracts human attention over videos.
These, plus the need for high computational and memory requirements, makes video saliency prediction a considerably more challenging task than image saliency. 
Despite these impediments, there has been a growing interest in video saliency over the past few years partially driven by its applications (\eg video summarization, image and video captioning).

Conventionally, video saliency models pair bottom-up feature extraction
with an ad-hoc motion estimation that can be performed either by means of optical flow or feature tracking. In contrast, deep video saliency models learn the whole process end-to-end. In these works, the dynamic characteristics are modeled in two ways: a) adding temporal information to CNNs, or 2) developing a dynamic
structure using LSTMs. Both types of models are covered next.  

\textbf{Two-stream network:} As one of the first attempts, Bak~\etal~\cite{bak2018spatio} applied a two-stream (5 layer each) CNN architecture for video saliency prediction. RGB frames and motion maps were fed to the two streams.

\textbf{Chaabouni \etal:} Chaabouni \etal~\cite{chaabouni2016transfer} employed transfer learning to adapt a previously trained deep network for saliency prediction in natural videos. 
They trained a 5-layer CNN on RGB color planes and residual motion for each video frame. However, their model uses only the very short-term temporal relations of two consecutive frames.

\textbf{Bazzani \etal:} In~\cite{bazzani2016recurrent}, a recurrent mixture density network is proposed for saliency prediction. The input clip of 16 frames is fed to a 3D CNN, whose output becomes the input to a LSTM. Finally, a linear layer projects the LSTM representation to a Gaussian mixture model, which describes the saliency map.
In a similar vein,
Liu \etal~\cite{liu2017predicting} applied LSTMs to predict video saliency maps, relying on
both short- and long-term memory of attention deployment.

\textbf{OM-CNN:} Proposed by Jiang \etal~\cite{jiang2017predicting}, the Object-to-Motion CNN model includes two  subnets of objectness and motion that are trained end-to-end. 
The objectness and object motion information are used to predict the intra-frame
saliency of videos. Inter-frame saliency is computed by means of a structure-sensitive ConvLSTM architecture. 

\textbf{Leifman \etal:} In~\cite{leifman2017learning}, RGB color planes, dense optical flow map, depth map and the previous saliency map are fed to a 7-layered encoder-decoder structure to predict fixations of observers who
viewed RGBD videos on a 2D screen.

\textbf{Gorji \& Clark:} 
As in their previous work~\cite{gorji2018going}, here they used a multi-stream ConvLSTM to augment state-of-the-art static saliency models with dynamic attentional push (shared attention). Their network contains a saliency pathway and three push pathways including gaze following, rapid scene changes, and attentional bounce. The multi-pathway structure is followed by a CNN that learns to combine the complementary and time-varying outputs of the CNN-LSTMs by minimizing the relative entropy between the augmented saliency and viewers fixations on videos. 

\textbf{ACLNet:} In~\cite{wang2018revisiting}, we proposed the Attentive CNN-LSTM Network which augments a CNN-LSTM with a supervised attention mechanism to enable fast end-to-end saliency learning. The attention mechanism explicitly encodes static saliency information allowing LSTM to focus on learning a more flexible temporal saliency representation across successive frames. Such a design fully leverages existing large-scale static fixation datasets, avoids overfitting, and significantly improves training efficiency. 

\textbf{SG-FCN:} Sun~\etal~\cite{sun2018sg} proposed a robust deep model that utilizes 
memory and motion information to capture salient points across successive frames. The memory information was exploited to enhance the model generalization by considering the fact that changes between two adjacent frames are limited within a certain range, and hence the corresponding fixations should remain correlated.

\subsection{Deep vs. Classic Saliency Models} 
Consider a CNN with a single convolutional layer followed by a fully connected layer trained to predict fixations. This model generalizes the classic Itti model and also models built upon it that learn to combine feature maps (\eg~\cite{Judd2009,borji2012boosting,xu2014predicting}). The learned features in the CNN will correspond to orientation, color, intensity, etc. which can be combined linearly by the fully connected layer (or 1x1 convolutions in a fully convolutional network). To handle the scale dependency of saliency computation, classic models often recruit multiple image resolutions. In addition to this technique (as in SALICON), deep saliency models concatenate maps from several convolutional layers (as in ML-Net), or feed input from different encoder layers to the decoder (\eg using skip connections) to preserve fine details. 

Despite the above resemblances, the most evident shortcoming of classic models (\eg the Itti model) with respect to today's deep architectures is the lack of ability to extract higher level features, objects, or parts of objects. Some classic models remedied this shortcoming by explicitly incorporating object detectors such as face or text detectors.
The hierarchical deep structure of CNNs (\eg 152 layers in the ResNet~\cite{he2016deep}) allows capturing complex cues that attract gaze automatically. This is perhaps the main reason behind the big performance gap between the two types of models. In practice, however, research has shown that there are cases where classic saliency models win over the deep models indicating that current deep models still fall short in fully explaining low-level saliency (Fig.~\ref{fig:bruce}). Further, deep models fail in capturing some high-level attention cues (\eg gaze direction). I will discuss these further in Section~\ref{challenges}.

\vspace*{-5pt}
\section{State of the Art Performance}
\label{sta}
A quantitative comparison of static and dynamic saliency models is presented here over two image benchmarks (MIT~\cite{bylinskii2014saliency} and SALICON~\cite{huang2015salicon}), as well as two video datasets (DHF1K~\cite{jiang2018deepvs} and LEDOV~\cite{jiang2018deepvs}). For the MIT and DHF1K benchmarks, I compile the results from our online benchmark portals, whereas for the other two I resort to publications by other researchers in the area.



\vspace*{-10pt}
\subsection{Performance on Images}

\subsubsection{The MIT Benchmark} This benchmark has the most comprehensive set of traditional and deep saliency models evaluated. Eight scores are computed. 
In addition to models, 
the following 5 baselines are also considered.


\noindent {\bf 1) Infinite humans:} How well a fixation map of infinite observers predicts fixations from a different set of infinite observers, computed as a limit? Two groups of $n$ observers are considered. Fixations of the first group are used to predict the fixations of the second group. By varying $n$ and fitting a power function to the points, an empirical limit is computed. Please see~\cite{bylinskii2018different} for details. 

\noindent  {2) \bf One human:} How well a fixation map of one observer (taken as a saliency map) predicts the fixations of the other $n-1$ observers. This is computed for each observer in turn, and averaged over all $n$ observers. Different individuals are more or less predictive of the rest of the population, and so a range of prediction scores is obtained. 


\noindent  {3) \bf Center:} A symmetric Gaussian is stretched to fit the aspect ratio of a given image, under the assumption that the center of the image is most salient~\cite{tatler2007central}.

\noindent  {4) \bf Permutation control:} For each image, instead of randomly sampling fixations, fixations from a randomly-sampled image are chosen as the saliency map. This process is repeated 5 times per image, and the average  performance is computed. This method allows capturing observer and center biases that are independent of the image 
\cite{koehler2014saliency}.

\noindent  {5) \bf Chance:} A random uniform value is assigned to each image pixel to build a saliency map. Average performance is computed over 5 such chance saliency maps per image. 


\begin{table}
\caption{Performance of saliency models over the MIT300 dataset. Models are sorted based on the AUC-Judd score. The best score is shown in bold and is underlined.  
See~\cite{BylinskiiECCV2016} for results using the IG score.}
\vspace{-10pt}

\centering
\renewcommand{\tabcolsep}{.8mm}
\renewcommand{\arraystretch}{.85}%
\begin{scriptsize}
\begin{tabular}{ | l | cccccccc | } 
\hline
& AUC-J & SIM & EMD $\downarrow$ & AUC-B & sAUC & CC & NSS & KL $\downarrow$ \\ \hline
\hline

Baseline: infinite h.	&  0.92	& 1	& 0	& 0.88	& 0.81	& 1	& 3.29& 	0 \\	
\hline
Deep Gaze 2	&     \underline{{\bf 0.88}}	& 0.46	& 3.98	& \underline{{\bf 0.86}}	& 0.72	& 0.52	& 1.29	& 0.96	\\
EML-NET	& \underline{{\bf 0.88}}	& 0.68	& \underline{{\bf 1.84}}	& 0.77	& 0.7	& 0.79	& \underline{{\bf 2.47}}	& 0.84	\\
SALICON	& 0.87	& 0.6	& 2.62	& 0.85	& \underline{{\bf 0.74}}	& 0.74	& 2.12	& 0.54	\\
DeepFix	& 0.87	& 0.67	& 2.04	& 0.8	& 0.71	& 0.78	& 2.26	& 0.63	\\
DSCLRCN	& 0.87	& 0.68	& 2.17	& 0.79	& 0.72	& 0.8	& 2.35	& 0.95	\\
SAM-ResNet	& 0.87	& 0.68	& 2.15	& 0.78	& 0.7	& 0.78	& 2.34	& 1.27 \\	
SAM-VGG	& 0.87	& 0.67	& 2.14	& 0.78	& 0.71	& 0.77	& 2.3	& 1.13	\\
DenseSal	& 0.87	& 0.67	& 1.99	& 0.81	& 0.72	& 0.79	& 2.25	& \underline{{\bf 0.48}} \\	
DPNSal	& 0.87	& \underline{{\bf 0.69}}	& 2.05	& 0.8	& \underline{{\bf 0.74}}	& \underline{{\bf 0.82}}	& 2.41	& 0.91	\\
CEDNS	& 0.87	& 0.64	& 2.23	& 0.74	& 0.69	& 0.75	& 2.43	& 0.63 \\	
SalGAN	& 0.86	& 0.63	& 2.29	& 0.81	& 0.72	& 0.73	& 2.04	& 1.07 \\	
LRL2	& 0.86	& 0.57	& 2.57	& 0.84	& 0.7	& 0.61	& 1.57	& 0.73	\\
PDP	& 0.85	& 0.6	& 2.58	& 0.8	& 0.73	& 0.7	& 2.05	& 0.92	\\
ML-Net	& 0.85	& 0.59	& 2.63	& 0.75	& 0.7	& 0.67	& 2.05	& 1.1	\\
DVA	& 0.85	& 0.58	& 3.06	& 0.78	& 0.71	& 0.68	& 1.98	& 0.64	\\
DeepPeek	& 0.85	& 0.63	& 2.37	& 0.81	& 0.7	& 0.71	& 1.98	& 1.15	\\
Deep Gaze 1	& 0.84	& 0.39	& 4.97	& 0.83	& 0.66	& 0.48	& 1.22	& 1.23\\	
iSEEL	& 0.84	& 0.57	& 2.72	& 0.81	& 0.68	& 0.65	& 1.78	& 0.65\\	
IML-AELSAL	& 0.84	& 0.54	& 3.08	& 0.82	& 0.69	& 0.63	& 1.7	& 0.71\\	
BMS	& 0.83	& 0.51	& 3.35	& 0.82	& 0.65	& 0.55	& 1.41	& 0.81	\\
SalNet	& 0.83	& 0.52	& 3.31	& 0.82	& 0.69	& 0.58	& 1.51	& 0.81\\	
Mixture of Models	& 0.82	& 0.44	& 4.22	& 0.81	& 0.62	& 0.52	& 1.34	& 0.91	\\
eDN	& 0.82	& 0.41	& 4.56	& 0.81	& 0.62	& 0.45	& 1.14	& 1.14	\\
OS	& 0.82	& 0.51	& 3.35	& 0.81	& 0.64	& 0.54	& 1.41	& 0.84	\\
Judd	& 0.81	& 0.42	& 4.45	& 0.8	& 0.6	& 0.47	& 1.18	& 1.12	\\
CovSal	& 0.81	& 0.47	& 3.39	& 0.67	& 0.57	& 0.45	& 1.22	& 2.68	\\
GBVS	& 0.81	& 0.48	& 3.51	& 0.8	& 0.63	& 0.48	& 1.24	& 0.87	\\
SWD	& 0.81	& 0.46	& 3.89	& 0.8	& 0.59	& 0.49	& 1.27	& 0.97	\\
RARE2012-Imp.	& 0.81	& 0.48	& 3.74	& 0.8	& 0.66	& 0.51	& 1.34	& 0.89 \\	
LDS	& 0.81	& 0.52	& 3.06	& 0.76	& 0.6	& 0.52	& 1.36	& 1.05	\\
GoogLeNetCAM	& 0.81	& 0.45	& 4.04	& 0.8	& 0.67	& 0.49	& 1.26	& 0.99	\\
Cross Pixels Sal	& 0.81	& 0.5	& 3.36	& 0.79	& 0.58	& 0.51	& 1.31	& 0.88	\\
SMSPM	& 0.81	& 0.5	& 3.36	& 0.79	& 0.58	& 0.51	& 1.31	& 0.88	\\
FENG-GUI (FG)	& 0.81	& 0.41	& 4.63	& 0.79	& 0.64	& 0.46	& 1.25	& 1.13	\\
FES	& 0.8	& 0.49	& 3.36	& 0.73	& 0.59	& 0.48	& 1.27	& 1.2	\\
\hline
Baseline: 1 human	& 0.8	& 0.38	& 3.48	& 0.66	& 0.63	& 0.52	& 1.65	& 6.19	\\
\hline
COHS	& 0.8	& 0.5	& 3.49	& 0.76	& 0.62	& 0.49	& 1.27	& 1.36	\\
KalSal	& 0.8	& 0.44	& 4.18	& 0.79	& 0.64	& 0.46	& 1.18	& 1.03	\\
JuntingNet	& 0.8	& 0.46	& 4.06	& 0.79	& 0.64	& 0.54	& 1.43	& 0.96	\\
CVPR-1843	& 0.8	& 0.45	& 3.99	& 0.79	& 0.63	& 0.56	& 1.47	& 0.95	\\
WEPSAM	& 0.8	& 0.45	& 4.22	& 0.78	& 0.62	& 0.51	& 1.36	& 1	\\
UHF	& 0.8	& 0.45	& 4.11	& 0.79	& 0.64	& 0.47	& 1.21	 & 1	\\
STC	& 0.79	& 0.39	& 4.79	& 0.78	& 0.54	& 0.4	& 0.97	& 1.23	\\
RC	& 0.79	& 0.48	& 3.48	& 0.78	& 0.55	& 0.47	& 1.18	& 0.93	\\
CWS	& 0.79	& 0.46	& 3.81	& 0.78	& 0.55	& 0.45	& 1.11	& 0.99	\\
Mr-CNN	& 0.79	& 0.48	& 3.71	& 0.75	& 0.69	& 0.48	& 1.37	& 1.08	\\
CNN-VLM	& 0.79	& 0.43	& 4.55	& 0.79	& 0.71	& 0.44	& 1.18	& 1.06	\\
CORS	& 0.79	& 0.47	& 3.91	& 0.77	& 0.66	& 0.46	& 1.22	& 1.03	\\
SP	& 0.79	& 0.5	& 3.52	& 0.75	& 0.61	& 0.48	& 1.24	& 1.16	\\
MKL	& 0.78	& 0.42	& 4.4	& 0.78	& 0.61	& 0.42	& 1.08	& 1.1	\\
\hline
Baseline: Center	& 0.78	& 0.45	& 3.72	& 0.77	& 0.51	& 0.38	& 0.92	& 1.24	\\
\hline
Aboudib Magn.	& 0.78	& 0.48	& 3.56	& 0.75	& 0.56	& 0.45	& 1.12	& 1.19	\\
LS	& 0.78	& 0.43	& 4.4	& 0.77	& 0.64	& 0.39	& 1.02	& 1.16	\\
Rosin Saliency 2	& 0.78	& 0.48	& 3.43	& 0.73	& 0.53	& 0.45	& 1.13	& 1.21	\\
I-VGG16 (UID)	& 0.78	& 0.47	& 3.96	& 0.76	& 0.67	& 0.45	& 1.24	& 1.02	\\
FBFF	& 0.78	& 0.47	& 3.98	& 0.77	& 0.66	& 0.48	& 1.3	& 0.95	\\
SIM	& 0.77	& 0.46	& 4.17	& 0.76	& 0.64	& 0.43	& 1.14	& 1.53	\\
RARE2012	& 0.77	& 0.46	& 4.11	& 0.75	& 0.67	& 0.42	& 1.15	& 1.01	\\
LMF	& 0.77	& 0.45	& 4.22	& 0.76	& 0.64	& 0.41	& 1.07	& 1.02	\\
AIM	& 0.77	& 0.4	& 4.73	& 0.75	& 0.66	& 0.31	& 0.79	& 1.18	\\
SPPM	& 0.77	& 0.46	& 4.17	& 0.76	& 0.66	& 0.42	& 1.1	& 1.18	\\
EYMOL	& 0.77	& 0.46	& 3.64	& 0.72	& 0.51	& 0.43	& 1.06	& 1.53	\\
LGS	& 0.76	& 0.42	& 4.63	& 0.76	& 0.66	& 0.39	& 1.02	& 1.11	\\
MR-AIM	& 0.75	& 0.43	& 4.1	& 0.75	& 0.59	& 0.36	& 0.9	& 2.6	\\
RCSS	& 0.75	& 0.44	& 3.81	& 0.74	& 0.55	& 0.38	& 0.95	& 1.08	\\
Image Signature	& 0.75	& 0.43	& 4.49	& 0.74	& 0.66	& 0.38	& 1.01	& 1.09	\\
IttiKoch2	& 0.75	& 0.44	& 4.26	& 0.74	& 0.63	& 0.37	& 0.97	& 1.03	\\
VICO	& 0.75	& 0.44	& 4.38	& 0.71	& 0.6	& 0.37	& 0.97	& 1.96	\\
QCUT	& 0.75	& 0.39	& 4.57	& 0.67	& 0.57	& 0.4	& 1.07	& 6.71	\\
Aboudib Magn. Sal-v1	& 0.74	& 0.44	& 4.24	& 0.72	& 0.58	& 0.39	& 0.99	& 2.45	\\
Context-Aware sal.	& 0.74	& 0.43	& 4.46	& 0.73	& 0.65	& 0.36	& 0.95	& 1.06\\	
AWS	& 0.74	& 0.43	& 4.62	& 0.73	& 0.68	& 0.37	& 1.01	& 1.07	\\
WMAP	& 0.74	& 0.42	& 4.49	& 0.67	& 0.63	& 0.34	& 0.97	& 1.38	\\
GNM	& 0.74	& 0.42&	 4.49	& 0.67	& 0.63	& 0.34	& 0.97	& 1.21	\\
GMR	& 0.74	& 0.38	& 4.28	& 0.64	& 0.53	& 0.36	& 0.94 & 	7.38	\\
NARFI	& 0.73	& 0.38 &	 4.75	& 0.61	& 0.55	& 0.31	& 0.83	& 5.17	\\
SDWT	& 0.72	& 0.4 &	 4.71	& 0.7	& 0.61	& 0.31	& 0.84	& 1.19	\\
Self-res. LARK	& 0.71	& 0.41	& 4.55	& 0.69	& 0.64	& 0.31	& 0.83	& 1.54\\	
SFP	& 0.71	& 0.41&	 4.56	& 0.7	& 0.62	& 0.3	& 0.8	& 1.2	\\
Rosin Saliency 1	& 0.71	& 0.4	 & 4.86	& 0.7	& 0.62	& 0.29	& 0.76	&  1.13	\\
CIWM  & 0.7	& 0.38 &	 5.18	& 0.69	& 0.65	& 0.27	& 0.73	& 1.23	\\
Fabrice Urban	& 0.7	& 0.4 &	 5.03	& 0.7	& 0.64	& 0.29	& 0.78	& 1.23	\\
Torralba saliency	& 0.68	& 0.39	& 4.99	& 0.68	& 0.62	& 0.25	& 0.69	& 1.24	\\
\hline
Baseline: Perm.	& 0.68	& 0.34	& 4.59	& 0.59	& 0.5	& 0.2	& 0.49	& 6.12	\\
\hline
SUN saliency	& 0.67	& 0.38	& 5.1	& 0.66	& 0.61	& 0.25	& 0.68	& 1.27	\\
SOS CNN	& 0.65	& 0.37	&  4.96	& 0.65	& 0.61	& 0.24	& 0.67	& 1.32	\\
IttiKoch	& 0.6	& 0.2	  &  5.17	& 0.54	& 0.53	& 0.14	& 0.43	& 2.3	\\
Achanta	& 0.52	& 0.29	& 5.77	& 0.52	& 0.52	& 0.04	& 0.13	& 1.73	\\
\hline
Baseline: Chance	& 0.5	& 0.33 &	6.35	& 0.5	& 0.5 &	0	& 0	& 2.09	\\
\hline

\end{tabular}
\label{tab:MitRes}
\end{scriptsize}
\end{table}

\begin{figure}
	\centering
    \includegraphics[width=.93\linewidth, height=23.8cm ]{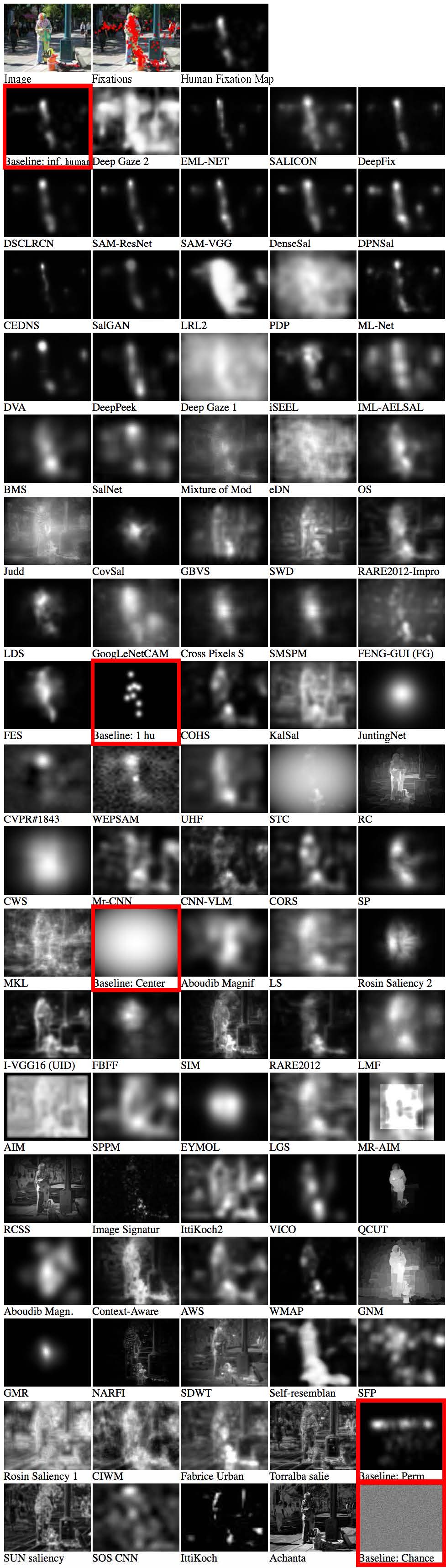} \\
    \vspace*{-7pt}
	\caption{Saliency maps of 90 models over a sample image. Models are sorted according to AUC-Judd score. Red boxes indicate baselines.}
	\label{fig:maps}
\end{figure}

\begin{figure}[t]
	\centering
    \includegraphics[width=.95\linewidth]{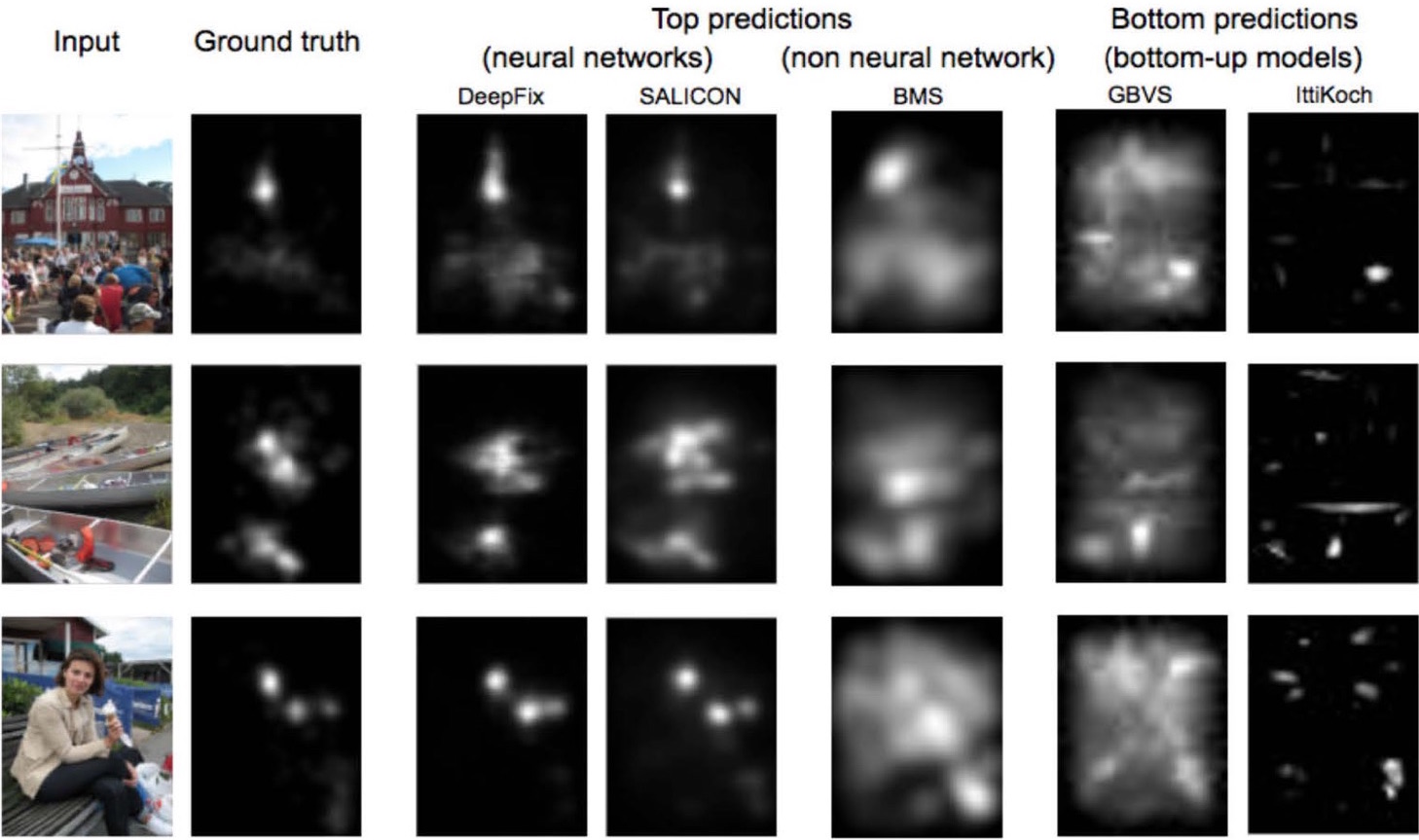} \\
   \vspace*{-5pt}    
	\caption{In Bylinskii~\etal~\cite{BylinskiiECCV2016}, we showed that a small set of 10 images can rank models with high accuracy. These images contain people at varying scales, as well as text amidst distracting textures. This figure shows some images for which deep models perform the best while classic non-deep models suffer the most. Refer to Fig. 2 in~\cite{BylinskiiECCV2016} to see all 10 images. }
	\label{fig:deepvsclassic}
	\vspace{-15pt}
\end{figure}

{\bf Results on MIT300:} Table~\ref{tab:MitRes} shows the results over the MIT300 dataset. According to all measures, the top 5 models are all NN-based. Using the AUC-Judd measure, DeepGaze II and EML-NET hold the top 2 spots with score of 0.88 (Inf. human scores 0.92). 
Based on the AUC-Borji measure, DeepGaze II and SALICON rank first and second scoring 0.86 and 0.85, respectively. The corresponding score for Inf. human is 0.88.
According to sAUC, SALICON and DPNSal are tied in first place with sAUC of 0.74 (sAUC of Inf. human is 0.81). Using this score, the only model scoring lower than the center baseline is EYEMOL.
While models perform close to each other according to AUC scores, switching to NSS widens the differences. EML-NET has the highest score with an NSS of 2.47 (NSS of Inf. human is 3.29). The second and third ranks here belong to CEDNS and DPNSal. Using CC and SIM scores, DPNSal is the winner (CC = 0.82 and SIM = 0.69). The Inf. human baseline scores 1 using both scores.
DenseSal and SALICON occupy the first two places using the KL measure (Inf. human scores 0). 
Using the EMD measure, the best model is EML-NE (Inf. human scores 0).

Rankings generated by different evaluation measures fluctuate drastically. For example, switching from AUC-Judd to AUC-Borji, DeepGaze II still ranks at the top, but EML-Net drops to the 44th place, even below the center baseline. Nevertheless, results suggest that
EML-Net, DeepGaze II, DPNSal, SALICON, CEDNS, DeepFix, SAM ResNet, SAM-VGG, DSCLRCN, and SalGAN 
alternate in providing the best prediction for ground-truth fixations. EML-Net is the only model that wins over 3 scores (AUC-Judd, EMD, NSS). DPNSal and DeepGaze II each win over 2 scores (DPNSal over CC and SIM; DeepGaze II over AUC-Judd and AUC-Borji).

BMS is the best-performing non neural network model (AUC-Judd = 0.83 and NSS = 1.41) and ranks better than the deep eDN model. The majority of models significantly outperform all the baselines, except the Inf. human. Among baselines, one human and center predict fixations better than permutation and chance.

Comparing the best results pre and during the deep learning era, shows about 43\% improvement in terms of NSS (EML-NET vs. BMS) and about 5.7\% improvement in terms of AUC-Judd (DeepGaze II vs. BMS). At the same time, the gap between the best model and the Inf. human shrinks from 57\% to 25\% using NSS, and from 9.8\% to 4.4\% in terms of AUC-Judd. Using the NSS score, about 73\% of the top 30 models are NN-based (about 67\% using AUC-Judd).

A careful examination hints towards performance saturation. Using the AUC-Judd, the top 2 models score 0.8, and the subsequent 8 models score 0.87. Using NSS, performance of the top 3 models varies between 2.41 and 2.47, which are still much less than 3.29 for the Inf. human baseline.

Among the scores, NSS, KL, and EMD provide a wider range of values ([0,$\infty$)), and thus can better separate models. Different scores favor different types of maps. For example, AUC-type scores favor smoothed maps, whereas NSS prefers pointy ones. This leads to a compromise is scoring, making one wonder whether a single model can win across all scores. See~\cite{kummerer2018saliency} and Section~\ref{challenges}.

A qualitative evaluation of model predictions over the sample image on the MIT benchmark website is presented in Fig.~\ref{fig:maps}. Recent deep models produce saliency maps that are very similar to the ground truth on this image (See Fig.~\ref{turing}).

 {\bf Representative stimuli:}
Is there a subset of few images that can offer a reliable model ranking? 
In~\cite{BylinskiiECCV2016}, we defined representative images as those that best preserve model rankings when tested on, compared to the whole dataset and followed a correlation-based greedy image selection to find them.
We found that a subset of 10 images can already rank the models on the MIT benchmark with a Spearman correlation of 0.97 relative to their ranking on all dataset images (See Fig. 2 in~\cite{BylinskiiECCV2016}). These images help to accentuate model differences. By visualizing the predictions of some of the top and bottom models on these images, we can see that driving performance is a model's ability to detect people and text in images amidst clutter, texture, and potentially misleading low-level pop-out (See Fig.~\ref{fig:deepvsclassic}). We are planning to update the representative image list as new models are added to the benchmark.

\begin{table}
\caption{Performance of saliency models over the CAT2000 dataset.}
\vspace{-10pt}

\centering
\renewcommand{\tabcolsep}{.8mm}
\renewcommand{\arraystretch}{.9}%
\begin{scriptsize}
\begin{tabular}{ | l | cccccccc | } 
\hline
& AUC-J  & SIM & EMD $\downarrow$ & AUC-B & sAUC & CC & NSS & KL$\downarrow$  \\ \hline
\hline
Baseline: infinite h.	& 0.9 & 	1	& 0	& 0.84	& 0.62	& 1	& 2.85	&  0 \\
\hline
SAM-ResNet	& \underline{{\bf 0.88}}	& \underline{{\bf 0.77}}	& \underline{{\bf 1.04}}	& 0.8	& 0.58	& \underline{{\bf 0.89}}	& 2.38	& 0.56 \\
SAM-VGG	& \underline{{\bf 0.88}}	& 0.76	& 1.07	& 0.79	& 0.58	& \underline{{\bf 0.89}}	& 2.38	& 0.54 \\
CEDNS	& \underline{{\bf 0.88}}	& 0.73	& 1.27	& 0.74	& 0.58	& 0.85	& \underline{{\bf 2.39}}	& \underline{{\bf 0.34}} \\
DeepFix	& 0.87	& 0.74	& 1.15	& 0.81	& 0.58	& 0.87	& 2.28	& 0.37 \\
EML-NET	& 0.87	& 0.75	& 1.05	& 0.79	& 0.59	& 0.88	& 2.38	& 0.96 \\
MixNet	& 0.86	& 0.66	& 1.63	& 0.82	& 0.58	& 0.76	& 1.92	& 0.62 \\
eDN	& 0.85	& 0.52	& 2.64	& \underline{{\bf 0.84}}	& 0.55	& 0.54	& 1.3	& 0.97 \\ 
BMS	& 0.85	& 0.61	& 1.95	& \underline{{\bf 0.84}}	& 0.59	& 0.67	& 1.67	& 0.83 \\
SeeGAN	& 0.85	& 0.68	& 1.45	& 0.78	& 0.56	& 0.81	& 2.11	& 1.27 \\
Judd	& 0.84	& 0.46	& 3.6	& \underline{{\bf 0.84}}	& 0.56	& 0.54	& 1.3	& 0.94 \\
iSEEL	& 0.84	& 0.62	& 1.78	& 0.81	& 0.59	& 0.66	& 1.67	& 0.92 \\
SMSPM	& 0.84	& 0.64	& 1.72	& 0.8	& 0.54	& 0.76	& 1.93	& 0.85 \\
\hline
Baseline: Center	& 0.83	& 0.42	& 4.31	& 0.81	& 0.5	& 0.46	& 1.06	& 1.13 \\
\hline
LDS	& 0.83	& 0.58	& 2.09	& 0.79	& 0.56	& 0.62	& 1.54	& 0.79 \\
EYMOL	& 0.83	& 0.61	& 1.91	& 0.76	& 0.51	& 0.72	& 1.78	& 1.67\\
FENG-GUI (FG)	& 0.83	& 0.43	& 4.18	& 0.81	& 0.56	& 0.49	& 1.2	& 1.07\\
RARE2012- Improved	& 0.82	& 0.54	& 2.72	& 0.81	& 0.59	& 0.57	& 1.44	& 0.76\\
FES	& 0.82	& 0.57	& 2.24	& 0.76	& 0.54	& 0.64	& 1.61	& 2.1\\
Aboudib Magn. & 0.81	& 0.58	& 2.1	& 0.77	& 0.55	& 0.64	& 1.57	& 1.41\\
SDDPM	& 0.81	& 0.52	& 2.31	& 0.8	& 0.54	& 0.51	& 1.22	& 1.44\\
GBVS	& 0.8	& 0.51	& 2.99	& 0.79	& 0.58	& 0.5	& 1.23	& 0.8\\
\hline
Baseline: Permutation	& 0.8	& 0.55	& 2.25	& 0.71	& 0.5	& 0.63	& 1.63	& 2.42\\
\hline
IttiKoch2	& 0.77	& 0.48	& 3.44	& 0.76	& 0.59	& 0.42	& 1.06	& 0.92\\
Context-Aware saliency	& 0.77	& 0.5	& 3.09	& 0.76	& 0.6	& 0.42	& 1.07	& 1.04\\
\hline
Baseline: one human	& 0.76	&  0.43	 &  2.51	&  0.67 & 	0.56 & 	0.56 & 	1.54 & 	7.77	\\			
\hline
AWS	& 0.76	& 0.49	& 3.36	& 0.75	& \underline{{\bf 0.61}}	& 0.42	& 1.09	& 0.94 \\
AIM	& 0.76	& 0.44	& 3.69	& 0.75	& 0.6	& 0.36	& 0.89	& 1.13 \\
WMAP	& 0.75	& 0.47	& 3.28	& 0.69	& 0.6	& 0.38	& 1.01	& 1.65 \\
Torralba saliency	& 0.72	& 0.45	& 3.44	& 0.71	& 0.58	& 0.33	& 0.85	& 1.6 \\
CIWM	& 0.7	& 0.43	& 3.79	& 0.7	& 0.59	& 0.3	& 0.77	& 1.14 \\
SUN saliency	& 0.7	& 0.43	& 3.42	& 0.69	& 0.57	& 0.3	& 0.77	& 2.22 \\
Achanta	& 0.57	& 0.33	& 4.46	& 0.55	& 0.52	& 0.11	& 0.29	& 2.31 \\
IttiKoch	& 0.56	& 0.34	& 4.66	& 0.53	& 0.52	& 0.09	& 0.25	& 6.71 \\
\hline
Baseline: Chance	& 0.5	& 0.32	& 5.3	& 0.5	& 0.5	 & 0	& 0	& 2 \\
\hline
\end{tabular}

\label{tab:CatRes}
\end{scriptsize}

\vspace{-10pt}

\end{table}

{\bf Results on CAT2000:} Table~\ref{tab:CatRes} shows the results over the CAT2000 dataset. 
According to the AUC-Judd, SAM-ResNet, SAM-VGG, and CEDNS are tied in the top with a score of 0.88 (Inf. human scores 0.90). 
EML-Net, the winner on the MIT300 dataset, is ranked second with a score of 0.87 (tied with DeepFix). Notice that as of now (Oct. 2018), we do not have submissions from DeepGaze models on this dataset.
Switching to NSS, CEDNS wins with score of 2.39 slightly above SAM-ResNet, SAM-VGG, and EML-Net (all with NSS of 2.38). Interestingly, using to AUC-Borji, the classic Judd model has the best score of 0.84, same as eDN, BMS, and the Inf. human. 
Using CC and SIM scores, SAM-ResNet and SAM-VGG are the winners followed by the EML-Net. 
SAM-ResNet wins over 4 scores (AUC-Judd, SIM, CC, and EMD). AWS is the winner according to the sAUC score.
Among classic models, BMS~\cite{zhang2013saliency} and EYMOL~\cite{zanca2017variational} perform better than the others. Overall, models that do well over the MIT300 dataset perform well here as well.

There is a 3.5\% improvement from the best non-deep model to the best deep model in terms of the AUC-Judd score. Improvement in terms of the NSS score is 30\%. The performance gap between the best non-deep model and the Inf. human upper-bound is 5.55\% and shrinks to 2.22\% using the best deep learning model (based on the AUC-Judd measure). The corresponding values are 41.5\% and 16.2\% using the NSS score. Results suggest that performance is saturating on this dataset. Three models have a NSS of 2.38 (tied in second rank), three models have AUC-Judd of 0.88 (tied in first rank), and 2 with the best CC equal to 0.89. Notice that the upper-bound values (Inf. human) for some scores are dataset dependent (\eg AUC and NSS scores), whereas upper-bound are fixed for some others (CC=1, SIM=1, EMD=0, KL=0).


In our previous work (supplement in~\cite{BylinskiiECCV2016}), we observed that models underperform over these categories of CAT2000 dataset:
Jumbled, Indoor, Satellite, Cartoon, Inverted, OutdoorManMade, Social, LineDrawing, and Art. Interestingly, and perhaps counter intuitively, models perform better over Sketch and low-resolution images compared to other categories (according to NSS score).

 {\bf MIT300 vs. CAT2000:}
The NSS and AUC-Judd values for the Inf. human (3.29 and 0.92) over the MIT300 dataset are higher than their corresponding values over the CAT2000 dataset (2.85 and 0.90). This indicates that observers are more consistent over the MIT300 dataset and partly contributes to the larger human-model gap over the MIT300 compared to the CAT2000 dataset (4.34\% vs. 2.22\% using AUC-Judd, and 25\% vs. 16\%; both using NSS).
This suggests that MIT300 is a harder dataset for models. We also found that CAT2000 is more center-biased witnessed by the higher performance of the center baseline on this dataset (AUC-Judd = 0.83, NSS = 1.06), compared to the corresponding values over MIT300 dataset (AUC-Judd = 0.78, NSS = 0.92).

\begin{table}[t]
\caption{Performance of models over the SALICON test set. Results are compiled from Cornia~\etal~\cite{cornia2016predicting} and Sen Jia~\cite{jia2018eml}.}
\vspace{-10pt}
\centering
\renewcommand{\tabcolsep}{2mm}
\renewcommand{\arraystretch}{.95}%
\begin{scriptsize}
\begin{tabular}{ | l | cccc | } 
\hline
& CC & sAUC & AUC-Judd & NSS \\ \hline
\hline
EML-NET~\cite{jia2018eml} & \underline{{\bf 0.886}} & 0.746 & 0.866 & 2.050 \\
SAM-ResNet~\cite{cornia2016predicting} & 0.842  & \underline{{\bf 0.779}}  & 0.883  & \underline{{\bf 3.204}} \\
DSCLRCN~\cite{liu2016deep} & 0.831  & 0.776  & 0.884  & 3.157\\
SAM-VGG~\cite{cornia2016predicting} & 0.825  & 0.774  & 0.881  & 3.143\\
ML-Net~\cite{cornia2016deep} & 0.743  & 0.768  & 0.866  & 2.789\\
MixNet~\cite{dodge2018visual} & 0.730  & 0.771  & 0.861  & 2.767\\
Kruthiventi \etal~\cite{kruthiventi2016saliency} & 0.780  & 0.760  & 0.880  & 2.610\\
SalGAN~\cite{pan2017salgan} & 0.781  & 0.772  & 0.781  & 2.459\\
SalNet~\cite{pan2016shallow} & 0.622  & 0.724  & 0.858  & 1.859\\
DeepGaze II~\cite{kummerer2017understanding} & 0.509  & 0.761  & \underline{{\bf 0.885}}  & 1.336\\ \hline
Human Inter-observer (IO)~\cite{kummerer2017understanding} & -  & -  & 0.89  & -\\ \hline
\end{tabular}

\label{tab:saliconRes}
\end{scriptsize}
\vspace{-15pt}
\end{table}

\begin{figure*}[t]
	\centering
  \includegraphics[width=1\linewidth,angle =0]{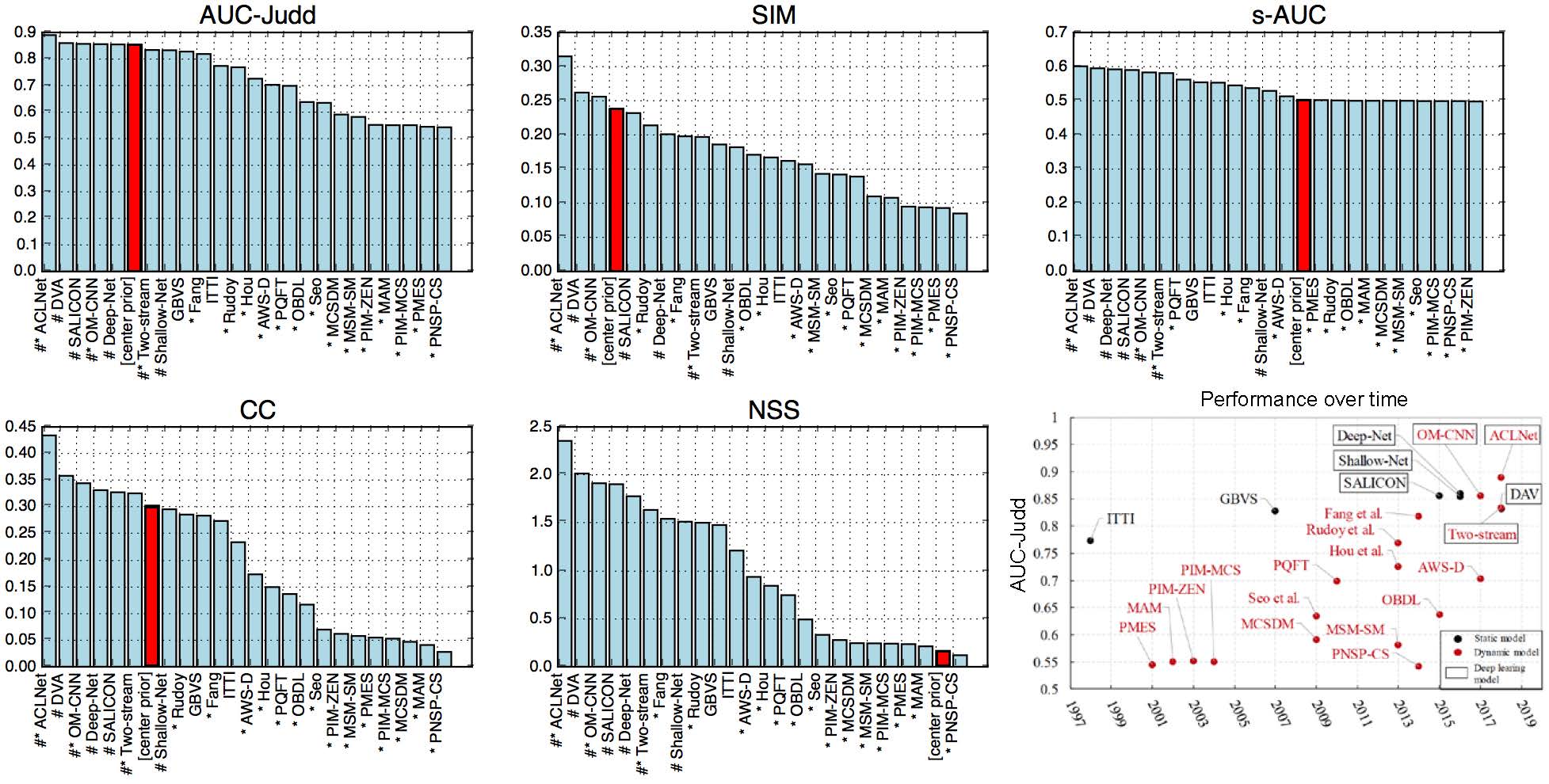} 
 \vspace{-13pt}    
	\caption{Performance of models over DHF1K dataset~\cite{wang2018revisiting}. 
	Models using the motion feature are marked with `*'. Deep models are marked with `\#'. Bottom-right: Dynamic saliency prediction performance over time, evaluated on
the DHF1K test set. Results are compiled from our work in~\cite{wang2018revisiting}.} 
	\label{fig:wangBench}
 \vspace{-10pt}    		
\end{figure*}

\begin{figure}[t]
	\centering
        \includegraphics[width=\linewidth]{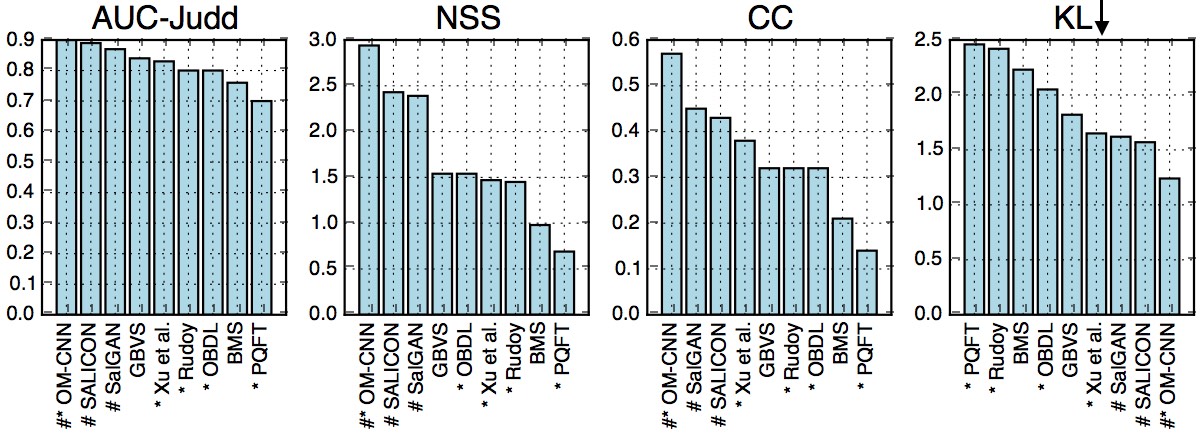} 
    \vspace{-20pt} \caption{Performance of models over LEDOV dataset~\cite{jiang2018deepvs}. Xu~\etal is the model in~\cite{xu2017learning}. Models using the motion are marked with `*'. Deep models are marked with `\#'.  Results are compiled from Jiang~\etal~\cite{jiang2018deepvs}.} 	
	\label{fig:DeepVS}
 \vspace{-15pt}    	
\end{figure}

\subsubsection{The SALICON Benchmark}
Unlike the MIT benchmark, SALICON does not provide a leaderboard. Consequently, I had to resort to the recent publications~\cite{cornia2016predicting,jia2018eml} to distill the 
results (Table~\ref{tab:saliconRes}).
EML-Net wins over CC (0.886), SAM-ResNet wins over sAUC and NSS (0.779 and 3.204), and DeepGaze II wins over the AUC-Judd score (0.885). Unfortunately, it is not possible to reliably characterize the gap between the models and humans on this dataset, as the scores of the upper-bound human model (as well as other baselines) are not available.   
However, based on the only available score, models still underperform humans (0.885 vs. 0.89~\cite{huang2015salicon}). Model accuracies vary in a narrow band here, suggesting possible saturation.

\subsection{Performance on Videos}

Twenty four models (23 models plus the center baseline) are evaluated on the DHF1K dataset using 5 scores NSS, CC, s-AUC, SIM, and AUC-Judd (Fig.~\ref{fig:wangBench}). Among these models, 7 are NN-based and 6 do not use temporal information. Scores are computed per frame and are then averaged. 
ACLNet~\cite{wang2018revisiting} consistently holds the first spot over all the scores. DVA~\cite{wang2018video} occupies the second rank. 
OM-CNN~\cite{jiang2018deepvs} and DeepNet~\cite{pan2016shallow} alternate in taking the third and forth places. Among the static models, DVA, DeepNet and SALICON perform better than the others. They also perform better than non-deep video saliency models. This result suggests that incorporating deep CNNs is more effective than augmenting temporal information into classic models, for predicting video saliency.
The center baseline performs better than several models due to strong center-bias in fixations over video datasets~\cite{borji2013quantitative}.


Results over the LEDOV dataset are compiled from~\cite{jiang2018deepvs} and are presented in Fig.~\ref{fig:DeepVS}. Nine models (3 NN-based, 4 without motion) are compared using 4 scores (AUC-Judd, NSS, CC, and KL). On this dataset, OM-CNN performs the best over all scores. Notice that ACLNet is not tested here. Consistent with the results on the DHF1K dataset, here deep models perform better than classic ones. GBVS is the best classic model here. PQFT model ranks at the bottom. Models perform better on the LEDOV dataset than DHF1K. This means that the latter is more difficult for current models.

Progression of model performance over time is illustrated in Fig.~\ref{fig:wangBench} (bottom-right). Overall, it seems that introduction of deep models has led to a big performance boost in video saliency prediction.

\section{Challenges and Open Problems}
\label{challenges}
Here, I discuss some challenges that need to be addressed to further narrow down the existing gap between current saliency models and the human inter-observer model.

\subsection{Characterizing the Errors of Models}
\begin{figure}[t]
	\centering
    \includegraphics[width=.9\linewidth]{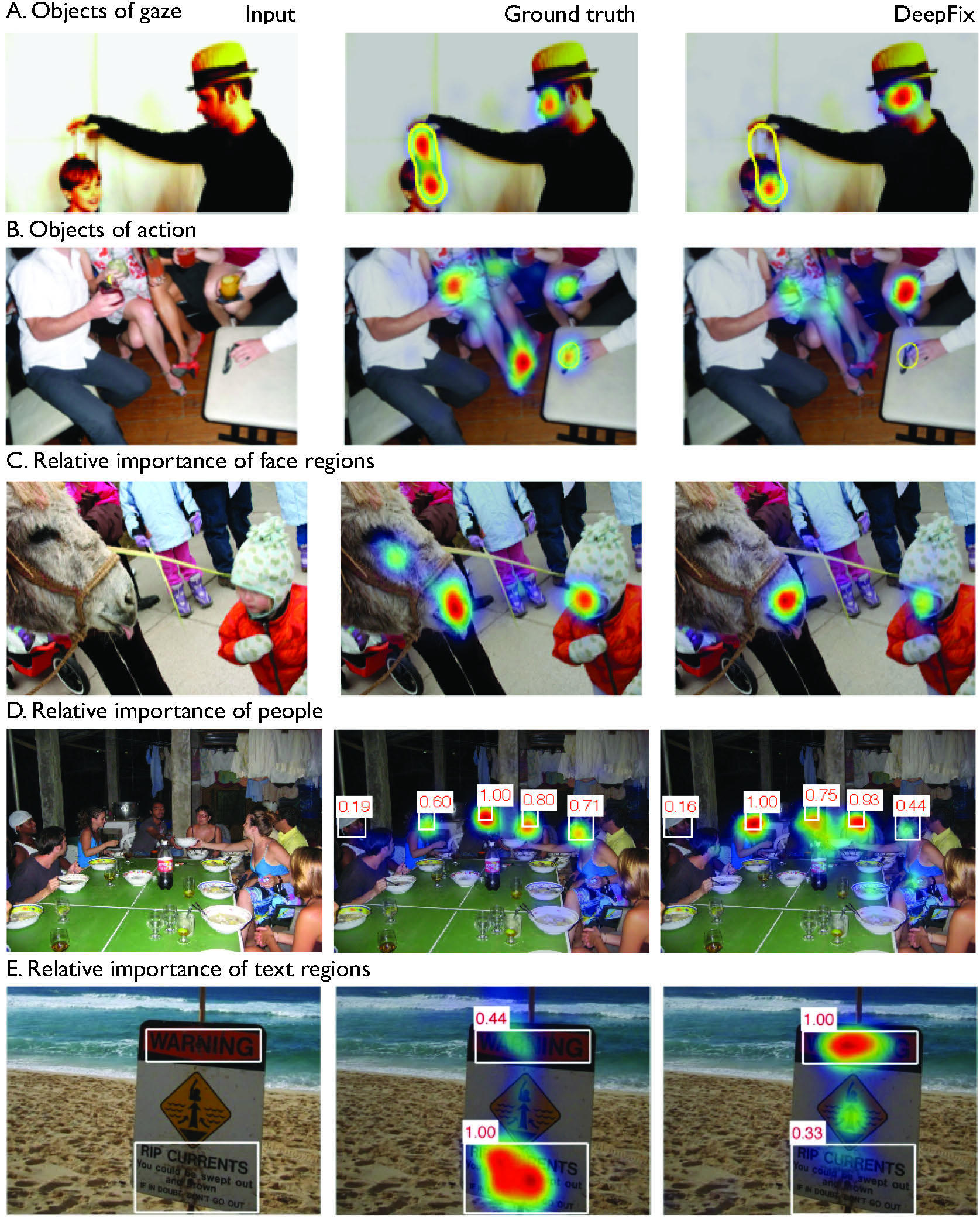} \\
    \vspace*{-7pt}
	\caption{Our finer-grained analysis in~\cite{BylinskiiECCV2016} shows that deep saliency models miss key elements in images. 
	Some example stimuli for which models under- or overestimate fixation locations due to gaze direction (A), or 
	locations of implied action or motion (B) are shown. 
Also, models fail to detect small faces or profile ones, and fail to assign correct relative importance to them (C). Cases for which models do not correctly assign relative importance to people (D), or text regions (E) in the scene are shown. Figure  borrowed from our work in Bylinskii~\etal~\cite{BylinskiiECCV2016}. 
}

   \vspace*{-5pt}
	\label{fig:failures}
\end{figure}

\begin{figure}
	\centering
    \vspace*{-5pt}
    \includegraphics[width=\linewidth]{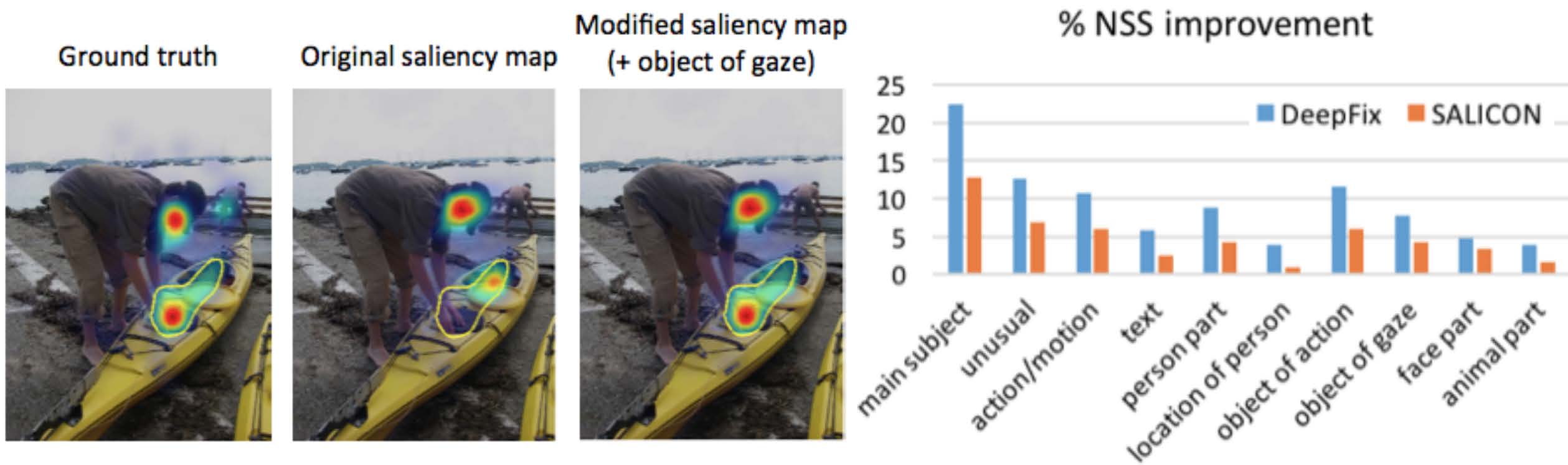} \\
        \vspace*{-10pt}
	\caption{Augmenting the prediction maps with the annotated regions dramatically increases the performance of saliency models, further indicating that models are missing fine-grained high-level concepts. 
	The right panel shows NSS of the enhanced models vs. original ones when adding different types of cues to the saliency models. Figure compiled from our work in Bylinskii~\etal~\cite{BylinskiiECCV2016}.} 
	\label{fig:augment}
   \vspace*{-10pt}
\end{figure}

\begin{figure}[t]
    \includegraphics[width=\linewidth]{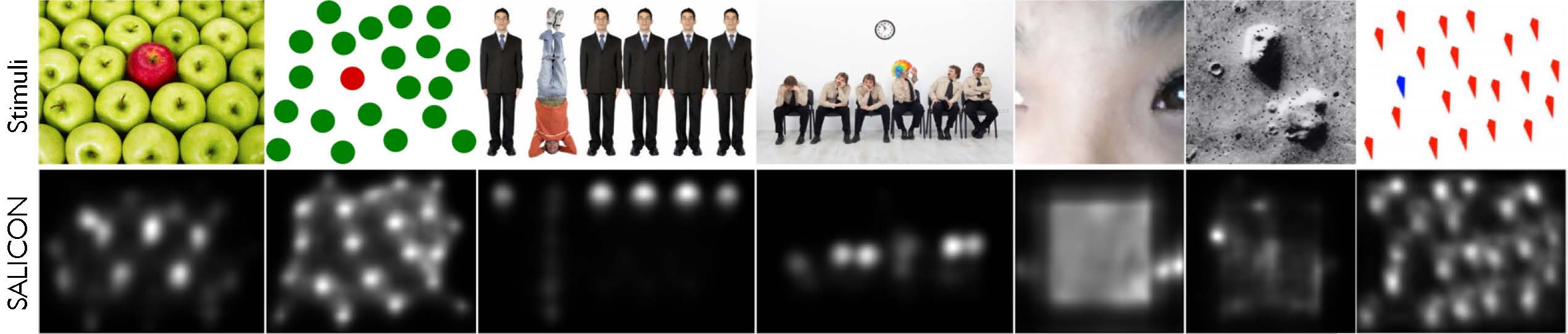}
    \vspace*{-15pt}
	\caption{Example stimuli where a deep saliency model produces counterintuitive results relative to models
based on feature contrast. 
Sometimes deep models neglect low-level image features (local contrast) and overweigh the contribution of high-level features (\eg faces or text). Figure compiled from Rahman and Bruce~\cite{rahman2015saliency} and Huang~\etal~\cite{huang2015salicon}.}
	\label{fig:bruce}
	    \vspace*{-15pt}
\end{figure}

As we saw, deep learning models have shown impressive performance in saliency prediction. A deeper look, however, reveals that they continue to miss key elements in images. 
In Bylinskii~\etal~\cite{BylinskiiECCV2016}, we investigated the state-of-the-art image saliency models using a fine-grained analysis on image types, image regions, etc. 
We conducted a behavioral study in which AMT workers were asked to label (1 out of 15 choices) image regions that fall on 
the top 5\% of the fixation heatmap. Analyzing the failures of models on those regions shows that about half the errors made by models are due to failures in accurately detecting parts of a person, faces, animals, and text. These regions carry the greatest semantic importance in images (Fig.~\ref{fig:failures}.A \& B). One way to ameliorate such errors is to train models on more instances of faces (\eg partial, blurry, small, or occluded faces, non-frontal views), more instances of text (different sizes and types), and animals. Saliency models may also need to be trained on different tasks, to learn to detect gaze and action and leverage this information for saliency. Moreover, saliency models need to reason about the relative importance of image regions, such as focusing on the most important person in the room or the most informative sign on the road (Fig.~\ref{fig:failures}.C, D \& E). Interestingly, when we added the missing regions (\ie under-predicted) to models, performance improved drastically (Fig.~\ref{fig:augment}; See also~\cite{nips15_recasens,gorji2017attentional,gorji2018going,henderson2018meaning}). Also, previous research (\eg ~\cite{rahman2015saliency,huang2015salicon}) has identified cases where deep saliency models produce counterintuitive results relative to models based on feature contrast (Fig.~\ref{fig:bruce}). 
In summary, more accurately predicting where people look in images and videos demands higher-level scene understanding. This needs to be studied further using both eye and mouse tracking, as will be detailed next.

\subsection{Mining Factors Driving Gaze during Free Viewing}
Several cues that attract gaze in free viewing have been discovered~\cite{wolfe2017five,2001Itti} (\eg color, orientation, intensity, depth, motion, flicker, texture, surprise, symmetry, scarcity, faces, text, signs, objects, memory, reward, semantic object distance, gaze and pointing directions~\cite{Borji_etal14jov,nips15_recasens}, emotional valence, vanishing point~\cite{borji2016vanishing}, focus of expansion, cultural differences, image center-bias, object center-bias~\cite{nuthmann2010object,borji2016reconciling}, scene context)
and have been incorporated into saliency models.
To fully close the gap between the human IO model and saliency models, parallel to developing saliency models, it is also necessary to understand how attention is deployed by humans (\eg by conducting behavioral studies). In other words, predicting where people look is not really the same as understanding attention. Better understanding where people look allows us to explain what models do, where they fail, and how to improve them.
Below, I enumerate some directions that need further exploration.

\textbf{Social cues.} It is well established that cues such as faces and text attract fixations in a task-independent manner (\eg Cerf~\etal~\cite{cerf2009faces}). Social interactions among people and finer properties of faces such as affect, gender, face-like patterns, number of faces, face size, pose, blur, familiarity, importance and beauty have been less explored. These are typical examples in which classic saliency models usually fail~\cite{BylinskiiECCV2016}. 
\textbf{Over images.} Existing datasets such as CAT2000~\cite{Borji_Itti15arxiv}, LFW~\cite{huang2007labeled}, and Crowd Scenes~\cite{jiang2014saliency} offer a rich set of stimuli 
that can be used to formulate and test several hypotheses addressing social cues. Some questions include: which facial components attract more fixations? how about frontal vs. profile faces? how do face and text compete in influencing gaze? how do relations among people guide attention (\eg one female among several males, and vice versa)? how do gaze and pointing directions compete~\cite{schauerte2010saliency}? 
{\bf Over videos.} Social cues seem to be stronger and more prevalent in dynamic conversation scenes. 
For example, sometimes gaze might switch from the active speaker to another person (to whom he is interacting with) for assessing emotional exchanges, etc. (\eg in response to an insult). Foulsham \etal~\cite{foulsham2010gaze} showed that high-status individuals were gazed at much more often, and for longer, than low-status individuals, even over short 20 second videos. Fixations were temporally coupled to the person who was talking at any one time. Further research is needed to probe the influence of the active speaker, main character, soundtrack, and subtitle on guiding gaze. The outcomes will break the ground for building multi-modal dynamic saliency models (\eg~\cite{coutrot2014saliency}).


\textbf{Action cues and affordance.} Previous research has shown that fixations are driven to action-related regions when performing daily tasks (\eg manipulating objects, reaching~\cite{belardinelli2016s}). Whether actions also guide fixations during free viewing is still unknown. Often, action cues are mixed with other cues (\eg gaze direction). Therefore, it is important to carefully disentangle them.
One way is designing a specialized set of images where an object is \textit{acted upon} vs. \textit{when it is not} (\ie control condition) similar to the approach in  
~\cite{borji2014complementary,Parks_etal15vr}, where we used paired stimuli. Some example actions that may guide fixations in free viewing include reaching, grasping, throwing, eating, and kicking (performed by the people in the scene).

\textbf{Saliency, importance, and interestingness.} These factors have been studied independently over small image sets (\eg~\cite{spain2011measuring,Elazary_Itti08jov,m2013fixations}). Since they are very subjective notions, a consistent, principled, comprehensive and large-scale investigation may bring new insights regarding their differences. Elazary and Itti~\cite{Elazary_Itti08jov} were the first to study interestingness and saliency over an existing dataset (LabelMe). Their study, however, is limited in two ways. First, their subjects were not given an explicit task. 
Due to this, subjects might have labeled easier objects (small objects as they were easier to draw a polygon around) instead of interesting ones. Second, they only used the Itti model. It would be interesting to verify that previous conclusions still hold using newer deep saliency models over both fixations and clicks. Also, the 
role of individual differences and culture~\cite{chua2005cultural}, viewer gender~\cite{shen2012top}, and alternative hypotheses than low-level saliency for predicting gaze (\eg scene and object meaning~\cite{henderson2018meaning}) need further exploration.


%

\textbf{Objects vs. low-level saliency.} Whether overt attention in natural scenes is guided by objects or low-level image features has become a matter of intense debate~\cite{einhauser2008objects,Borji_etal13jov,borji2016reconciling,stoll2015overt}. The object-based attention models have rivaled saliency models in predicting fixations. Therefore, it is important to study how they relate to low-level saliency models. Recent studies have shown that observers tend to fixate near the object center (\aka Preferred Viewing Location PVL~\cite{nuthmann2010object}). This is an impressive human ability but it is not clear whether new saliency models also pay more attention to the object center.
To gain insights, the following questions need to be answered: 1) does PVL, in humans and models, also hold over inverted scenes, line drawings, and blurred images? 2) how do object size and scale influence PVL? 3) how are symmetry and object center-bias related (\eg over the Kootstra dataset~\cite{kootstra2011predicting}),
4) how does PVL relate to actionable regions and affordance (\ie as in~\cite{belardinelli2016s}),
and 5) does PVL hold using mouse clicks?

\subsection{Explicit Saliency Judgments}

Rudoy~\etal~\cite{rudoy2012crowdsourcing} and Jiang~\etal~\cite{jiang2015salicon} were the first to utilize explicit saliency judgments for saliency modeling. Mouse clicks~\cite{jiang2015salicon,rudoy2012crowdsourcing} and salient objects~\cite{Borji_etal13vr} have been shown to correlate well, although imperfectly, with fixations. They have also been \textit{very instrumental} for training data-hungry deep models~\cite{huang2015salicon,kruthiventi2017deepfix}. 
Some important questions, however, remain unanswered: \textit{a) can clicks fully replace fixations}, and \textit{b) if not, what are the implications from a modeling perspective, and how best they can be utilized?}

\begin{figure}
	\centering
    \includegraphics[width=\linewidth]{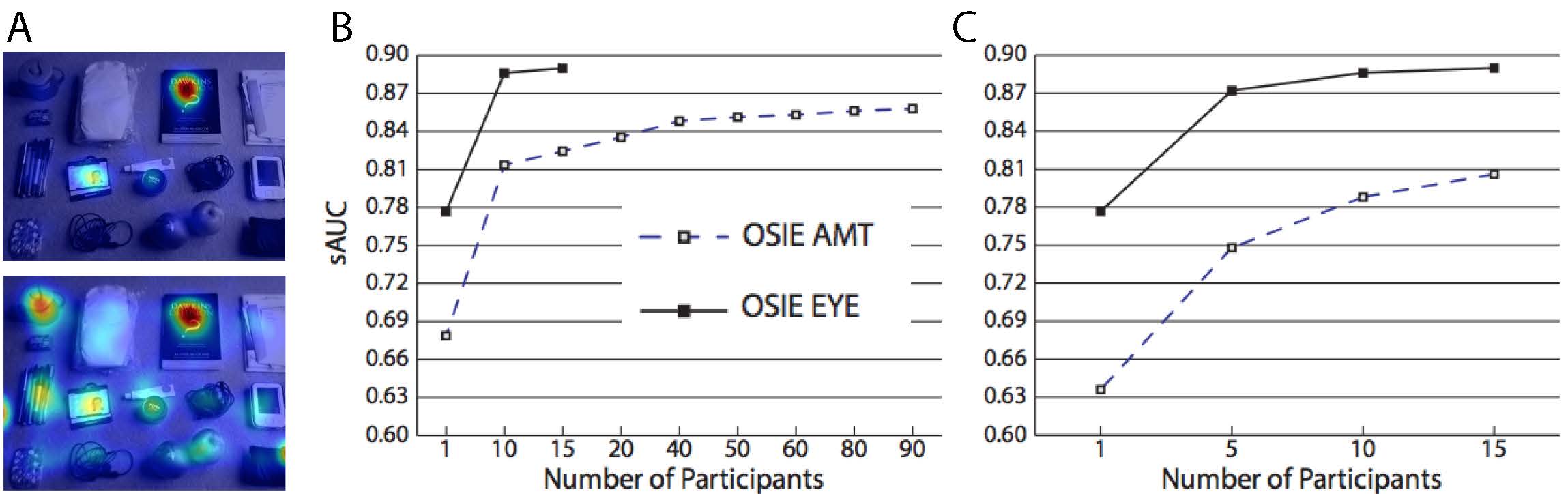}
    \vspace*{-15pt}
	\caption{A) An example image where fixations (top) do not completely agree with mouse clicks (bottom). B) Using eye movements and mouse clicks of a subset of participants to predict fixations of another group of participants, for increasing number of training participants. C) Inter-subject agreement using mouse click subjects and eye movement observers. Please see our work in Tavakoli~\etal~\cite{tavakoli2017saliency} for details.}
	\label{fig:hamed}
	\vspace*{-10pt}
\end{figure}

In~\cite{tavakoli2017saliency}, we investigated the similarity of fixations and clicks (Fig.~\ref{fig:hamed}.A). 
We ran two analyses over the OSIE dataset~\cite{xu2014predicting}. In the {\bf first} one, we randomly chose $m$ participants from the mouse tracking data of OSIE AMT dataset, and built a mouse density map to predict eye movements. We found that mouse data from even 90 participants can not achieve the same accuracy when using eye movements of 10 observers, indicating a significant gap between mouse tracking and eye tracking (Fig.~\ref{fig:hamed}.B). In the {\bf second} analysis, we measured the inter-subject congruency by building a map from $m$ subjects and applying it to the rest of the subjects using each type of data (\ie inter-observer models). As Fig.~\ref{fig:hamed}.C shows, there exists a higher dispersion among the mouse tracking participants compared to the eye tracking participants. To summarize, despite high agreement between clicks and fixations, there are still minor discrepancies:  \textit{clicks have lower inter-participant congruency and higher dispersion than fixations}. 
This result suggests that mouse clicks can not fully replace eye movements. Thus, we are left with two options: a) resorting to large scale eye tracking using desktop eye trackers or webcams, or b) harvesting good data from mouse clicks for model training. I believe that a combination of both is a more viable strategy.

\begin{figure}
	\centering
    \includegraphics[width=.9\linewidth]{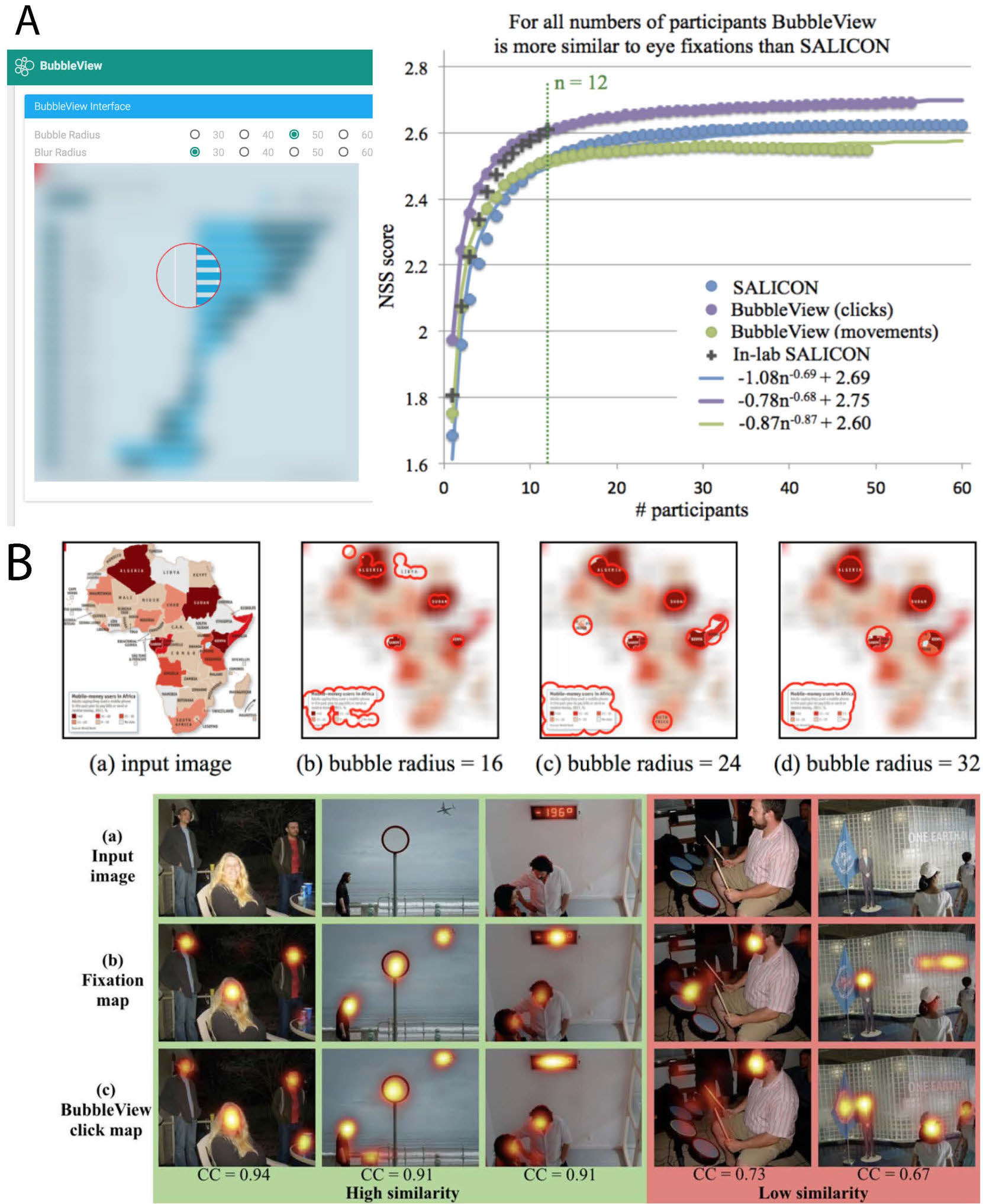} \\    
    \vspace*{-7pt}
	\caption{A) Left: An illustration of BubbleView by Kim~\etal~\cite{kim2017bubbleview} paradigm, Right: 
	The NSS score obtained by comparing mouse clicks and mouse movements to ground-truth fixations on natural images. 
	Each point represents the score obtained at a given number of participants, averaged over 10 random splits of participants and all 51 images used. It shows that BubbleView clicks better approximate fixations than SALICON mouse movements for all feasible numbers of participants ($n < 60$). 
	It also shows that clicks of 10 participants explain 90\% of fixations. B) Top: Clicks of 3 participants (b-d) who explored the same image (a) with BubbleView, but with different bubble sizes: 16, 24, and 32 pixel radius, respectively. The same regions of interest tended to be clicked on, despite differences in bubble sizes. Bottom: Example images from the OSIE dataset, a) dataset images, b) corresponding ground-truth fixation maps, and c) BubbleView click maps. Both cases when BubbleView maps have high similarity to fixation maps, and when they have low similarity to fixation maps are shown. Fig. compiled from~\cite{kim2017bubbleview}.}
	\label{fig:zoyaBubble1}
  \vspace*{-10pt}
\end{figure}

Kim~\etal~\cite{kim2017bubbleview} have recently developed a technique called BubbleView for collecting click data on images in scale. The subject is shown a blurred image and is asked to successively click on regions to unveil the story behind the scene. With each click a circular image region is revealed. They show that this strategy results in better data compared to SALICON paradigm and is closer to actual eye movements. They also showed that clicks of 10 participants explain about 90\% of fixations. Please see Fig.~\ref{fig:zoyaBubble1}. While subjects can correctly guess gaze locations on regions such as faces and text, it is not clear whether they can also estimate attention and gaze due to 
non-trivial subconscious cues (\eg spatial outliers, gaze direction, vanishing point, and symmetry). One way to approach this is by interspersing some images containing these cues (one cue at a time) among images without such cues (\eg outdoor natural scenes, line drawings). 
It would be also interesting to study \textit{1) whether subjects can be trained to better predict fixations when provided with feedback, 2) whether utilizing touch-screens (\eg tablets, cell phones) may yield better data\footnote{Touch-screens have the advantage that motor movement is in the same plane as the fixations and touching needs less effort than clicking.}, and 3) whether these paradigms can be extended to the video domain.} 

\subsection{ Invariance to Transformations} 
To truly approach human-level accuracy in fixation prediction, saliency models should be able to approximate gaze over a wide range of stimuli. Traditionally, researchers have resorted to upright images to estimate saliency. As a result, models trained on these images may lack generalization. Probing attention on transformed images not only offers invaluable insights regarding how attention is deployed by humans, also allows testing generalization power of models.

\begin{figure}[t]
	\centering
    \includegraphics[width=1\linewidth]{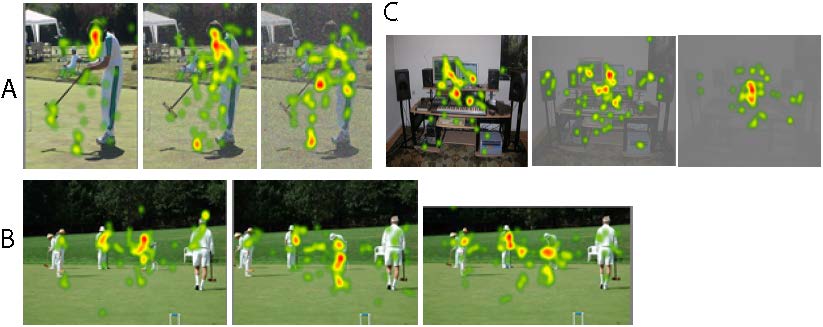} \\
    \vspace*{-8pt}
	\caption{Pilot data from our work in Che~\etal~\cite{invariance} showing fixations of one subject on images with low and high noise (A), high and low contrast (B), left and top crop (C).}
	\label{fig:invariance}
  \vspace*{-5pt}
\end{figure}

The hallmark of human object recognition ability is invariance to transformations such as rotation, scale, lighting, pose, and occlusion. I argue that, as in the object recognition literature, saliency models should perform similar to humans when applied to transformed images. 
A few previous works have studied fixations over blurred~\cite{judd2011fixations,gide2016effect}, noisy~\cite{Kim_milanfar,gide2016effect}, and low-resolution images~\cite{yohanandan2017saliency,gide2016effect}.
In a preliminary investigation~\cite{invariance}, we considered a larger array of transformations including horizontal flip, shear, cropping, line drawings, low and high contrast, low and high motion blur, etc. (20 in total). See Fig.~\ref{fig:invariance}. 
We found that model performances dramatically drop over all transformations. To build invariant saliency models, some questions need to be answered: \textit{1) how do bottom-up saliency and high level factors (\eg context, face) guide fixations on transformed images compared to original stimuli? 2) on which transformations subjects agree more with each other, 
3) how similar clicks and fixations are on transformed images, and 4) can models be trained to mimic human gaze over image transformations?} This research can be extended to the video domain by investigating the role of additional factors such as playing speed, viewing for comprehension vs. skimming, inversion, and mirroring.

\vspace{-10pt}
\subsection{New Image Datasets}
Saliency datasets are still orders of magnitude smaller than their counterparts in other areas of computer vision (\eg 15,000 images in SALICON vs. 1 million images in imagenet~\cite{deng2009imagenet}). Increasingly bigger datasets are still appearing. For example, Sun \etal~\cite{sun2017revisiting} collected a dataset of 300 million images (the JFT-300M) and showed that object recognition models performed better when trained on this dataset.
Motivated by this, I believe that future research in saliency will highly benefit from more data. Some possibilities are listed below. Further, this data provides new dimensions along which models can be compared.

\begin{figure}[t]
	\centering
    \includegraphics[width=\linewidth]{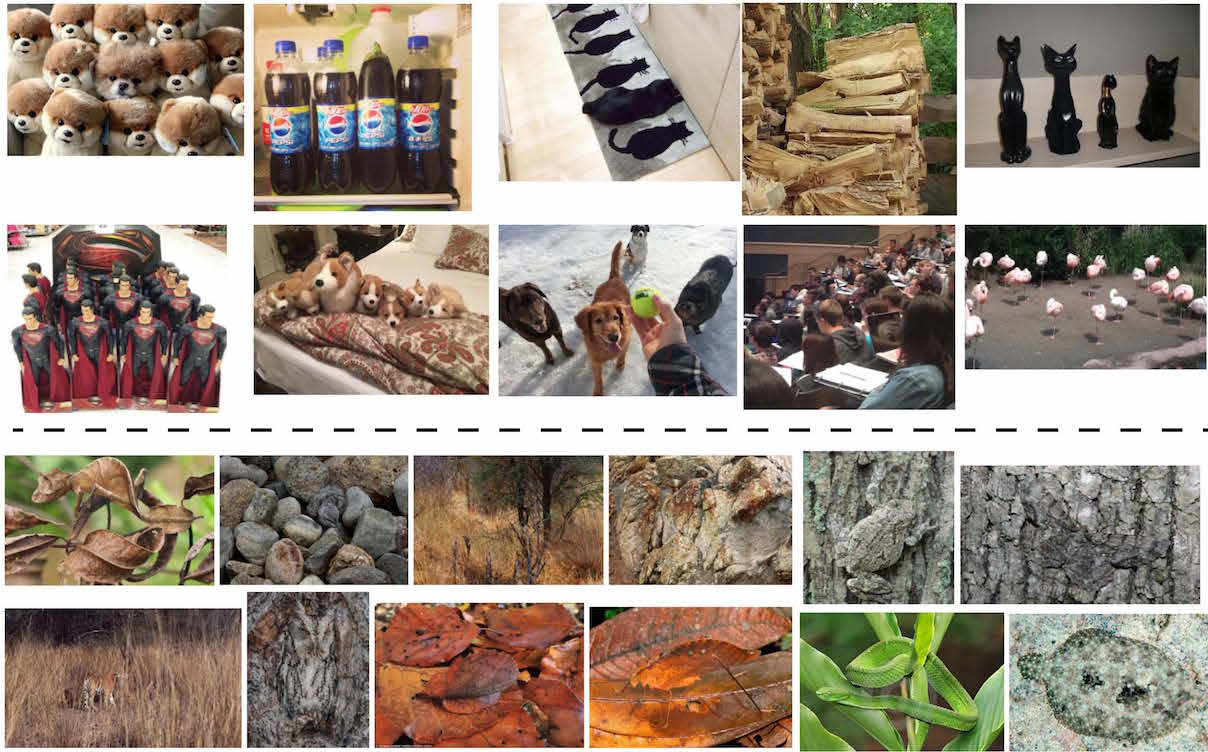} \\
    \vspace*{-10pt}
	\caption{Top) Example images where an odd item stands out from the rest of the items. Bottom) Example camouflage images. In contrast to above, here a careful investigation is needed to spot the target.}
	\label{fig:pattern}
  \vspace{-15pt}
\end{figure}

\begin{figure}[t]
	\centering
    \includegraphics[width=.9\linewidth]{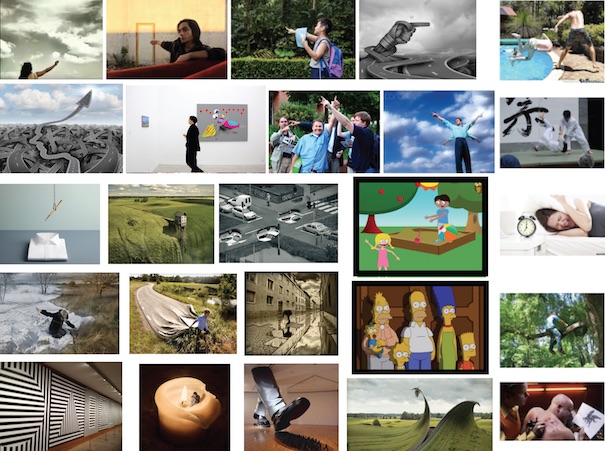} 
    \vspace{-8pt}
	\caption{Sample images containing rich semantic information (referred to as the `'Extreme Stimuli'' in the text). 
	 These images can be used to discriminate the models in their ability to capture factors that guide gaze at the semantic level. The two images with the black border show an example stimulus from the Clipart dataset by Zitnick and Parikh~\cite{zitnick2013bringing}, and a frame from ``The Simpsons'' cartoon (bottom).}
	\label{fig:stim}
	    \vspace{-5pt}
\end{figure}

\begin{figure}[t]
	\centering
    \includegraphics[width=.9\linewidth]{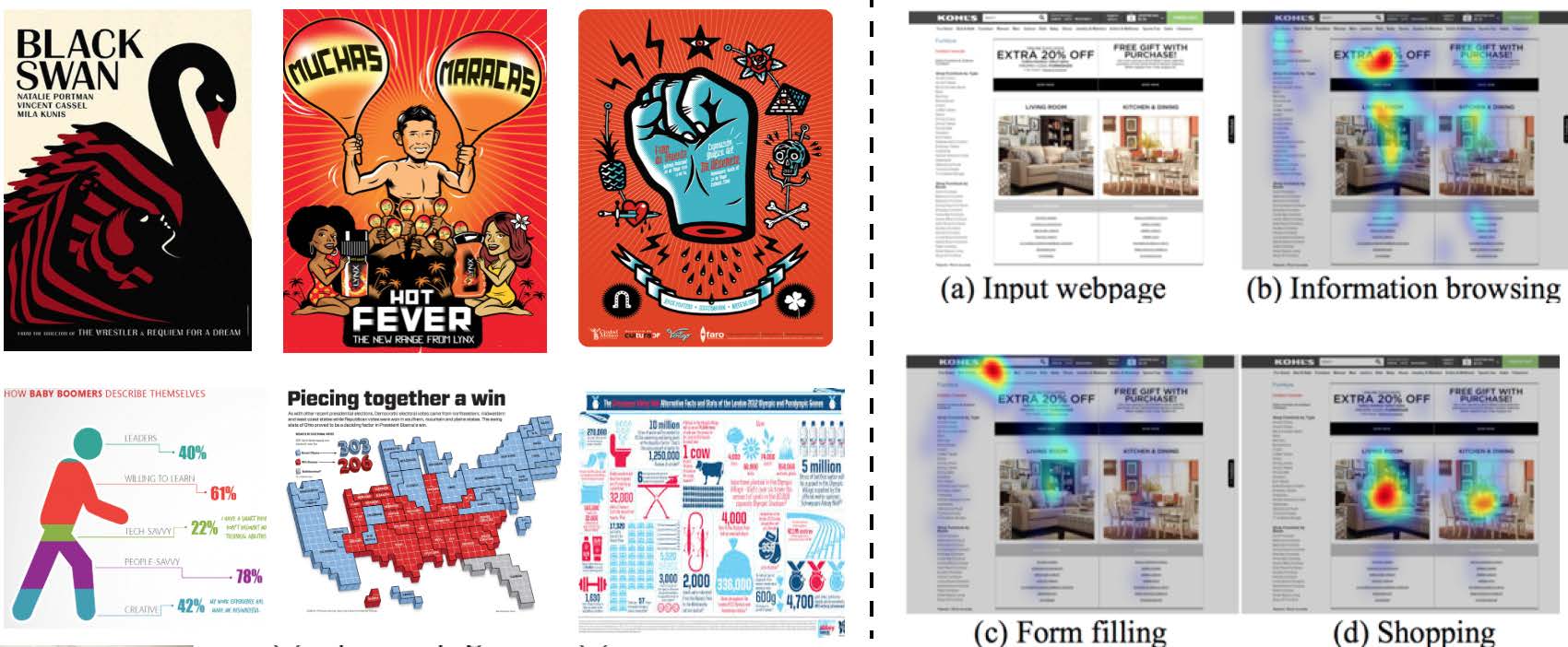} \\
    \vspace*{-8pt}
	\caption{ Left) Sample images from the \href{http://massvis.mit.edu/}{MASSVIS} dataset by Bylinskii~\etal~\cite{bylinskii2017understanding}, Right) Task-driven fixations over webpages by Zheng~\etal~\cite{zheng2018task}. }
	\label{fig:newStims}
 \vspace{-15pt}	
\end{figure}

\textbf{Natural oddball images and camouflages.}
Researchers have been relying heavily on complex natural scenes for training saliency models and to capture high-level and semantic attentional cues. Few works (\eg~\cite{borji2013quantitative}) have also tested models on synthetic pop-out search arrays (See Fig.~\ref{fig:bruce}). Natural oddball images such as the ones shown in Fig.~\ref{fig:pattern} (top) are largely ignored. Camouflage images, where a target is hidden in a cluttered background (Fig.~\ref{fig:pattern} (bottom)), also make an interesting set of stimuli for testing models.


\textbf{Extreme stimuli.}
The majority of the existing datasets are a mix of easy and difficult examples (with respect to humans) to make a general evaluation of models, but this can potentially create an imbalanced biased estimation and be problematic for model comparison. A model can be very good at difficult cases, but may fail over easy ones and vice versa. Averaging the accuracies over all types of easy and difficult stimuli conceals the strengths and weaknesses of models. 
In addition to applying models to a large set of various test images, comparing them against a small but very difficult set of images, such as the ones shown in Fig.~\ref{fig:stim}, can bring new insights. Such a dataset will particularly be useful when measuring the gap between models and humans, and is timely since saliency models are gradually approaching human level accuracy. 
These images can be collected in at least two ways: a) cases where best saliency models drastically fail when eye tracking data is available, and b) recording eye movements on images where subjects highly disagree to guess gaze locations, when only explicit saliency judgments are available.

\begin{figure}[t]
	\centering
    \includegraphics[width=1\linewidth]{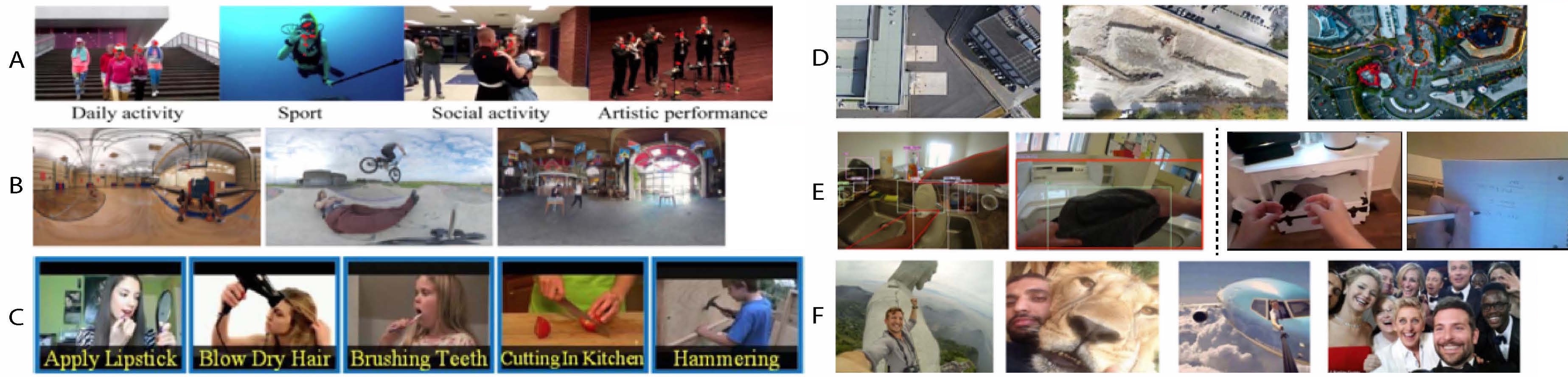} \\
        \vspace*{-10pt}
	\caption{Example frames from A) DHF1K dataset~\cite{wang2018revisiting}, B) Sal-360 dataset~\cite{zhang2018saliency}, C) UCF-101 dataset~\cite{soomro2012ucf101}, D) VIRAT~\cite{oh2011large}, E) ADL~\cite{pirsiavash2012detecting} and Charades-Ego~\cite{sigurdsson2018charades}, and F) YouTube Selfie videos and images.}
	\label{fig:newtd}
    \vspace*{-15pt}
\end{figure}

\textbf{Graphic designs, visualizations, and web pages.}
In a recent effort, Bylinskii \etal~\cite{bylinskii2017understanding} curated a new dataset of crowd-sourced graphic designs and visualizations, and analyzed the predictions of saliency models with respect to ground truth importance judgments (collected via BubbleView paradigm; Fig.~\ref{fig:newStims} (left)). They found that the current saliency models underperform a model specifically trained to predict fixations on this dataset. Similar observations have been made over webpages~\cite{zheng2018task} (Fig.~\ref{fig:newStims} (right)) and crowd scenes~\cite{jiang2014saliency}.
Collecting new data gives saliency researchers the opportunity to build domain-specific models that perform better than general saliency models.

\subsection{Large Scale Video Data and Benchmark} 
In spite of recent efforts in extending the scale and diversity of video saliency datasets (\eg DHF1K and Sal-360; Fig.~\ref{fig:newtd}.A \& B), available datasets still fall short in covering all cues (\eg semantic concepts, motion cues, main actors, background activities, and emotional content) that may guide gaze during watching dynamic scenes. For instance, over selfie videos, gaze might be directed more towards motions and objects related to the person's activity, rather than other stuff happening in the background or person's face. Thus, a saliency model that highlights just face regions would not perform well. Some new types of video data are as follows.

\textbf{Third-person videos and cartoons.}
One possibility is collecting gaze data on a subset of videos from the UCF-101~\cite{soomro2012ucf101} dataset. This dataset contains 13,320 YouTube videos from 101 action categories. It contains high action diversity and large variations in camera motion, object appearance and pose, object scale, viewpoint, background clutter, illumination, etc. Some example actions are shown in Fig.~\ref{fig:newtd}.C. Another possibility is 
drone videos which record activities from the top view(\eg VIRAT~\cite{oh2011large}; Fig.~\ref{fig:newtd}.D). Synthetic videos such as cartoons and games allow emphasizing more on high-level semantic and cognitive cues that may attract gaze. They have relatively lower amount of variability (\eg texture, clutter) compared to natural scenes. 
Some cartoons come with captions and audio that can be used to study the influence of language on attention, and vice versa (\eg ``The Simpsons''; Fig.~\ref{fig:stim}). 
One advantage of exploiting cartoons is that they can be annotated by adopting state of the art semantic segmentation models.

\begin{figure}[t]
	\centering
    \includegraphics[width=1\linewidth]{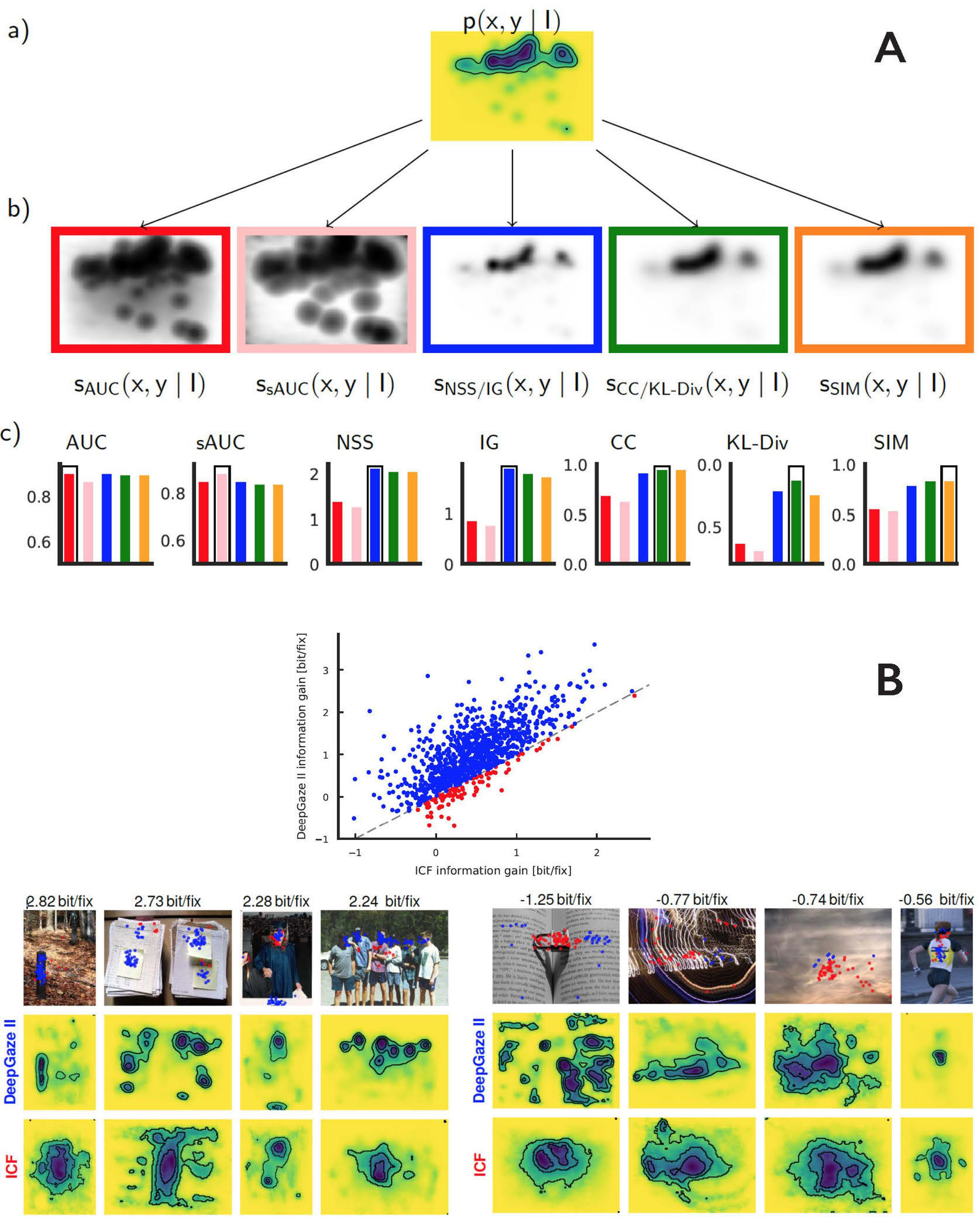} \\
    \vspace*{-5pt}
	\caption{A) In a series of works, Kummerer~\etal~\cite{kummerer2018saliency} showed that no single saliency map can perform best across all measures even when fixation distribution is known,
a) A probability distribution built from the ground truth fixations,
b) Predicted optimal saliency maps for each saliency measure.
These maps differ drastically due to the different
properties of the measures but reflect the same underlying model. 
Maps for the NSS and IG measures are the same, as are those for CC and KL-Div, 
c) Performances of the saliency maps from b) under seven measures on a large number of fixations sampled from the model distribution in a). 
The predicted saliency map for the specific measure (framed bar) yields the best performance in all cases. 
B) Performances of DeepGaze II and ICF models over the MIT1003 dataset~\cite{kummerer2017understanding}.
Each point corresponds to one image (information gain relative to the center bias).
Bottom-left) Images for which DeepGaze II has the largest improvement
over ICF. Fixations that are better explained by DeepGaze
II model (ICF model) are colored in blue (red). 
Predicted fixation densities for both models are plotted
below the images. Numbers above each stimulus are the difference
in information gain between DeepGaze II and ICF for that image. 
Bottom-right) Images for which ICF has the largest improvement over DeepGaze II. Figure compiled from~\cite{kummerer2018saliency,kummerer2017understanding}.}
	\label{fig:krummer}
    \vspace*{-15pt}
\end{figure}

\textbf{First-person and selfie videos.}   
First person videos pose unique challenges such as non-static cameras, unusual viewpoints, motion blur, variations in illumination, varying positions of the camera wearer, and real time video analysis requirements. Due to these, they need to be treated differently than third-person videos. In the context of saliency, 
studying how people view active videos in a passive manner can help build more general video saliency models.
Some example datasets that can be used here include ADL (Activities of Daily Living~\cite{pirsiavash2012detecting}) and Charades-Ego~\cite{sigurdsson2018charades} shown in Fig.~\ref{fig:newtd}.E. These datasets contain annotations such as
activities, object tracks, hand positions, and interaction events which allow a rich analysis of fixation guidance during daily activities. Lastly, selfie videos (Fig.~\ref{fig:newtd}.F) offer the opportunity to study how faces, emotions and actions relate to each other in guiding gaze.

\begin{figure*}[t]
	\centering

        \includegraphics[width=\linewidth]{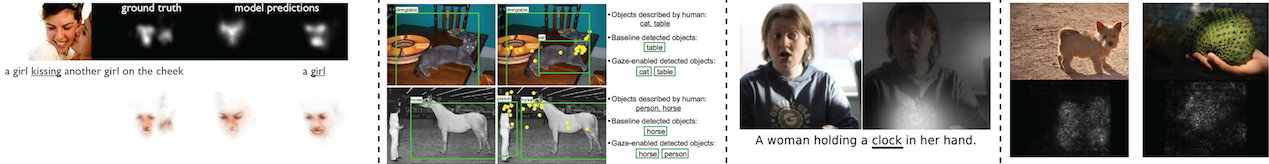} \\
    	   \vspace*{-5pt}
	\caption{Left) Fixations
	help uncover the story of images: who are the main actors, what are they doing, and what is going to happen next. As a result, fixations and saliency models can give us clues about human image captioning and understanding~\cite{BylinskiiECCV2016}.	
	In these examples, miss prediction by a saliency model has led to wrong or incomplete descriptions about the scene, Middle left) An example of how gaze can help improve object detection by Yun~\etal~\cite{yun2013studying}, Middle right) saliency predictors can be used to scrutinize image captioning models or clarify and improve otherwise ambiguous captions~\cite{xu2015show}, and Right) Saliency maps can be used to inspect where recognition models look in classifying scenes and objects.}
	\label{fig:imageUnderstanding}
	    \vspace*{-10pt}
\end{figure*}

\textbf{Video saliency benchmark.} Video saliency is lacking a comprehensive benchmark for assessing dynamic saliency models and identifying challenges. Such benchmarks are highly needed to accelerate future research in this area.
We have recently started such an effort by organizing the DHF1K benchmark (Wang~\etal~\cite{wang2018revisiting}) which in addition to a large dataset, provides a public evaluation server to compare models on a preserved test set. To become widespread, however, our effort requires community involvement.

\subsection{Fair Model Comparison and Scoring} 
Lack of consensus in evaluating models has been a big hindrance to saliency progress~\cite{borji2013quantitative,tatler2007central}. Several scores have been proposed, but each one suffers from a different problem. 
For example, the NSS score seems to be a better option~\cite{bylinskii2018different}, but similar to AUC, it is affected by center-bias (Tatler~\cite{tatler2007central}). While sAUC is more resilient to center-bias, it discards a large portion of valuable central fixations. These shortcomings 
have attracted substantial research to analyze the pros and cons of measures (\eg~\cite{bylinskii2018different}). 

In a recent study, K{\"u}mmerer \etal~\cite{kummerer2018saliency} showed that
it is impossible to solve the benchmarking problem when measures are evaluated on the same saliency maps. To solve this problem, they proposed to decouple the notions of models, maps, and measures to compare models. They 
derived the optimal saliency maps for the most commonly used measures (AUC, sAUC, NSS, CC, SIM, KL) and showed 
that this leads to consistent rankings in all measures and avoids the penalties of using one saliency map for all measures (Fig.~\ref{fig:krummer}.A). In view of these results, the MIT benchmark will soon implement this approach for model comparison.

K{\"u}mmerer \etal~\cite{kummerer2017understanding} conclude that no matter which score is used, there still might be images for which a weaker model performs better than the best model. In other words, higher average score does not mean a model wins over all cases.
To illustrate this, they compared the predictive power of low- and high-level features for saliency prediction by introducing two saliency models that use the same readout architecture built on two different feature spaces. The first one is the DeepGaze II, which uses high-level concepts and features learned via VGG. The second model called ICF (Intensity Contrast Features) uses simple intensity contrast features. While the DeepGaze II model significantly outperforms the low-level ICF as a whole, for a surprisingly large set of images the latter wins over the DeepGaze II (see Fig.~\ref{fig:krummer}.B). This implies that low performing models still hold value. For example, they can be combined with better models to build even stronger models (as in the integrated models in~\cite{dodge2018visual,wang2016learning}).

\subsection{Relation to Other Domains and New Applications}

New advances in deep learning (\eg GANs~\cite{goodfellow2014generative}) have opened the door to new applications for saliency models (Fig.~\ref{fig:imageUnderstanding}). Some examples include saliency-based image enhancement, compression, visualization of CNNs, image inpainting, saliency-augmented GANs, image and video captioning~\cite{schwartz2017high}, visual question answering~\cite{lu2016hierarchical}, 
zero-shot image classification~\cite{jiang2017learning}, and patient diagnosis~\cite{wang2015atypical}.

\begin{figure}
	\centering

    \includegraphics[width=\linewidth]{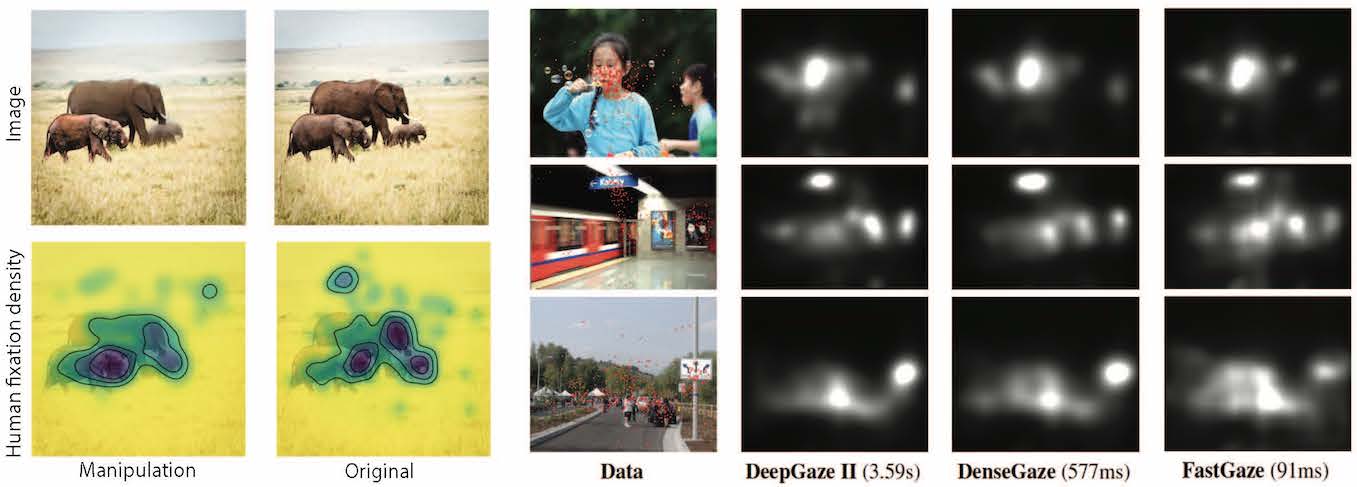} \\
        \vspace{-8pt}
	\caption{Left) Manipulating an image to transform its saliency by Gatys~\etal~\cite{gatys2017guiding}, Right) Example saliency maps from CAT2000 dataset along with predictions of DeepGaze II, DenseGaze and FastGaze models, with timings. All models produce similar saliency maps (Fig from ~\cite{theis2018faster}).}
	\label{fig:Manipulation}
   \vspace*{-13pt}	
\end{figure}

Here, I briefly describe two recent works that have utilized saliency for NN-based image manipulation or have used neural networks to speed up saliency computation. In the first one, Gatys~\etal~\cite{gatys2017guiding} proposed a method to automatically tweak the pictures in order to attenuate the distractors in favor of making objects of interest more salient (Fig.~\ref{fig:Manipulation} (left)). They trained a CNN to transform an image such that its saliency map satisfies a given distribution. For network training, their designed a loss function to achieve a perceptual effect while preserving naturalness of the transformed image. 
The second application regards the work by Theis \etal~\cite{theis2018faster}. They were able to speed up saliency computation by a factor of 10 while achieving about the same AUC score as the DeepGazeII, using dense networks and Fisher pruning (Fig.~\ref{fig:Manipulation} (right)).

\section{Discussion}
\label{discussion}


Saliency modeling falls at the intersection of human and computer vision. Thus, to boost the performance of models it is not enough to only try new architectures. Rather, it is important to study how humans view scenes, what guides gaze, what models learn, and where they fail. Here, I highlight some directions to transcend capability of models.

{\bf The need for higher level visual understanding.}
A finer-grained analysis of the top-performing models on the MIT Saliency Benchmark shows that they share common failures. 
Deep saliency models are good face and text detectors, much better than their predecessors. But, how good they are compared to the state of the art face and text detectors on labeled face datasets? Likewise, how much face and text detectors can explain fixations and 
how competitive classic saliency models augmented with face and text are against deep saliency models. 
For example, it would be desirable if a model does well on face detection and misses the ones that people do not pay much attention to.

Even the best saliency predictors today tend to place a disproportionate amount of importance on text and humans, even when they are not necessarily the most semantically interesting parts of the image. Saliency models will need to reason about the relative importance of image regions, such as focusing on the most important person in the room or the most informative sign on the road (which is image- and context-dependent). For example, in cases where there are multiple faces in a scene, determining important people requires extracting gaze direction, body pose, and higher-level features of individuals. Similarly, in the presence of different text regions in an image, some high-level understanding of meanings is necessary (\eg what is the warning sign about? what is the book about? what is inside the box?). To gain better insights, laboratory experiments can be conducted to assess how humans prioritize different cues (\eg text and faces) over stimuli designed for a special purpose (\eg free viewing, visual search). 

In short, saliency models still cannot fully understand the high-level semantics in semantically rich scenes (\ie the `'semantic gap''). To continue to approach human-level performance, saliency models will need to discover increasingly higher-level concepts in images: text, objects of gaze and action, locations of motion, and expected locations of people in images as well as determining the relative importance of objects. Cognitive attention studies can help overcome some of these limitations.


{\bf Harvesting high quality data.}
Collecting data for constraining, training and evaluating attention models is crucial to progress. Bruce \etal~\cite{bruce2015computational} pointed out that the manner in which data is selected, ground truth is created, and prediction error is measured (\eg loss function) are critical to model performance. Large scale click datasets~\cite{huang2015salicon,kim2017bubbleview} have been highly useful to train deep models and to achieve high accuracy. However, clicks occasionally disagree with fixations. Thus, separating good clicks from noisy ones, can improve model training. Moreover, studying discrepancies between mouse movements and eye movements, collecting new types of image and video data, as well as refining available datasets can be rewarding.

{\bf Multi-modal and multi-label data.} Data from different modalities (visual, auditory, captions, etc.) or with richer labels can help build more predictive saliency models. For example, Coutrot~\etal~\cite{coutrot2014saliency} showed that incorporating audio data can help fixation prediction over videos. Gorji \& Clark~\cite{gorji2017attentional,gorji2018going} and Recasens~\etal~\cite{nips15_recasens} showed that augmenting deep saliency models with gaze direction leads to higher performance over images and videos. Multi-label data allows training multi-task networks which can lead to better performance over individual tasks. For example joint learning of saliency, captioning, and visual question answering (VQA) could benefit all three. Existing datasets in one domain (\eg VQA) can be augmented with annotations of other tasks (\eg fixations, captions). Also correlations between saliency maps generated by VQA or captioning models against bottom-up saliency models can reveal good insights regarding how these models work.



{\bf Visualizing and understanding visual saliency models.} 
There has been a lot of progress in understanding the representations learned by CNNs for scene and object recognition in recent years (\eg~\cite{zhou2014object}). Our understanding of what is learned by deep saliency models, however, is limited. The question is 
how saliency computation emerges inside deep saliency architectures and how the learned patterns in different network layers differ from those learned for object recognition? One way to probe this by looking inside the same deep network, once trained for object recognition and once trained for saliency prediction, on the same set of stimuli containing the two types of labels: object category labels and saliency annotations (\eg clicks). 

{\bf Analysis of evaluation procedures. }
There has been a significant recent progress in understanding saliency evaluation measures (\eg~\cite{bylinskii2018different,kummerer2018saliency}). Some measures such as AUC, NSS, SIM, KL and IG are more favored over the others. Even using these measures, several saliency methods compete closely with one another at the top of the existing benchmarks and performances vary in a narrow band. 
As the number of saliency models grows and score differences between models shrink, evaluation measures should be adjusted to a) elucidate differences between models and fixations (\eg by taking into account the relative importance of spatial and temporal regions), and b) mitigate sensitivity to map smoothing and center-bias. There is a also an urge for designing appropriate measures for comparing video saliency models. Complementary to measures, finer-grained stimuli such as image regions in a collection or in a panel (such as the one in Fig. 11 in~\cite{BylinskiiECCV2016} to measure how well models predict the relative importance of image content), psychophysical patterns (pop-out search arrays and natural oddball scenes), transformed images, as well as extremely difficult stimuli can be used to further differentiate model performances. In sum, saliency evaluation, as in other areas of computer vision (\eg GANs~\cite{borji2018pros}) is still evolving.

\vspace{-10pt}

\section{Conclusion}

Saliency prediction performance has improved dramatically in the last few years, in large part due to
deep supervised learning and large scale mouse click datasets. The new NN-based models are trained in a single end-to-end manner, combining feature extraction, feature integration, and saliency value prediction, and have created a large gap in performance relative to traditional saliency models. The success of these saliency prediction models suggests that the high-level image features encoded by deep networks (\eg sensitivity to faces, objects and text), as well as the ability of CNNs to capture global context are extremely useful to predict fixation locations. The new deep saliency models, however, still suffer from several shortcomings that need to be addressed to reach human level accuracy. Analysis of model failures can help us design new and better models for that next qualitative leap in performance.


In this review, I covered a large volume of recent work in saliency prediction and proposed several avenues for future research. In spite of tremendous efforts and huge progress, there is still room for improvement in terms 
finer-grained analysis of deep saliency models, understanding models and evaluation procedures, datasets, tasks, annotation methods, cognitive attention studies, and applications. 



%
%
%
%

\vspace{-10pt}

\ifCLASSOPTIONcompsoc
  \section*{Acknowledgments}
\else
  \section*{Acknowledgment}
\fi

I would like to thank Zoya Bylinskii, Fredo Durand, Laurent Itti, Tilke Judd, Aude Oliva, and Antonio Torralba for their help in organizing the MIT saliency benchmark, and anonymous reviewers for their careful reading of the manuscript and their many insightful comments and suggestions.

\ifCLASSOPTIONcaptionsoff
  \newpage
\fi



%

\vspace{-10pt}

{\small
\bibliographystyle{IEEEtran}

\bibliography{ilab.bib}
}

%

%

%





\setcounter{section}{0}
\renewcommand\thesection{\Alph{section}}

\section{Appendix}



This appendix provides additional quantitative and qualitative results over the MIT benchmark including MIT300 and CAT2000 datasets~\cite{mit-saliency-benchmark}. \\

\begin{figure*}[t]
\vspace{-15pt}
	\centering
    \includegraphics[width=1.3\linewidth, angle =90]{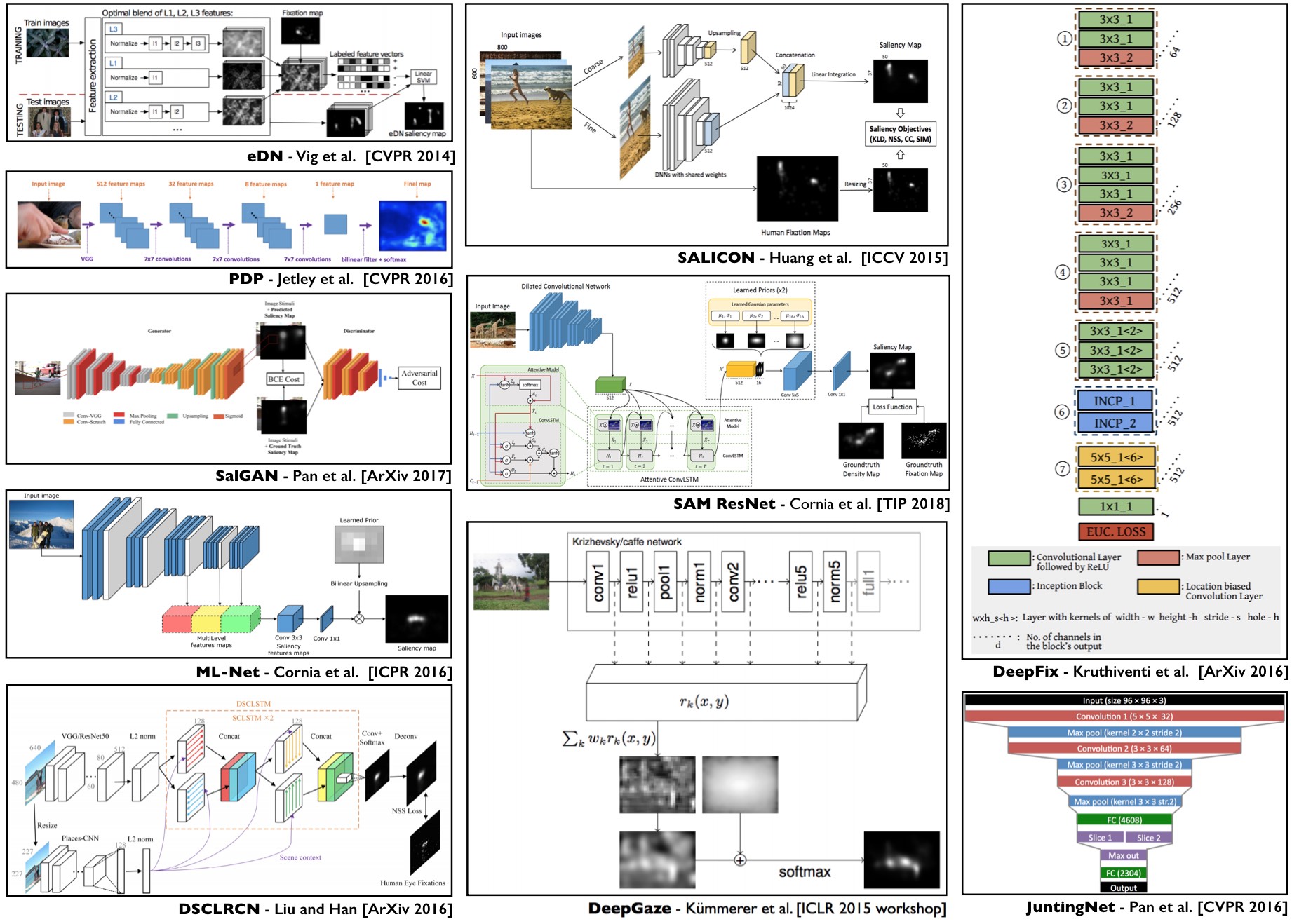} \\
	\caption{An illustration of 10 deep architectures for saliency prediction.}
	\label{fig:models}
\end{figure*}

Fig.~\ref{fig:models} illustrates a number of recent image-based saliency models grounded in deep learning. \\

\begin{figure*}
	\centering
    \includegraphics[width=\linewidth]{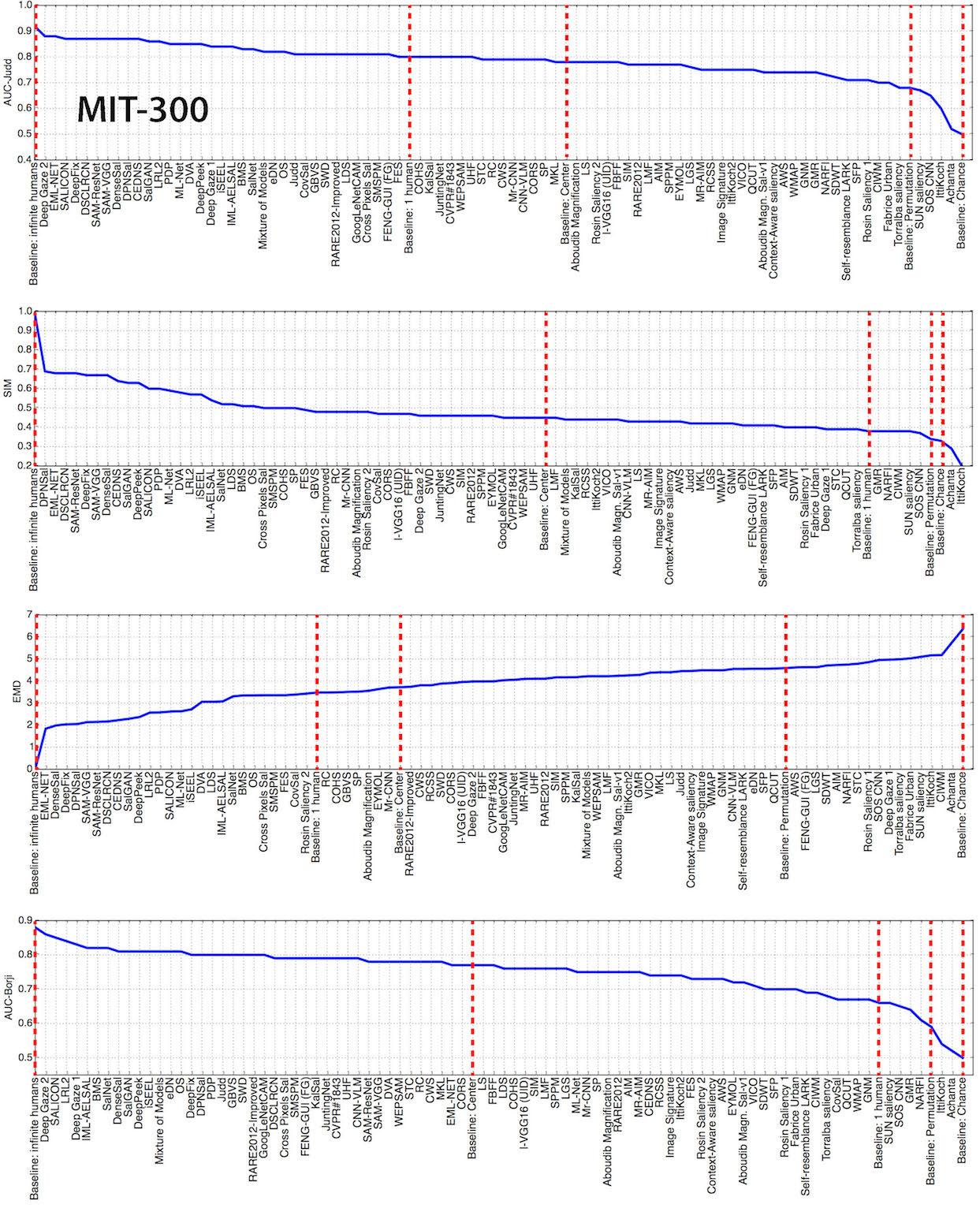} \\
    \vspace*{-7pt}
	\caption{Performance of static visual saliency models on the MIT300 dataset for AUC-Judd, SIM, EMD, and AUC-Borji scores. For all of these scores, the higher the number the better (except EMD where the opposite is true). Numbers are collected from the MIT saliency benchmark. Dashed lines represent the baselines.}
	\label{fig:newMIT-1}
\end{figure*}

Fig.~\ref{fig:newMIT-1} shows the ranking of models over the MIT300 dataset sorted based on the AUC-Judd, SIM, EMD, and AUC-Borji measures. Note that the lower the EMD, the better. \\

\begin{figure*}
    \includegraphics[width=\linewidth]{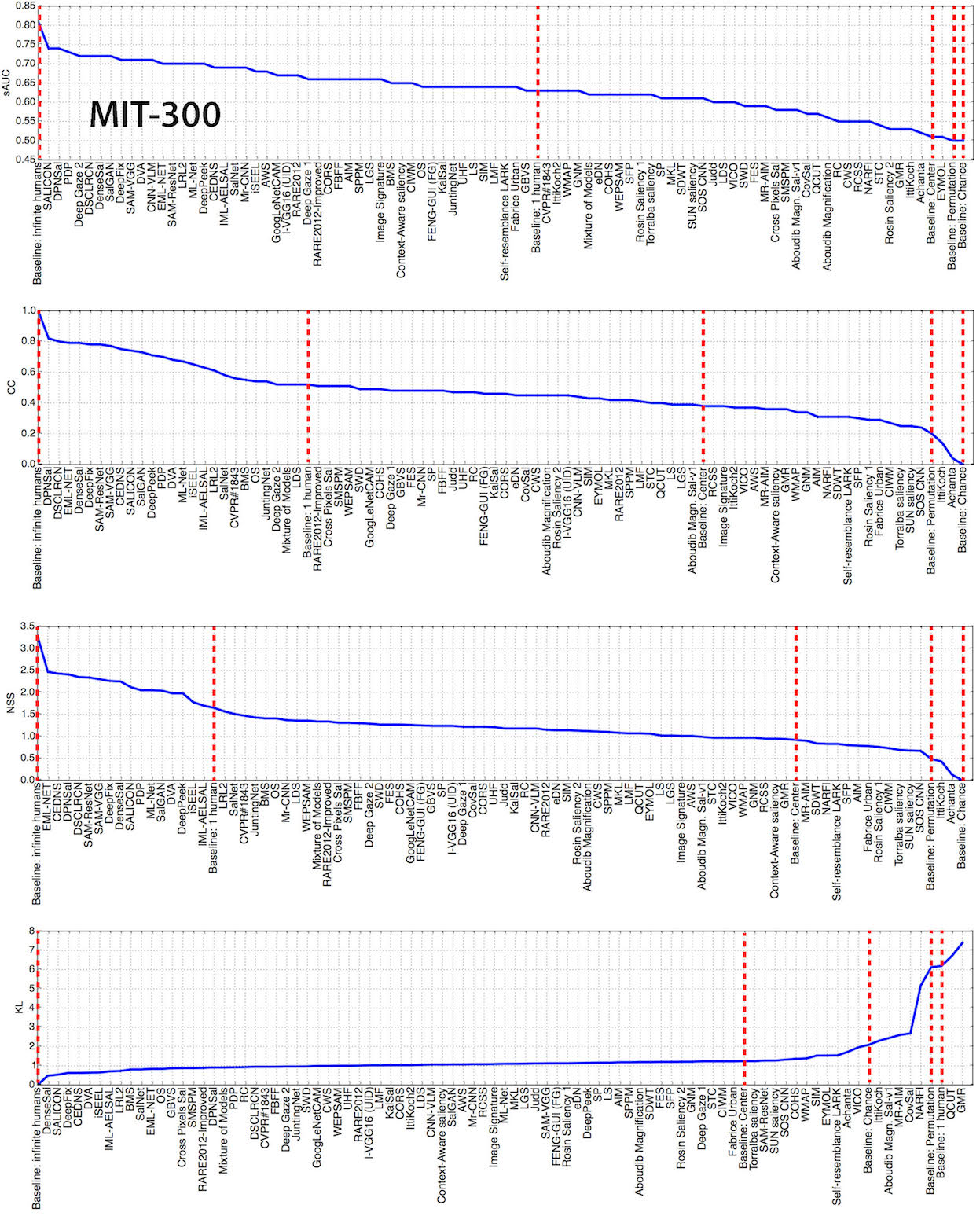} \\
    \vspace*{-10pt}
	\caption{Performance of static visual saliency models on the MIT300 dataset for sAUC, CC, NSS, and KL scores.  For all of these scores, the higher the number the better (except KL where the opposite is true). Numbers are collected from the MIT saliency benchmark. Dashed lines represent the baselines.}
	\label{fig:newMIT-2}
\end{figure*}

Fig.~\ref{fig:newMIT-2}  shows the ranking of models over the MIT300 dataset sorted based on the sAUC, CC, NSS, and KL measures. Note that the lower the KL, the better. \\

\begin{figure*}
	\centering
    \includegraphics[width=\linewidth, height= 20cm]{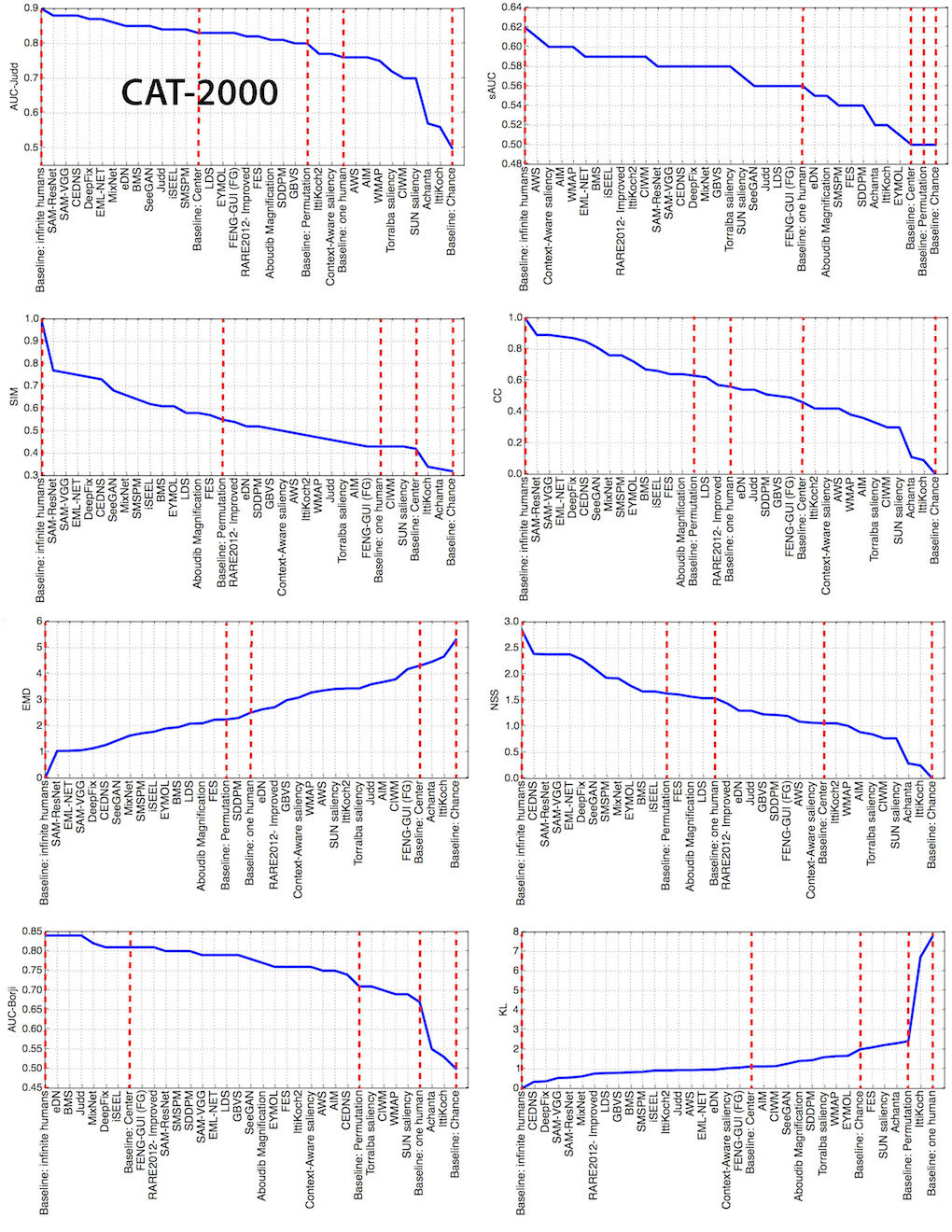} \\
	\caption{Performance of static visual saliency models over the CAT2000 dataset. Numbers are collected from the MIT saliency benchmark. Dashed lines represent the baselines.}
	\label{fig:newCat}
\end{figure*}

Fig.~\ref{fig:newCat} presents the model performance over all eight scores over the CAT2000 dataset. \\

\begin{figure*}
	\centering
    \includegraphics[width=1\linewidth ]{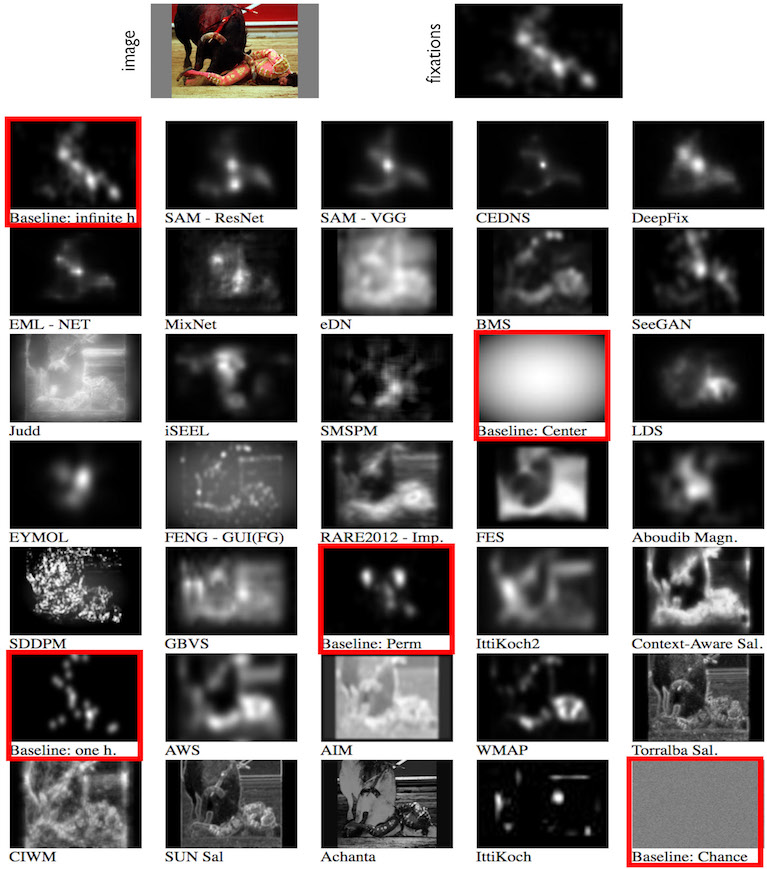} \\
    \vspace*{-7pt}
	\caption{Saliency prediction maps of 30 models (5 are baselines; enclosed with the red boxes) on the sample image at the MIT saliency benchmark for the CAT2000 dataset.}
	\label{fig:CATmaps}
\end{figure*}

Fig.~\ref{fig:CATmaps} illustrates predictions of saliency models over the sample image from the CAT2000 dataset on the MIT benchmark website. \\

\begin{figure*}
	\centering
    \includegraphics[width=.8\linewidth, angle =0]{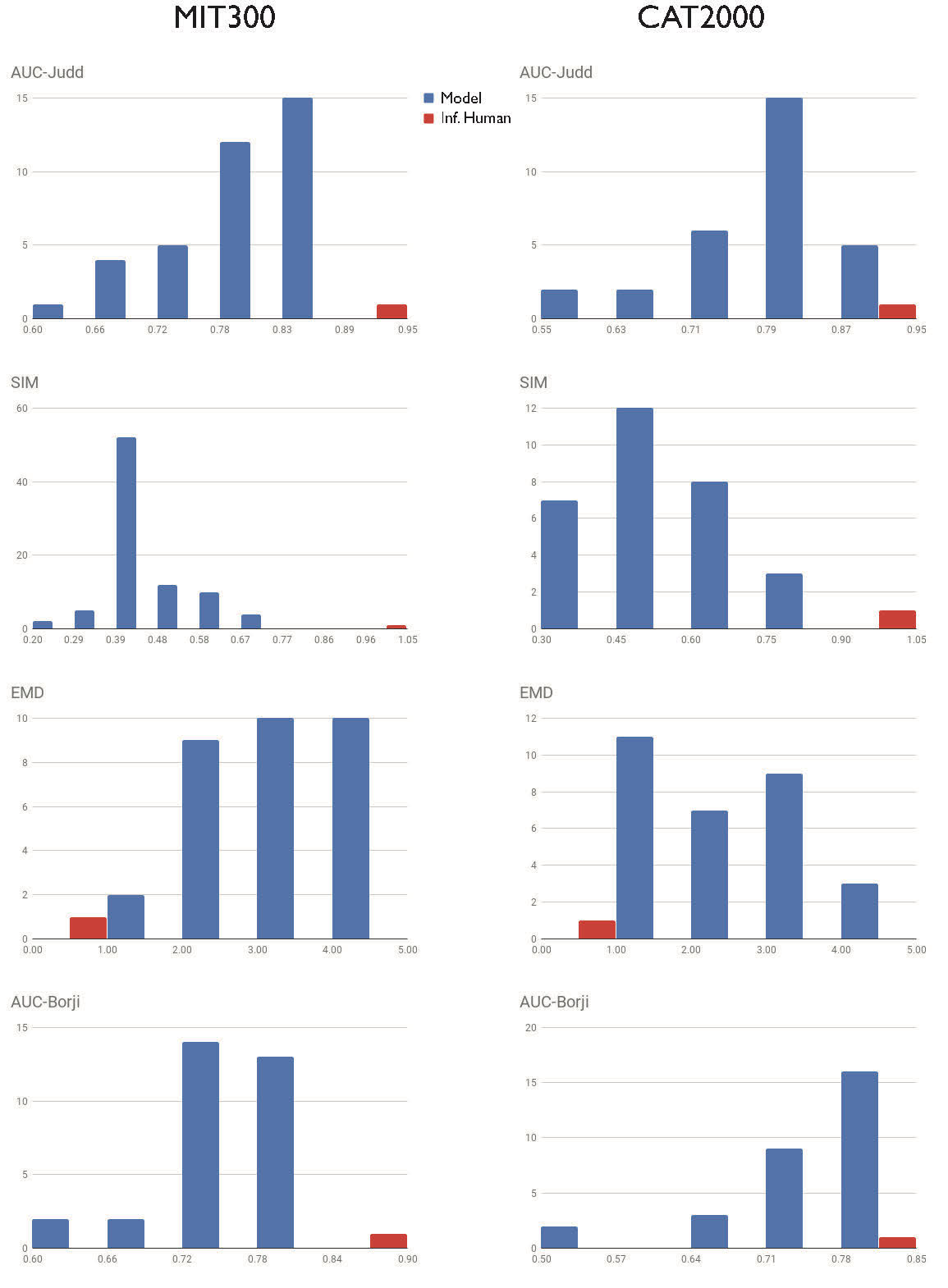} \\
	\caption{Histogram of model scores over both MIT300 and CAT2000 datasets. The red bar represents the human inter-observer (IO) model. As it stands, there is still a big gap between performance of models and humans across all scores. Further, several models share the same high score hinting towards performance saturation.}
	\label{fig:hists-MIT}
\end{figure*}

\begin{figure*}
	\centering
    \includegraphics[width=.8\linewidth,angle =0]{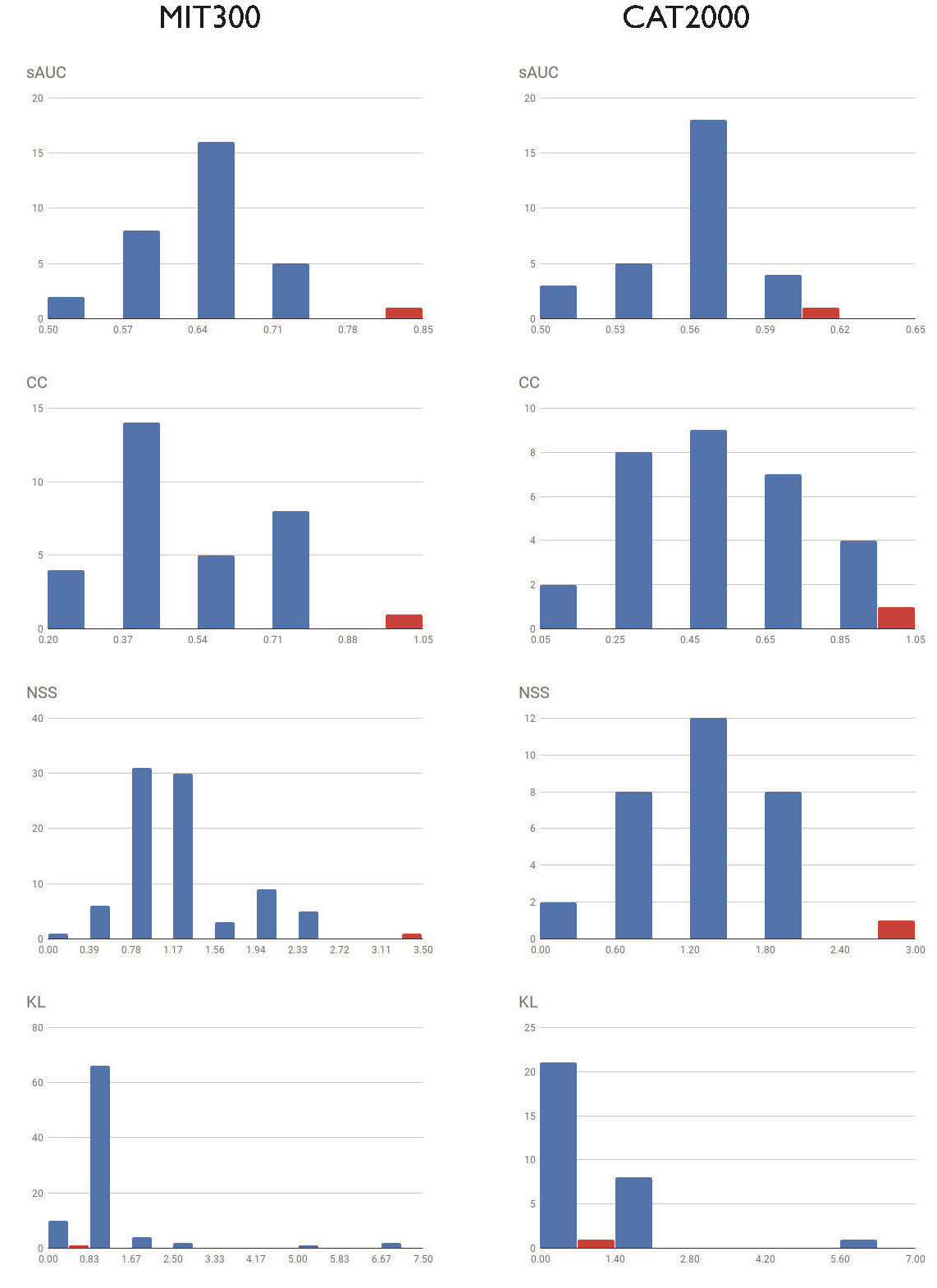} \\
	\caption{Histogram of model scores over both MIT300 and CAT2000 datasets. The red bar represents the human inter-observer (IO) model. As it stands, there is still a big gap between performance of models and humans across all scores. Further, several models share the same high score hinting towards performance saturation.}
	\label{fig:hists-MIT2}
\end{figure*}

Figs.~\ref{fig:hists-MIT} \&~\ref{fig:hists-MIT2} show the histogram of model scores over both MIT300 and CAT2000 datasets. The red bar represents the human inter-observer (IO) model. As it stands, there is still a big gap between performance of models and humans across all scores. Further, several models share the same high score hinting towards performance saturation.\\

\begin{figure}
	\centering
    \includegraphics[width=1\linewidth,height = 10cm, angle =0]{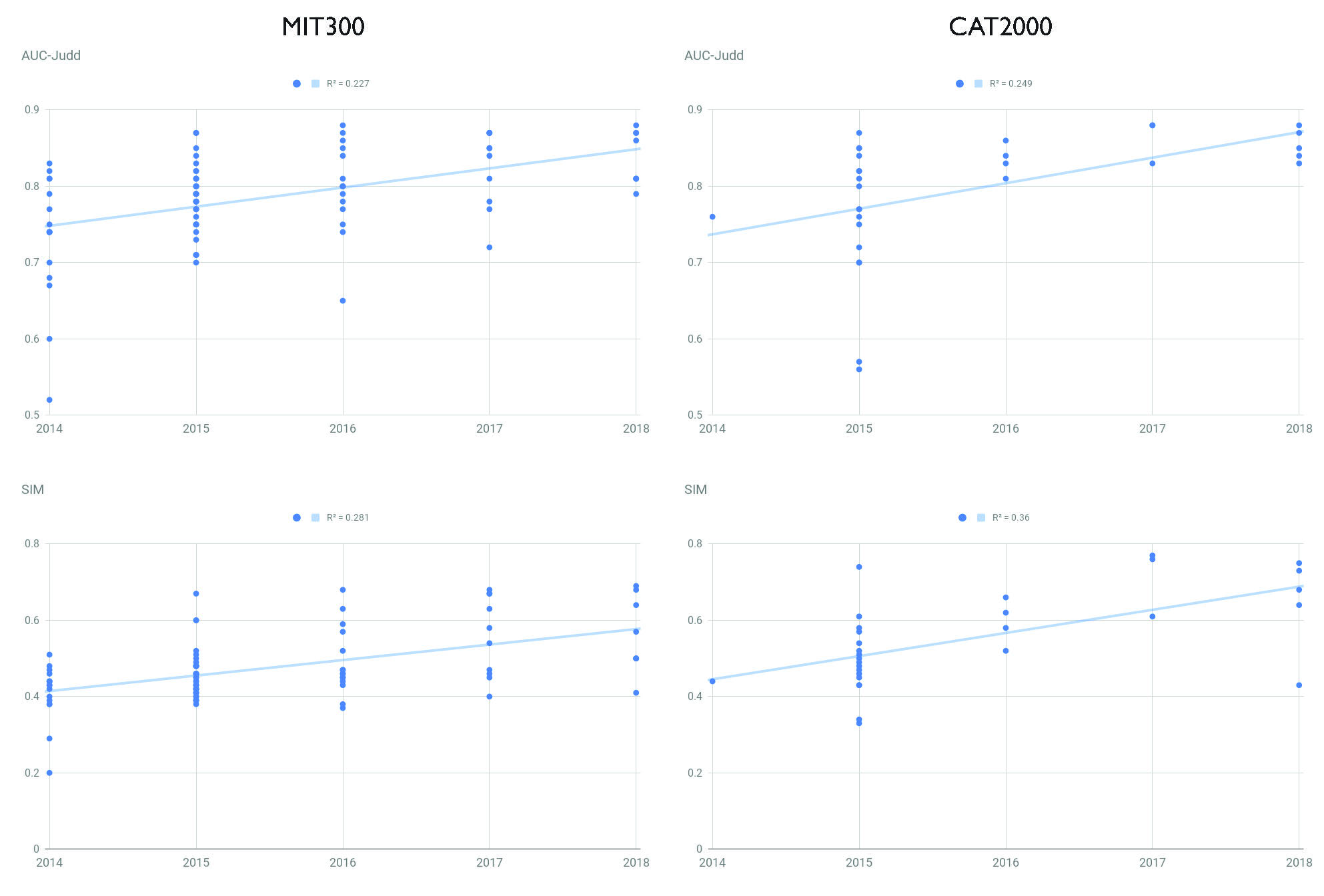} \\
	\caption{Performance improvement over time (from 2014-2018) over the MIT300 and CAT2000 datasets. Data shows that there is an increasing trend in performance (see the $R^2$ values above the plots).}
	\label{fig:progress-both-1}
\end{figure}

\begin{figure}
	\centering
    \includegraphics[width=1\linewidth,height = 10cm, angle =0]{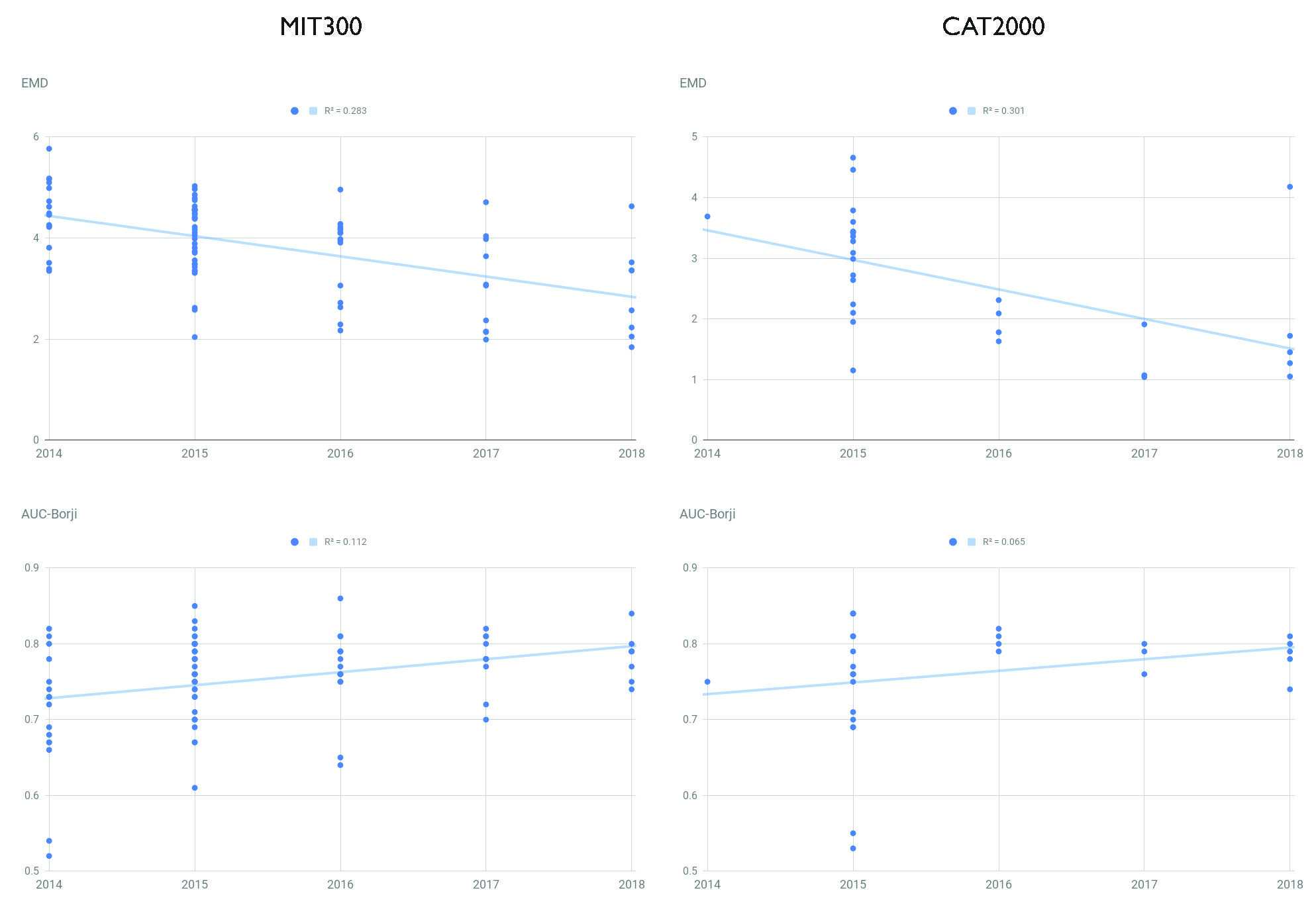} \\
	\caption{Performance improvement over time (from 2014-2018) over the MIT300 and CAT2000 datasets. Data shows that there is an increasing trend in performance (see the $R^2$ values above the plots).}
	\label{fig:progress-both-2}
\end{figure}

\begin{figure}
	\centering
    \includegraphics[width=1\linewidth,height = 10cm, angle =0]{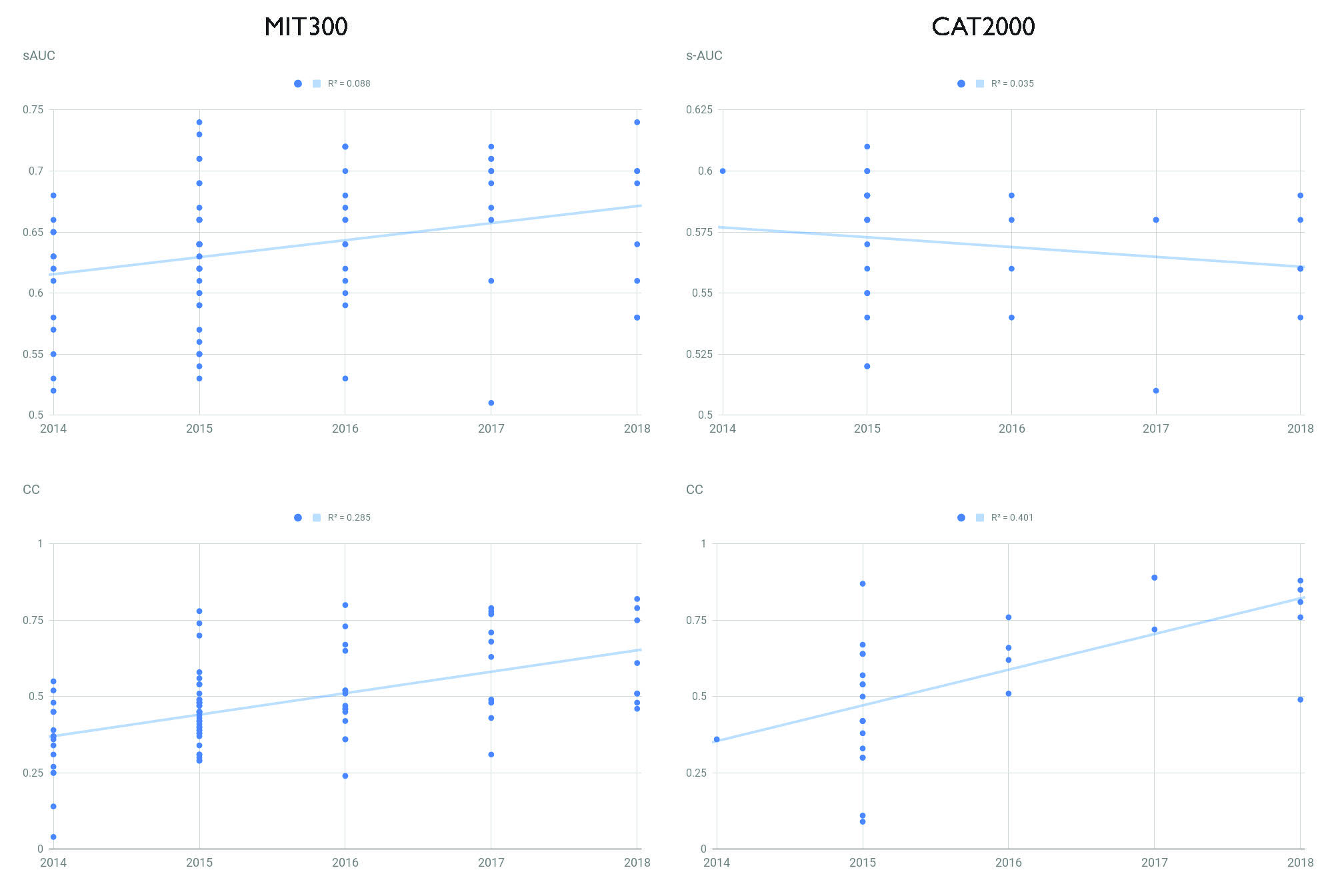} \\
	\caption{Performance improvement over time (from 2014-2018) over the MIT300 and CAT2000 datasets. Data shows that there is an increasing trend in performance (see the $R^2$ values above the plots).}
	\label{fig:progress-both-3}
\end{figure}

\begin{figure}
	\centering
    \includegraphics[width=1\linewidth,height = 10cm, angle =0]{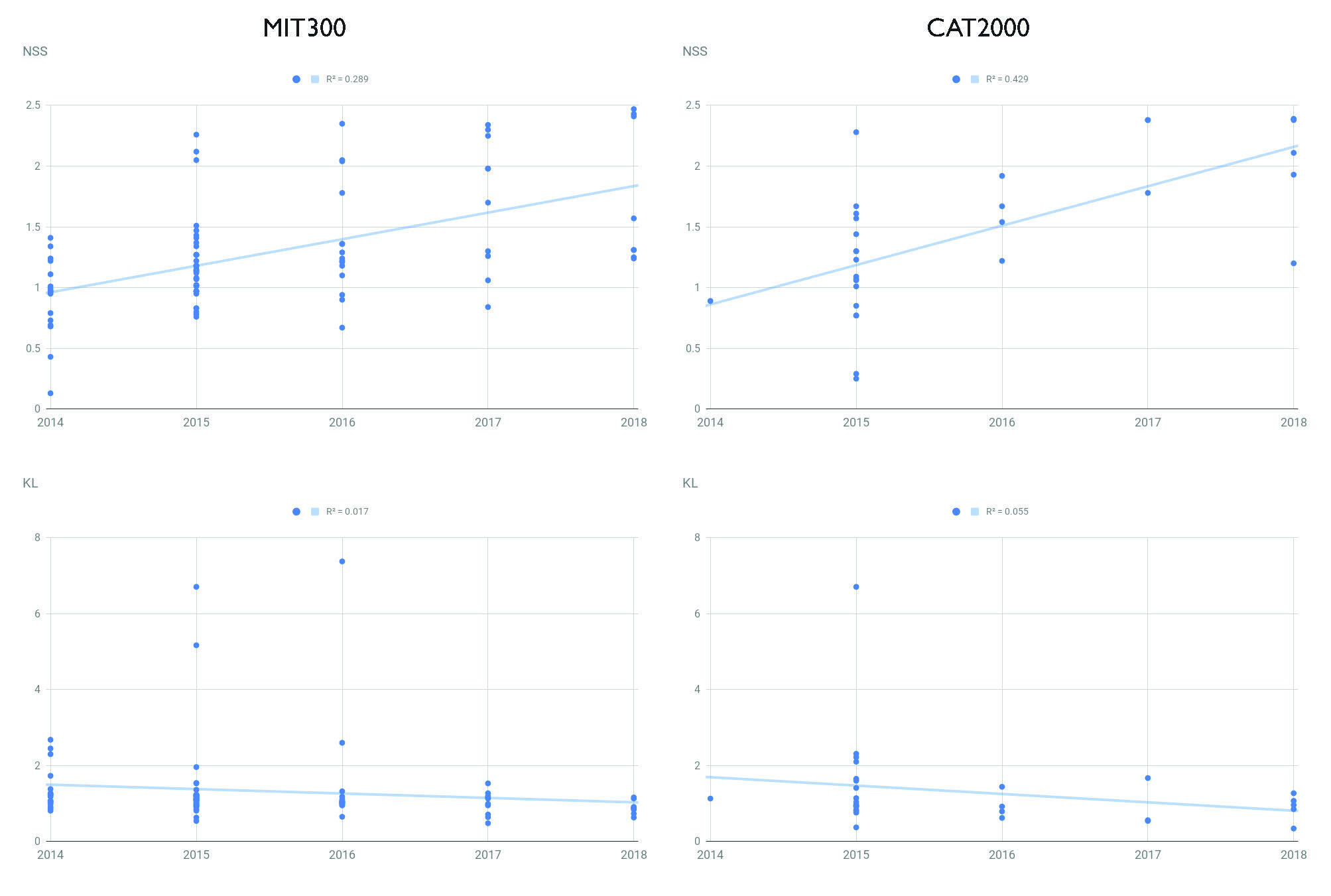} \\
	\caption{Performance improvement over time (from 2014-2018) over the MIT300 and CAT2000 datasets. Data shows that there is an increasing trend in performance (see the $R^2$ values above the plots).}
	\label{fig:progress-both-4}
\end{figure}

Figs.~\ref{fig:progress-both-1} \&~\ref{fig:progress-both-2} \&~\ref{fig:progress-both-3} \&~\ref{fig:progress-both-4} show the performance improvement over time (from 2014-2018) for each of the 8 scores over the MIT300 and CAT2000 datasets. Data shows that there is an increasing trend in performance (see the $R^2$ values above the plots).\\

\begin{figure*}
	\centering
    \includegraphics[width=.9\linewidth, angle =0]{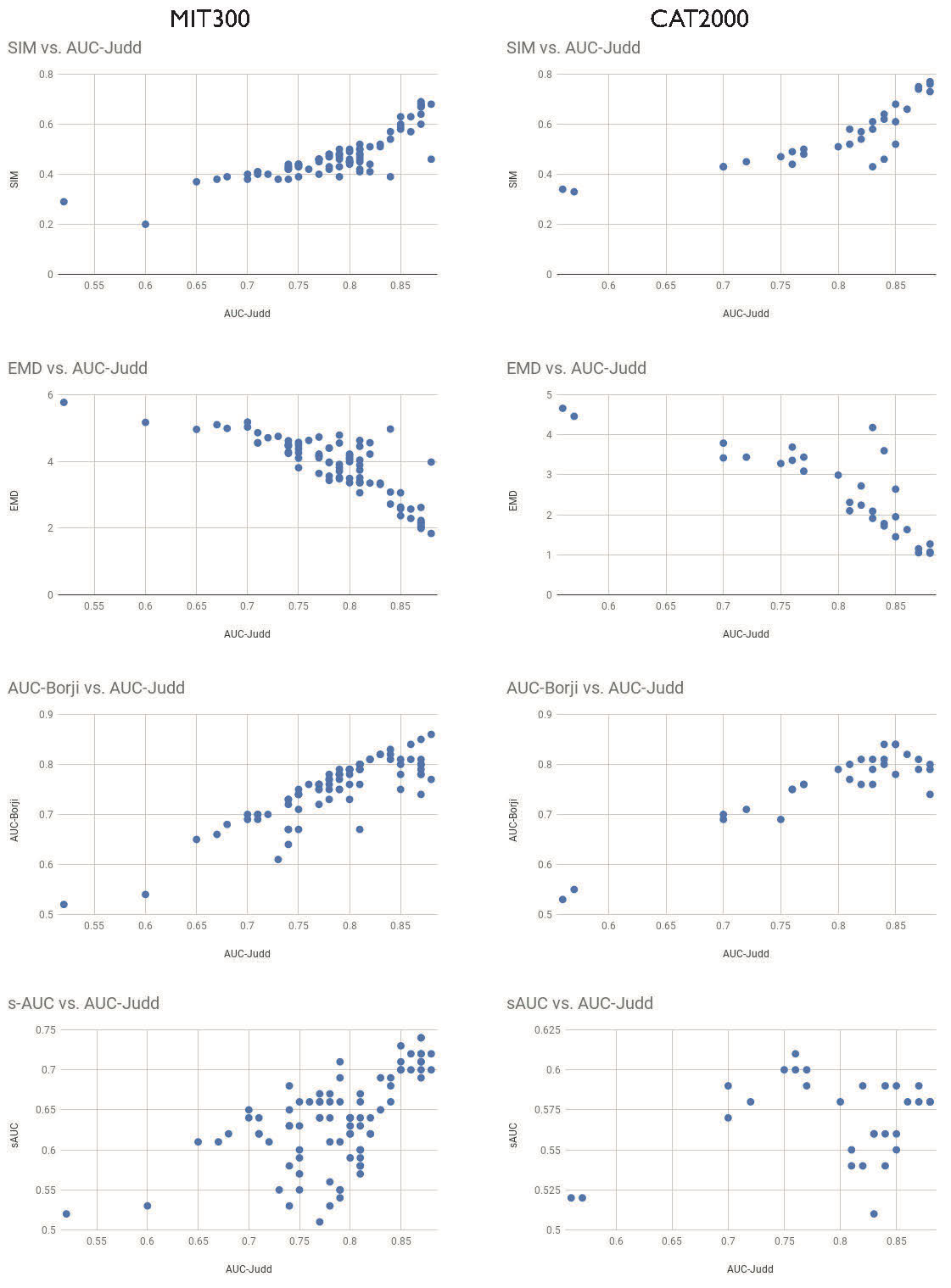} \\
	\caption{Plot of scores versus each other over the MIT300 and CAT2000 datasets. Each dot represents one model. There is a positive correlation between some scores (\eg SIM and AUC-Judd; AUC-Borji and AUC-Judd), whereas for some others the correlation is negative (\eg KL and AUC-Judd). }
	\label{fig:cross-Scores}
\end{figure*}

\begin{figure*}
	\centering
    \includegraphics[width=.9\linewidth, angle =0]{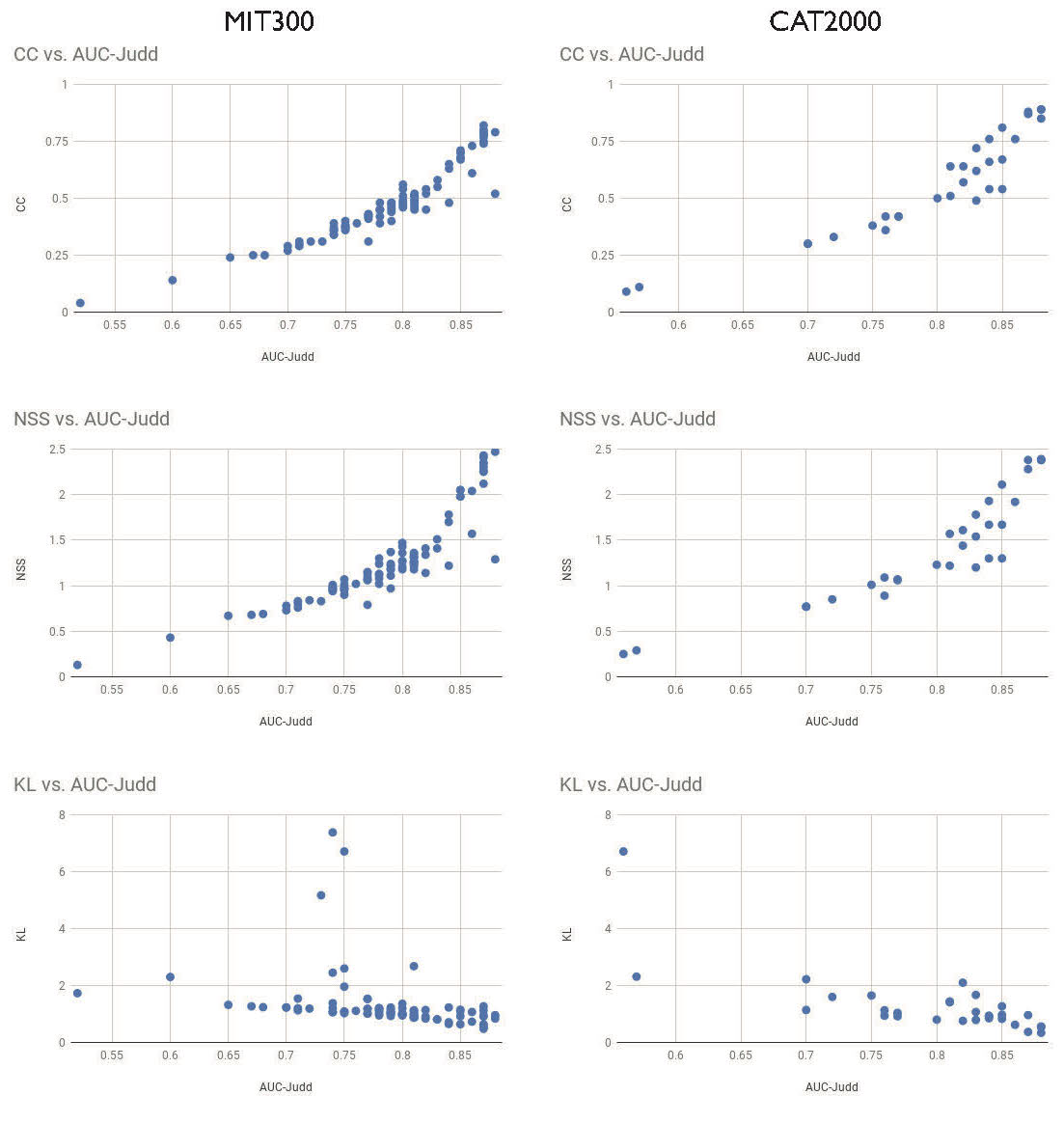} \\
	\caption{Plot of scores versus each other over the MIT300 and CAT2000 datasets. Each dot represents one model. There is a positive correlation between some scores (\eg SIM and AUC-Judd; AUC-Borji and AUC-Judd), whereas for some others the correlation is negative (\eg KL and AUC-Judd).}
	\label{fig:cross-Scores2}
\end{figure*}

Figs.~\ref{fig:cross-Scores} \&~\ref{fig:cross-Scores2} show the plot of scores versus each other over the MIT300 and CAT2000 datasets. Each dot represents one model. There is a positive correlation between some scores (\eg SIM and AUC-Judd; AUC-Borji and AUC-Judd), whereas for some others the correlation is negative (\eg KL and AUC-Judd). \\

\begin{figure*}
	\centering
    \includegraphics[width=.9\linewidth, angle =0]{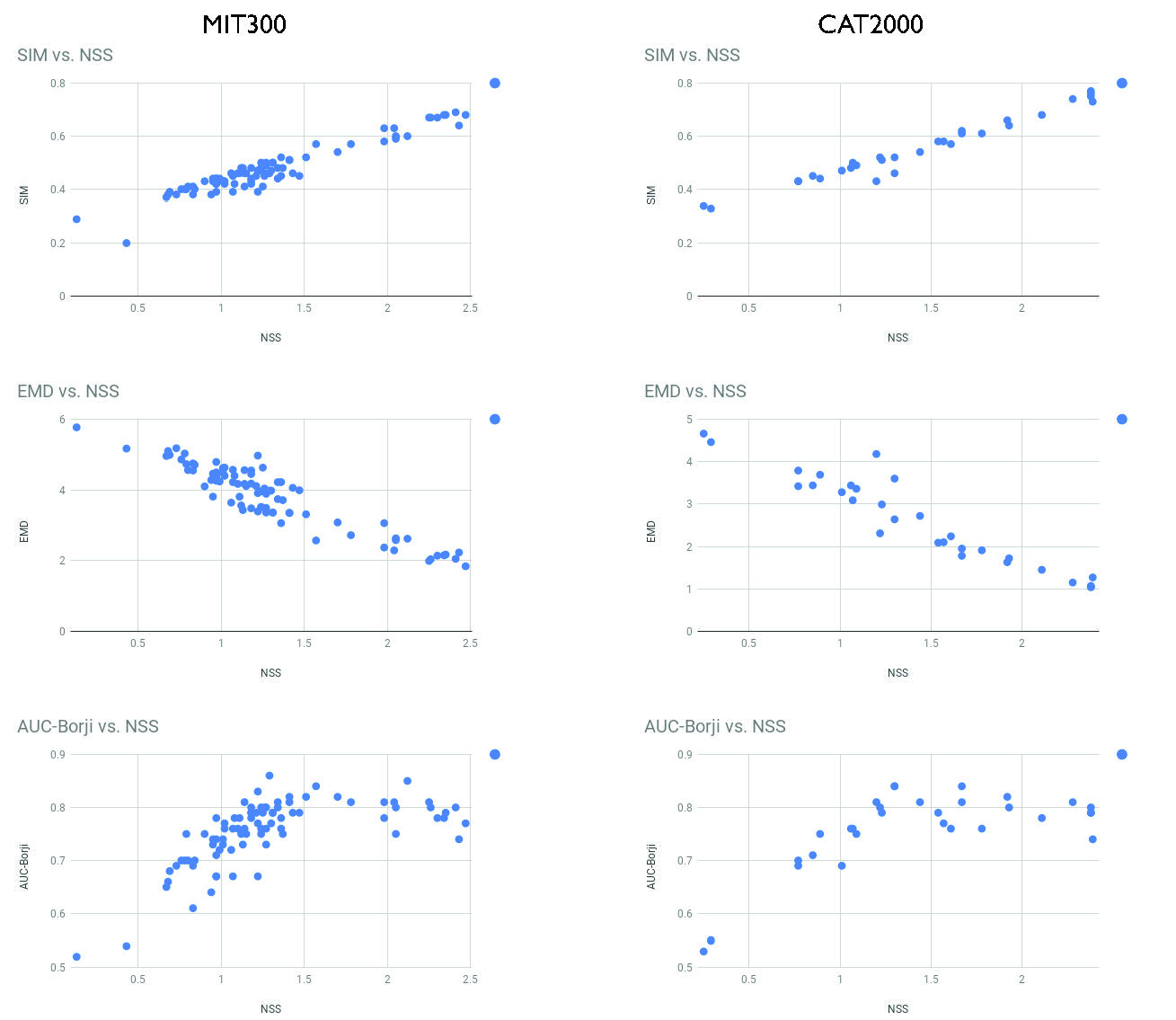} \\
	\caption{Correlations between scores versus the NSS score over MIT300 and CAT2000.}
	\label{fig:cross-Scores-NSS-1}
\end{figure*}

\begin{figure*}
	\centering
    \includegraphics[width=.9\linewidth, angle =0]{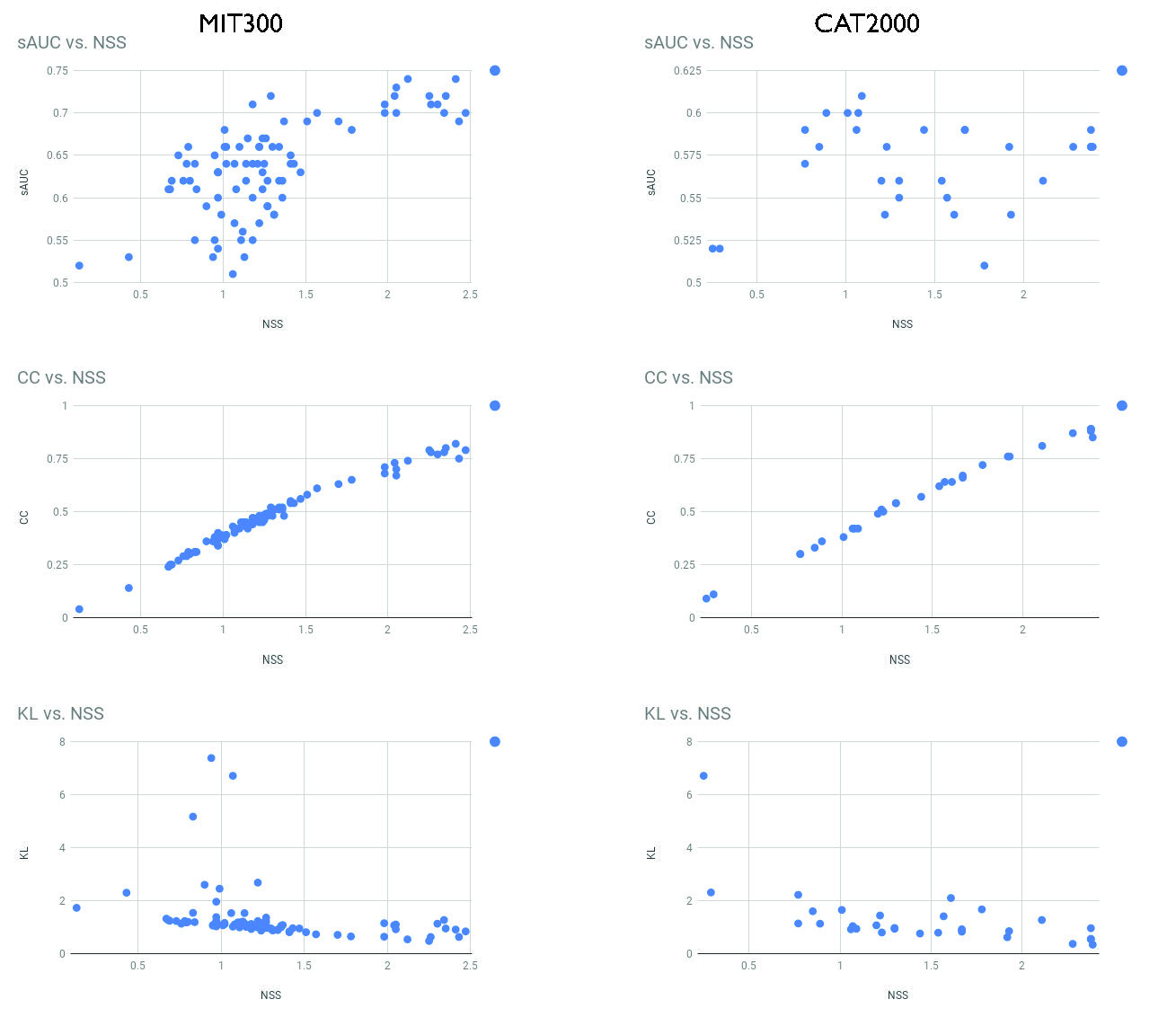} \\
	\caption{Correlations between scores versus the NSS score over MIT300 and CAT2000.}
	\label{fig:cross-Scores-NSS-2}
\end{figure*}

Figs.~\ref{fig:cross-Scores-NSS-1} \&~\ref{fig:cross-Scores-NSS-2} show the correlations between scores versus the NSS score.

\begin{figure*}
	\centering
    \includegraphics[width=.9\linewidth]{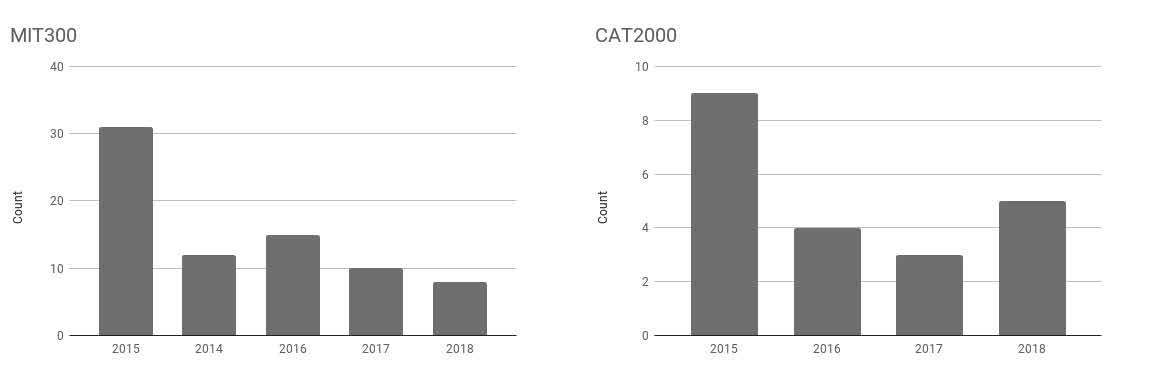} \\
	\caption{The number of models published per year. The number of submissions are declining over MIT300, but are increasing over the CAT2000 (As of Oct. 2018).}
	\label{fig:MIT300hist}
\end{figure*}

Fig.~\ref{fig:MIT300hist} illustrates the number of models published per year. The number of submissions are declining over MIT300, but are increasing over the CAT2000 (As of Oct. 2018).\\

\begin{figure*}
	\centering
    \includegraphics[width=1\linewidth]{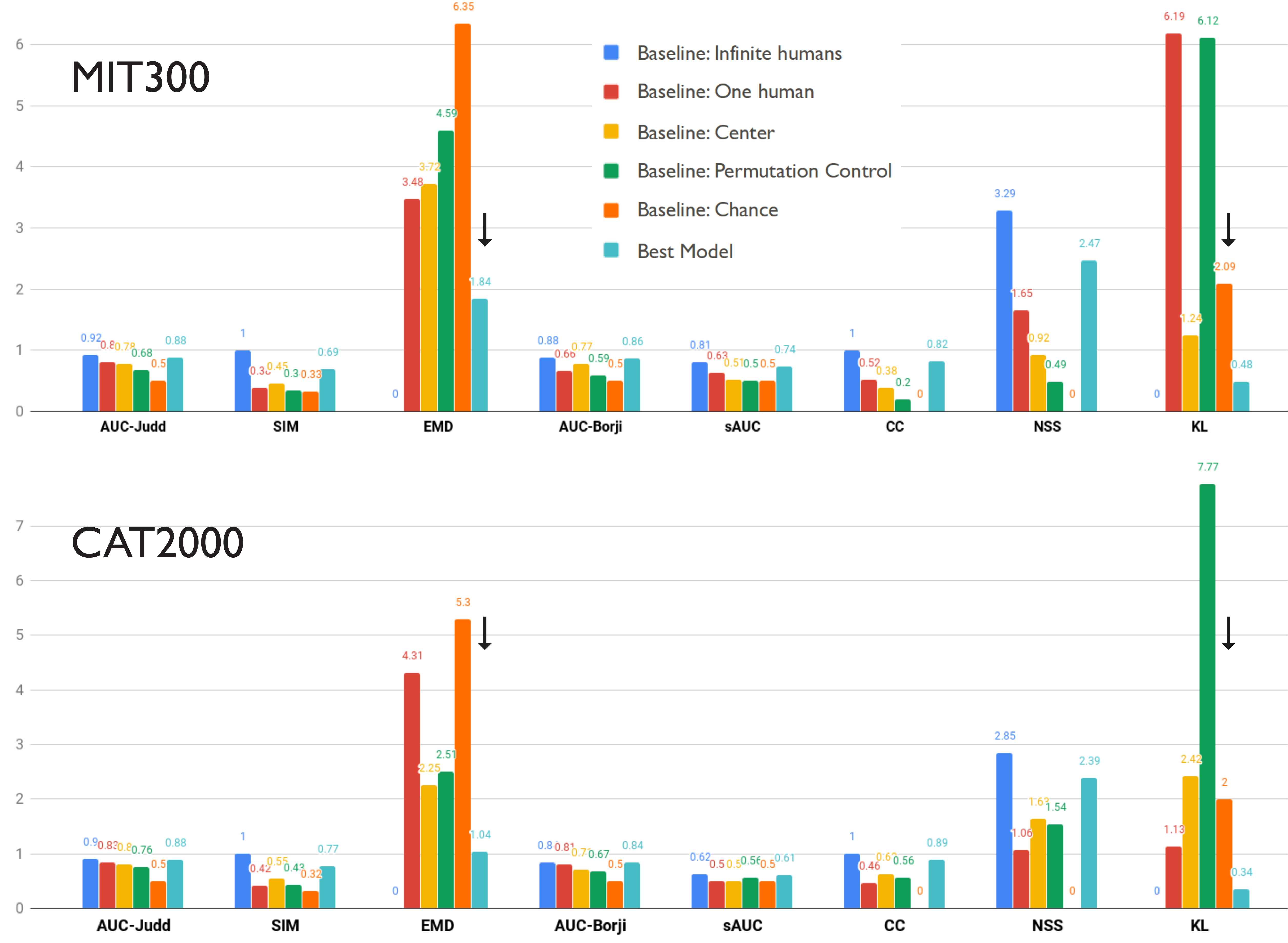} \\
	\caption{Scores of five baselines and the best saliency model (can be different for each score). As it can be seen the best model often wins over the baselines except the Inf. human baseline. This is the case over both MIT300 and CAT2000 datasets.}
	\label{fig:charts}
\end{figure*}

Fig.~\ref{fig:charts} shows the scores of five baselines and the best saliency model (can be different for each score). As it can be seen the best model often wins over the baselines except the Inf. human baseline. This is the case over both MIT300 and CAT2000 datasets.\\

\begin{figure*}
    \includegraphics[width=1\linewidth,height = 20.5cm, angle =0]{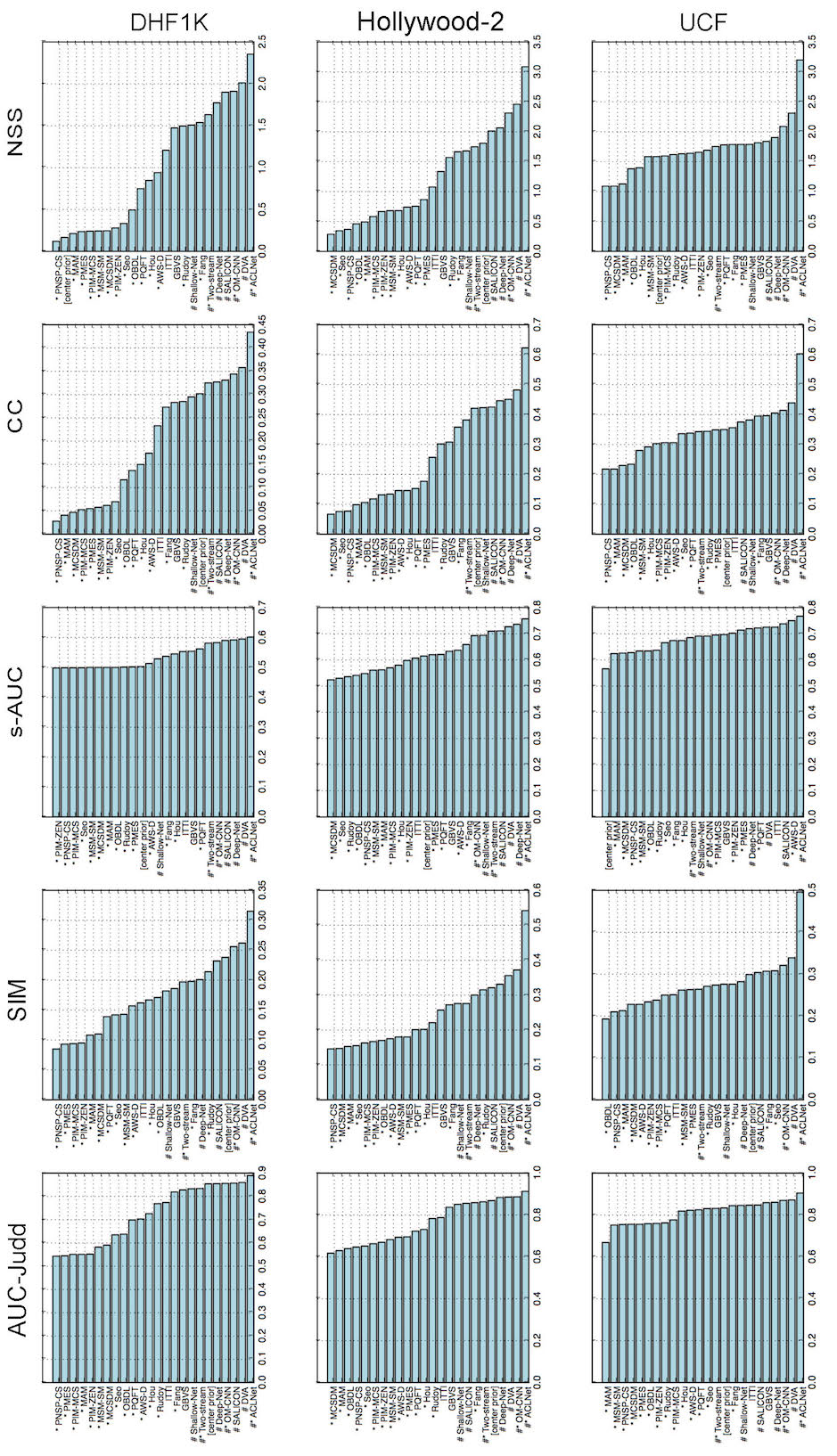} \\
	\caption{A comparison of static and dynamic, deep and non-deep, saliency models over the DHF1K~\cite{wang2018revisiting}, UCF sports~\cite{mathe2015actions}, and Hollywood-2~\cite{mathe2015actions} datasets. Models using the motion feature are marked with `*'. Deep models are marked with `\#'. Results are compiled from~\cite{wang2018revisiting}.}
	\label{fig:DHF1KNew}
\end{figure*}

Fig.~\ref{fig:DHF1KNew} shows a comparison of static and dynamic, deep and non-deep, saliency models over the DHF1K~\cite{wang2018revisiting}, UCF sports~\cite{mathe2015actions}, and Hollywood-2~\cite{mathe2015actions} datasets. Models using the motion feature are marked with `*'. Deep models are marked with `\#'. Results are compiled from~\cite{wang2018revisiting}. \\

\begin{figure*}
    \includegraphics[width=1\linewidth,height = 17cm, angle =0]{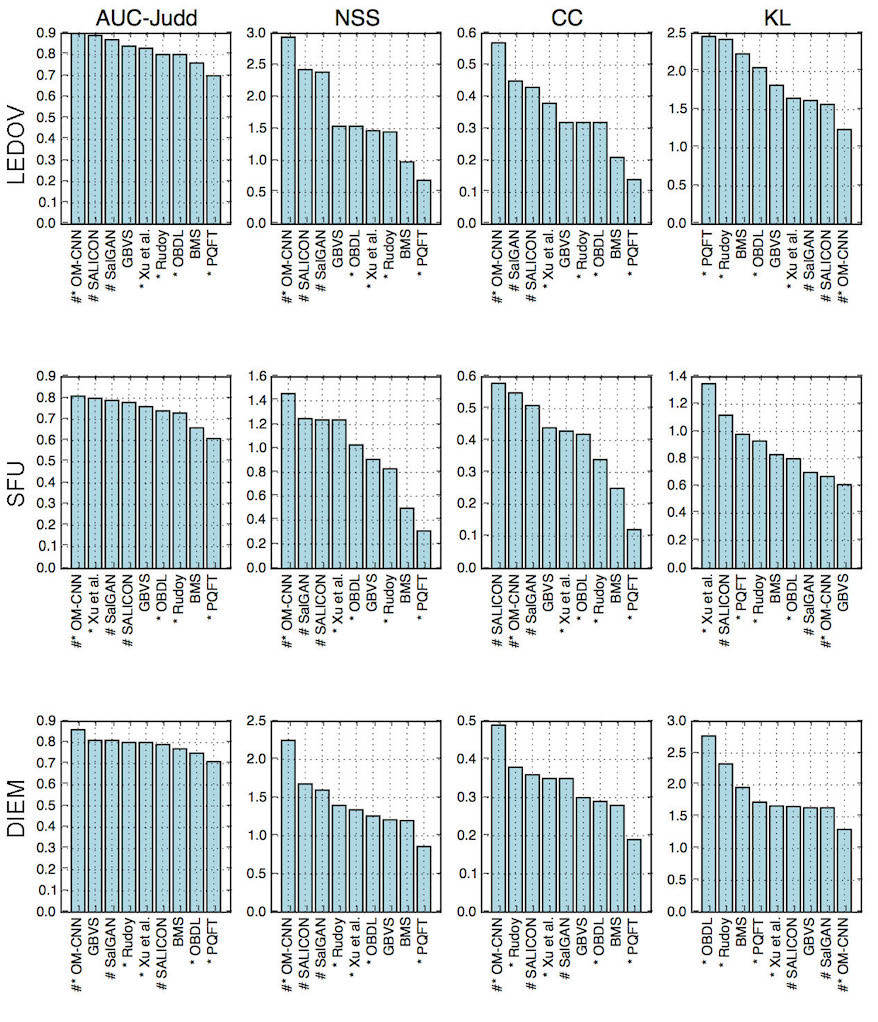} \\
	\caption{Performance of models over the LEDOV~\cite{jiang2018deepvs}, SFU~\cite{hadizadeh2012eye}, and
DIEM~\cite{mital2011clustering} datasets. Xu~\etal is the model in~\cite{xu2017learning}. Models using the motion feature are marked with `*'. Deep models are marked with `\#'. Results are compiled from~\cite{jiang2018deepvs}.}
	\label{fig:ledov}
\end{figure*}

Fig.~\ref{fig:ledov} shows performance of models over the LEDOV~\cite{jiang2018deepvs}, SFU, and
DIEM~\cite{mital2011clustering} datasets. Models using the motion feature are marked with `*'. Deep models are marked with `\#'. Results are compiled from~\cite{jiang2018deepvs}.\\

\begin{figure*}
	\centering
    \includegraphics[width=\linewidth]{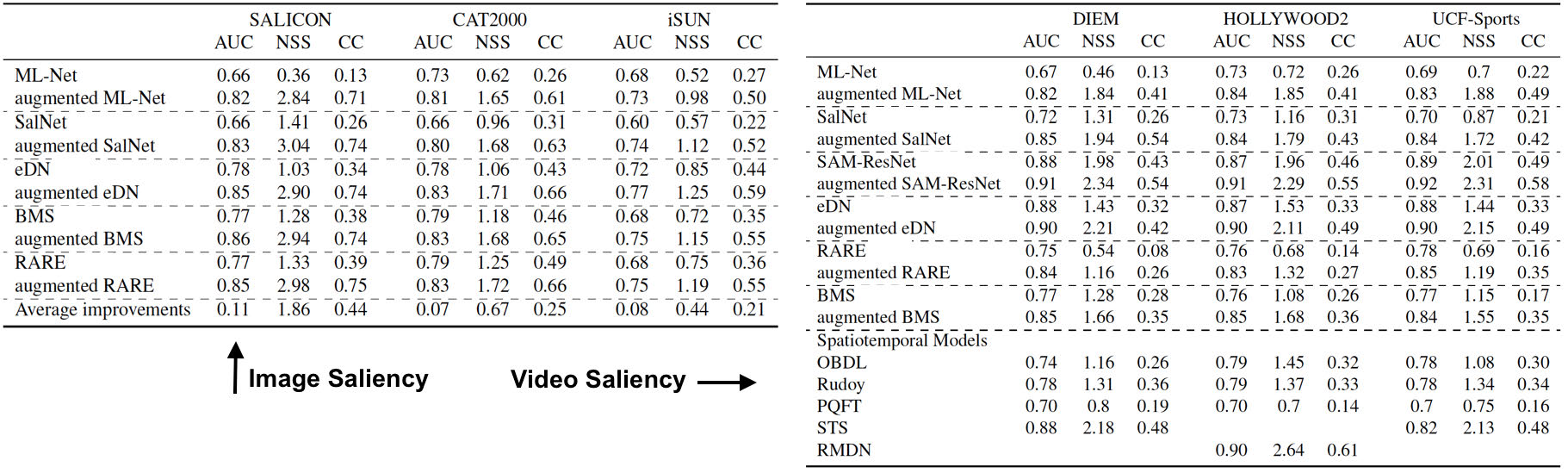} 
	\caption{Performance of the two models by Gorji and Clark~\cite{gorji2017attentional,gorji2018going}. The left table shows results of augmenting static deep models for image saliency prediction. The right table shows results of the augmented dynamic saliency model for video fixation prediction.}
	\label{fig:gorjiResults}
\end{figure*}

Fig.~\ref{fig:gorjiResults} shows the results of the two models by Gorji and Clark~\cite{gorji2017attentional,gorji2018going}. The left table shows results of augmenting static deep models for image saliency prediction. The right table shows results of the augmented dynamic saliency model for video fixation prediction.\\

\begin{figure*}
	\centering
    \includegraphics[width=\linewidth]{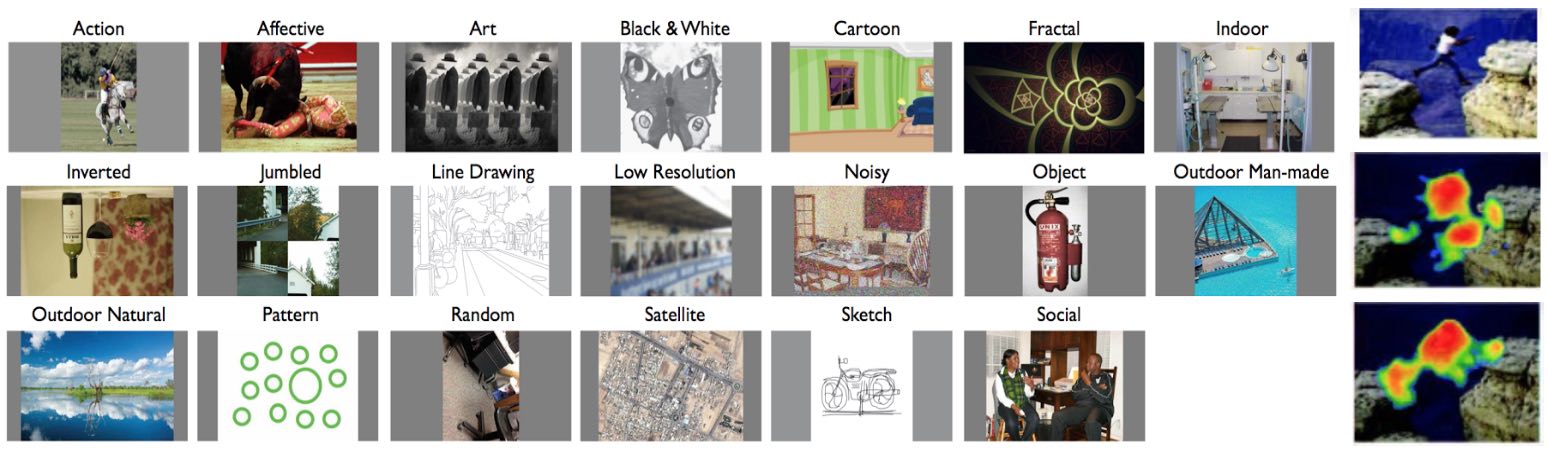} 
	\caption{Example stimuli from different categories of the CAT2000 dataset~\cite{Borji_Itti15arxiv}. The right panel shows an example image where future foot placement seems to be guiding the fixations (middle panel), but a saliency model fails to predict gaze (bottom panel). Image borrowed from~\cite{bruce2016deeper}.}
	\label{fig:CAT}
\end{figure*}

Fig.~\ref{fig:CAT} shows example stimuli from different categories of the CAT2000 dataset~\cite{Borji_Itti15arxiv}. The right panel shows an example image where future foot placement seems to be guiding the fixations (middle panel), but a saliency model fails to predict gaze (bottom panel). Image borrowed from~\cite{bruce2016deeper}.\\

\begin{figure*}
\centering
%
    \includegraphics[width=\linewidth]{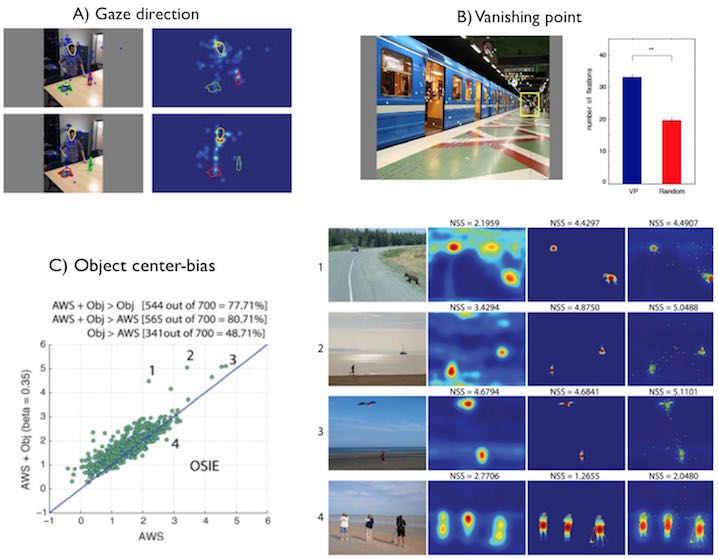} 
	\caption{Some cues that attract eye movements. A) an image pair with corresponding fixation maps made from saccades leaving the head region. An actor was explicitly instructed to look at one of the two objects. B) Left: Example stimuli with annotated vanishing point (VP; yellow box) and fixations (dots), Right: Histogram of saliency at VP vs. random windows, C) Left: NSS score of \textit{AWS saliency}~\cite{garcia2012saliency} + object center bias versus \textit{AWS alone}, Right: Sample images and corresponding prediction maps of the models. }
	\label{fig:cues}
\end{figure*}

Fig.~\ref{fig:cues} shows some cues that attract eye movements in free viewing of natural scenes, compiled from our previous studies including gaze direction~\cite{Borji_etal14jov,Parks_etal15vr}, vanishing point~\cite{borji2016vanishing}, and object center-bias~\cite{borji2016reconciling}. \\

\begin{figure*}
	\centering
    \includegraphics[width=1.\linewidth, angle =0]{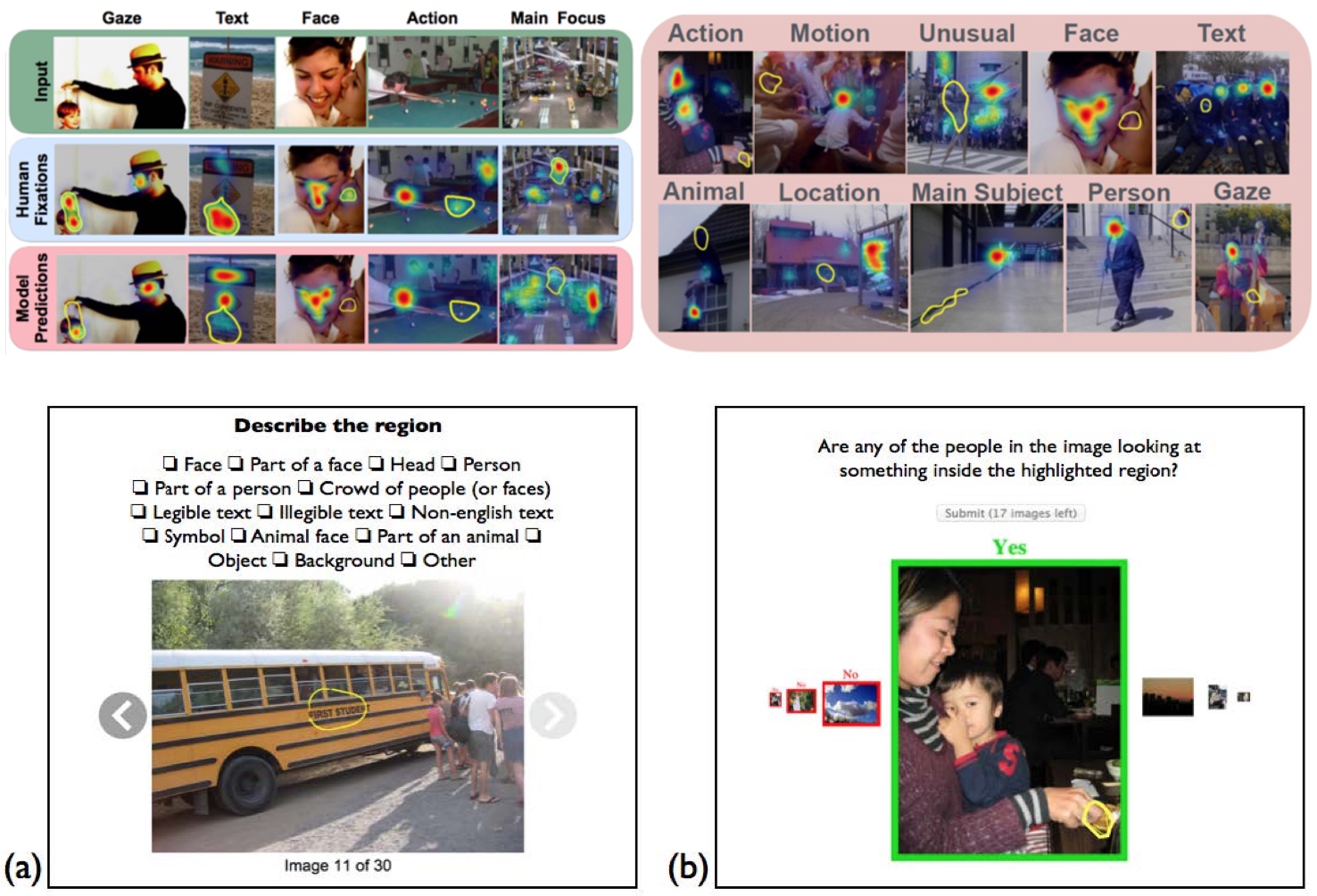} \\
	\caption{ Top: A finer-grained analysis reveals where models can still make significant improvements. High-density regions of human fixations are marked in yellow, and show that models continue to miss these semantically-meaningful elements. A large portion of model errors fall into these categories: action, motion, unusual, face, text, animal, location, main subject, person, and gaze direction. Bottom: Two types of Mechanical Turk tasks were used for gathering annotations for the highly-fixated regions in an image. These annotations were then used to quantify where people look in images. See~\cite{BylinskiiECCV2016} for details.}
	\label{fig:newZoya}
\end{figure*}

Fig.~\ref{fig:newZoya} shows cases where saliency models miss regions that attract human gaze. A large portion of model errors fall into these categories: action, motion, unusual, face, text, animal, location, main subject, person, and gaze direction. It also shows the types of Mechanical Turk tasks that were used for gathering annotations. Please see~\cite{BylinskiiECCV2016} for details. \\

\begin{figure*}[]
	\centering
    \includegraphics[width=\linewidth]{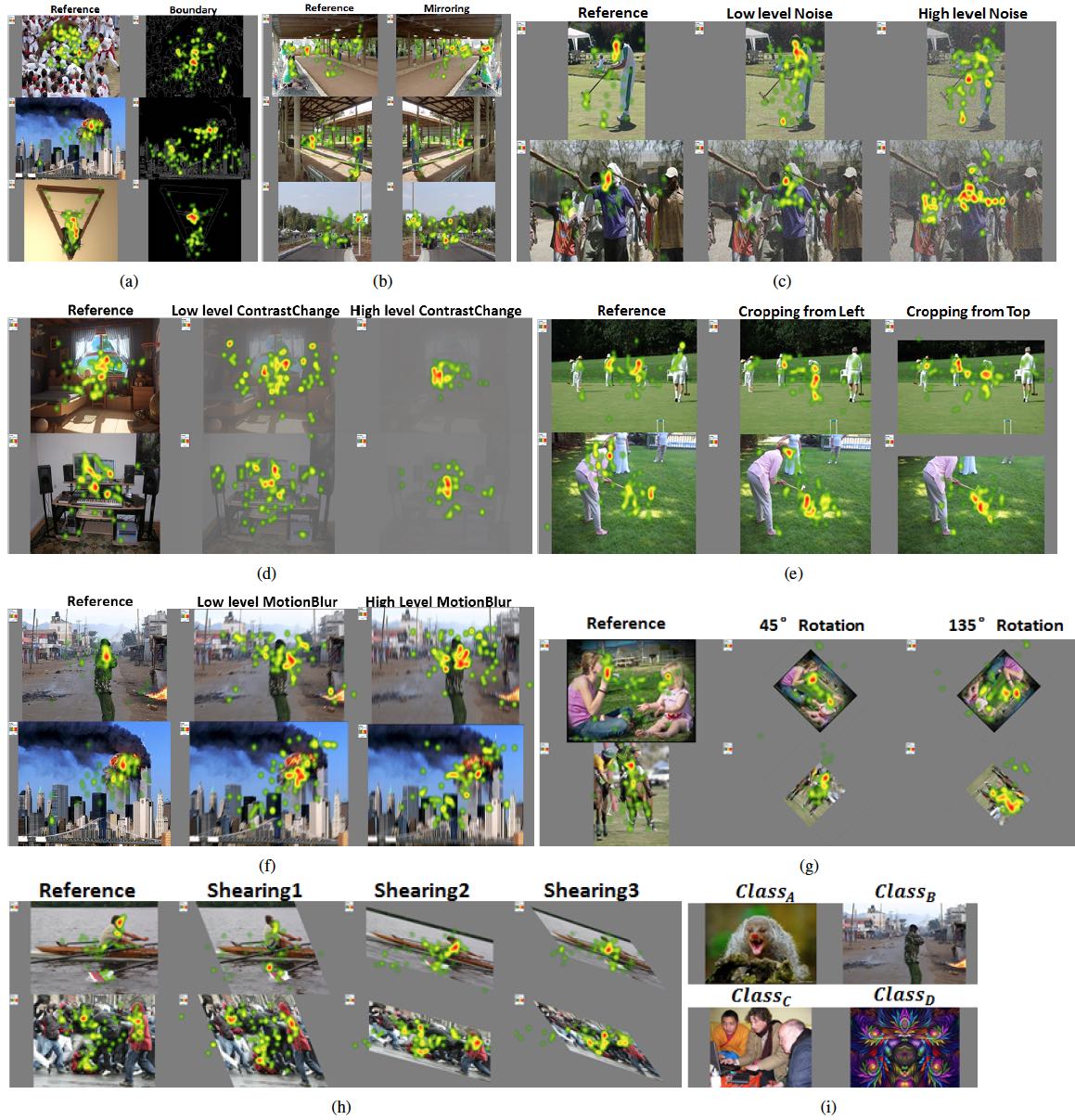} \\
    \vspace*{-7pt}
	\caption{Example stimuli under a number of image transformations such as shear, contrast change, adding noise, rotation, cropping, etc. See~\cite{invariance} for details.}
	\label{fig:invariance2}
\end{figure*}

Fig.~\ref{fig:invariance2} illustrates example stimuli under a number of transformations such as shear, contrast change, adding noise, rotation, cropping, etc. See~\cite{invariance} for details.\\

\begin{figure*}
	\centering
    \includegraphics[width=\linewidth]{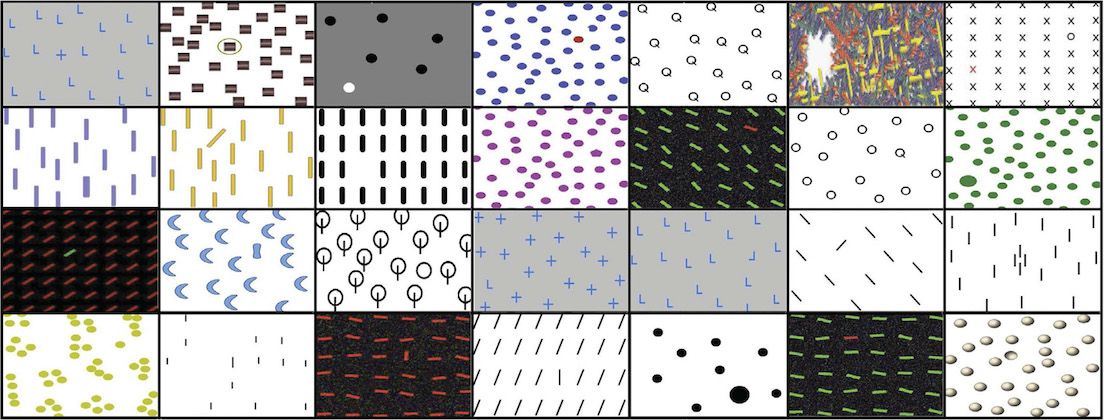} \\
    \vspace*{-7pt}
	\caption{Example pop out search arrays~\cite{borji2013quantitative}. These patterns can be used to test whether deep saliency models can capture saliency due to low-level feature contrast.}
	\label{fig:patterns}
\end{figure*}

Fig.~\ref{fig:patterns} shows example pop out search arrays~\cite{borji2013quantitative,borji2013state}. These patterns can be used to test whether deep saliency models can capture saliency due to low-level feature contrast.\\

\begin{figure*}
	\centering
    \includegraphics[width=\linewidth]{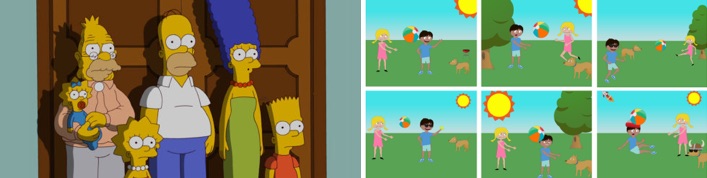} \\
    \vspace*{-7pt}
	\caption{Example images from the Clipart dataset by Zitnick and Parikh~\cite{zitnick2013bringing}, and a sample frame from ``The Simpsons'' cartoon.}
	\label{fig:simpson}
\end{figure*}

Fig.~\ref{fig:simpson} shows example images from the Clipart dataset by Zitnick and Parikh~\cite{zitnick2013bringing}, and a sample frame from ``The Simpsons'' cartoon. \\

Fig.~\ref{fig:stim} shows sample images containing rich semantic information (referred to as the Extreme Dataset in the main text). These images can be used to discriminate the models in their ability to capture factors that guide gaze at the semantic level. \\

Fig.~\ref{fig:manipulation} illustrates example applications of saliency for image manipulation and enhancement, from Mechrez~\etal, Gatys \etal~\cite{gatys2017guiding}, and Theis \etal~\cite{theis2018faster}. \\


Fig.\ref{fig:collage} illustrates an example image collection (all in one image) that can be used to test how saliency models prioritize images with respect to human (\ie importance assignment). \\

\vspace{-10pt}




%

 -----------------------------------------------------

%
%
%

\begin{figure*}
    \includegraphics[width=\linewidth]{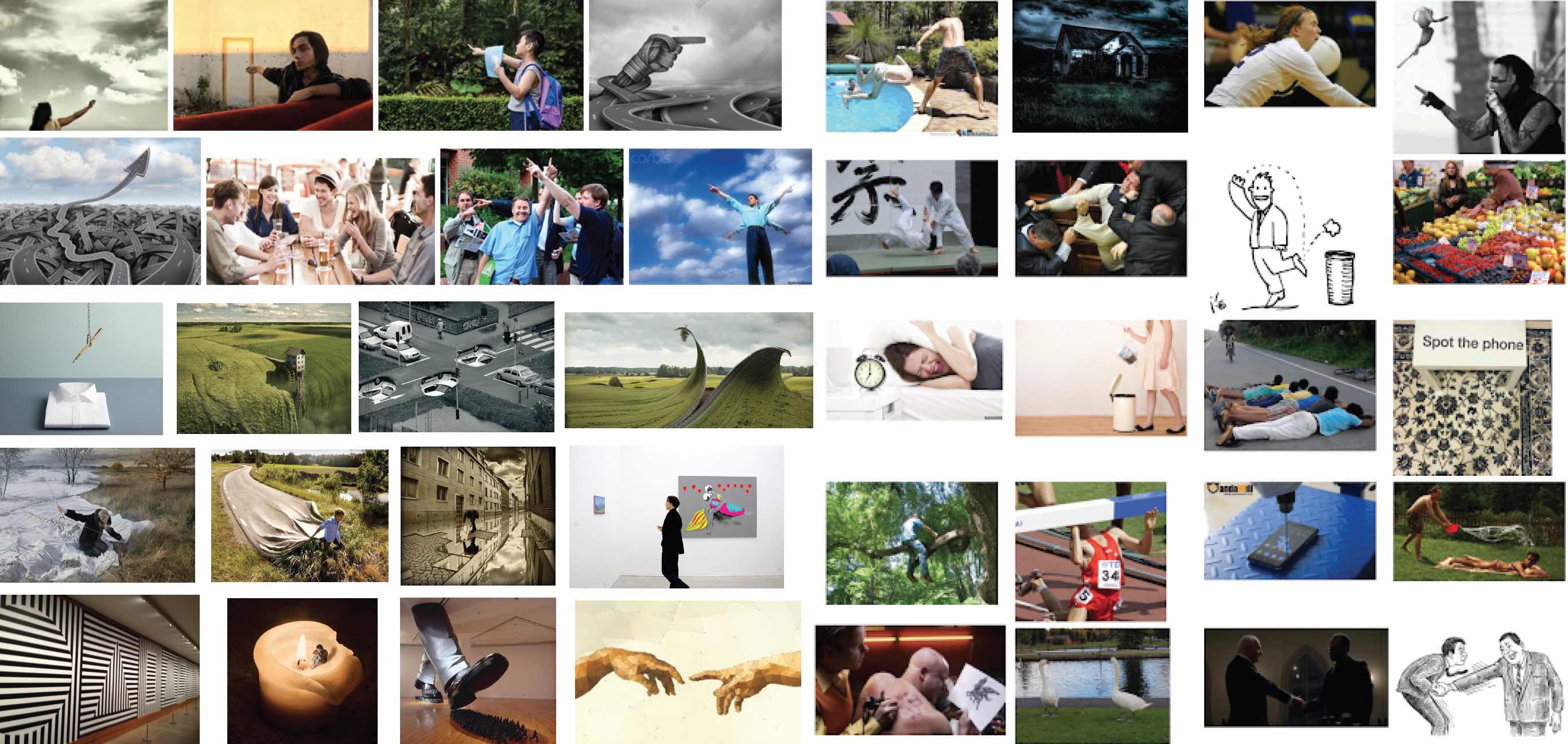} 
	\caption{Sample images containing rich semantic information (referred to as the Extreme Dataset in the main text). These images can be used to discriminate the models in their ability to capture factors that guide gaze at the semantic level.}
	\label{fig:stim}
\end{figure*}

%
%

\begin{figure*}
	\centering
    \includegraphics[width=1\linewidth, angle = 0]{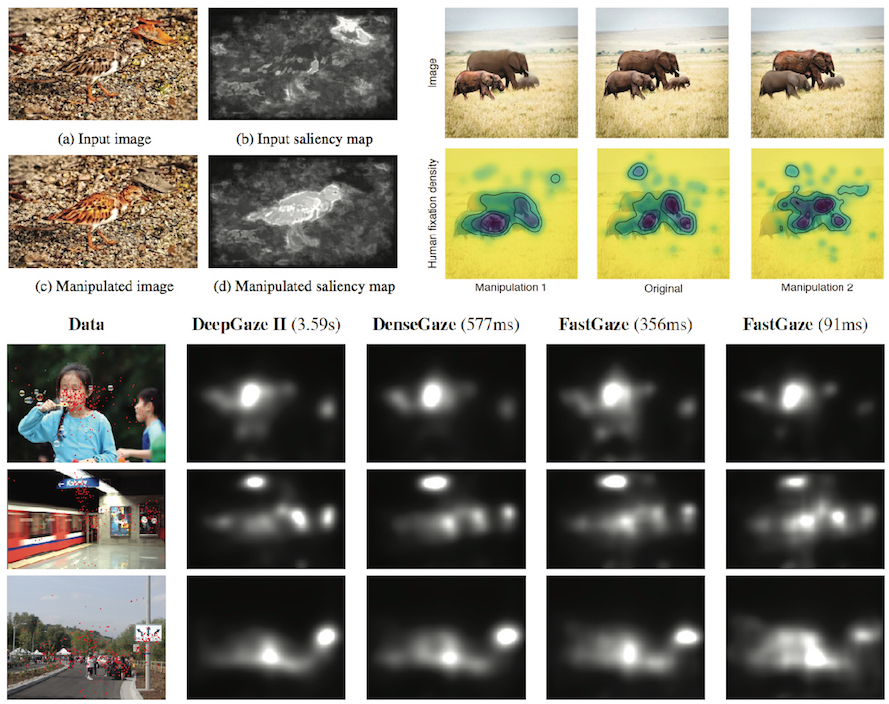} \\
	\caption{Example applications of saliency for image manipulation and enhancement, from Mechrez~\etal (top-left), Gatys \etal~\cite{gatys2017guiding} (top-right), and Theis \etal~\cite{theis2018faster} (bottom).}
	\label{fig:manipulation}
\end{figure*}


\begin{figure*}
	\centering
    \includegraphics[width=1\linewidth, angle = 0]{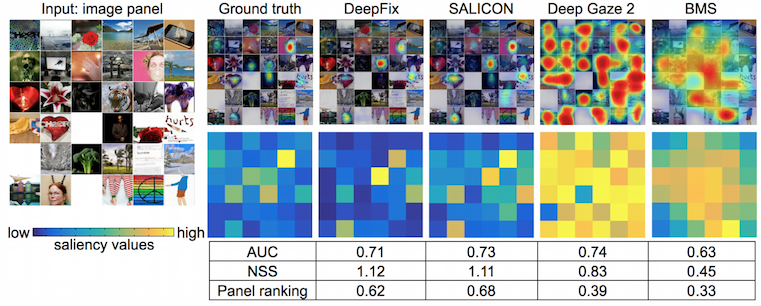} \\
	\caption{A finer-grained test proposed in~\cite{BylinskiiECCV2016} for determining how saliency models prioritize different sub-images in a panel, relative to each other. (left) A panel image from the MIT300 dataset. (right) The
saliency map predictions given the panel as an input image. The maximum response
of each saliency model on each subimage is visualized (as an importance matrix).
AUC and NSS scores are also computed for these saliency maps. The panel ranking is a measure of the correlation of values in the ground truth and predicted importance matrices.}
	\label{fig:collage}
\end{figure*}




\end{document}